\documentclass[onefignum,onetabnum,review]{siamonline190516}

\usepackage{xfrac}  
\usepackage{lipsum}
\usepackage{amsfonts,dsfont}
\usepackage{graphicx}
\usepackage{epstopdf}
\usepackage{algorithmic}
\usepackage{tikz}
\usepackage{graphicx}

\usepackage{enumitem,comment}
\usepackage{subcaption}
\usepackage{graphicx}
\ifpdf
\hypersetup{
  pdftitle={Inverse Transport},
  pdfauthor={blahlblah..., and Tan Bui-Thanh}
}
\usepackage{multirow}
\usepackage{colortbl}
\fi
\usetikzlibrary{positioning}

\ifpdf
  \DeclareGraphicsExtensions{.eps,.pdf,.png,.jpg}
\else
  \DeclareGraphicsExtensions{.eps}
\fi
\usepackage{comment}

\usepackage{tikz}
\usetikzlibrary{arrows,shapes}
\usetikzlibrary{decorations.pathreplacing,calligraphy}
\usetikzlibrary{patterns}
\usetikzlibrary{arrows,shapes}
\usetikzlibrary{decorations.pathreplacing,calligraphy}

\newsiamremark{remark}{Remark}
\newsiamremark{hypothesis}{Hypothesis}
\crefname{hypothesis}{Hypothesis}{Hypotheses}
\newsiamthm{claim}{Claim}

\headers{A Tikhonov network approach for Inverse Problems}{Hai V. Nguyen and Tan Bui-Thanh}

\title{\TNet: A Model-Constrained Tikhonov Network Approach for Inverse Problems\thanks{The first version of work was an Oden Institute technical report 21-09, May 11, 2021, and then presented at the Uncertainty Quantification and Probabilistic Modeling Technical Thrust Area, USNCCM, May 17, 2021.
}}

\author{Hai V. Nguyen\thanks{Department of Aerospace Engineering and
  Engineering Mechanics, the University of Texas at Austin, Texas (\email{hainguyen@utexas.edu})} \and Tan Bui-Thanh\thanks{Department of Aerospace Engineering and
  Engineering Mechanics, The Oden Institute for Computational
  Engineering and Sciences,  the University of Texas at Austin, Texas
  (\email{tanbui@oden.utexas.edu},
  \url{https://users.oden.utexas.edu/\string~tanbui/}).}
}
\usepackage{amsopn}

\newcommand{\nor}[1]{\left\| #1 \right\|} 
\newcommand{\LRp}[1]{\left( #1 \right)} 
\newcommand{\LRs}[1]{\left[ #1 \right]} 
\newcommand{\LRc}[1]{\left\{ #1 \right\}} 
\newcommand{\pp}[2]{\frac{\partial #1}{\partial #2}} 

\newcommand{\mc}[1]{\mathcal{#1}} 
\newcommand{\mb}[1]{\mathbf{#1}} 
\newcommand{\half}{\frac{1}{2}}
\newcommand{\halfv}[1]{\frac{#1}{2}}

\newcommand{\W}{W}

\newcommand{\WO}{\W^\text{0}}
\newcommand{\WI}{\W^\text{I}}

\newcommand{\DNN}{\Psi}
\newcommand{\F}{\mc{G}}

\newcommand{\bs}[1]{\boldsymbol{#1}}
\renewcommand{\P}{U}

\newcommand{\Ucov}{\Gamma}
\newcommand{\Ucovinv}{\Ucov^{-1}}

\newcommand{\Ycovinv}{\Lambda^{-1}}

\newcommand{\Pbar}{\overline{\P}}

\newcommand{\bb}{{\bf b}}

\newcommand{\bbO}{\bb^\text{0}}
\newcommand{\bbI}{\bb^\text{I}}
\newcommand{\B}{B}

\newcommand{\Y}{Y}
\newcommand{\Ybar}{\overline{\Y}}
\newcommand{\yb}{\bs{y}}
\newcommand{\ybbar}{\overline{\yb}}
\newcommand{\ybobs}{\bs{y}^{obs}}
\newcommand{\One}{\mathds{1}}
\newcommand{\nt}{n_{\mathrm{t}}}
\newcommand{\n}{n}
\newcommand{\m}{m}
\newcommand{\R}{{\bs{\mathbb{R}}}}
\newcommand{\I}{I}

\newcommand{\G}{G}

\newcommand{\pb}{\bs{u}}
\newcommand{\ub}{\pb}
\newcommand{\ubNDL}{\ub^\texttt{nDNN}}
\newcommand{\ubMCDL}{\ub^\texttt{mcDNN}}
\newcommand{\ubTNET}{\ub^\texttt{TNet}}

\newcommand{\pbbar}{\overline{\pb}}
\newcommand{\ubbar}{\pbbar}

\newcommand{\w}{w}
\newcommand{\wb}{\bs{\w}}
\newcommand{\s}{s}
\newcommand{\FW}{\mc{F}}
\newcommand{\fb}{\bs{f}}
\newcommand{\etab}{\bs{\eta}}

\usepackage{hyperref}
\hypersetup{
    colorlinks=true,
    linkcolor=blue,
    filecolor=magenta,      
    urlcolor=red,
    pdftitle={Model-Constrained Deep Learning Approaches for Inverse Problems},
    pdfpagemode=FullScreen,
    }

\newcommand{\figlab}[1]{\label{fig:#1}}
\newcommand{\eqnlab}[1]{\label{eq:#1}}
\newcommand{\theolab}[1]{\label{theo:#1}}
\newcommand{\corolab}[1]{\label{coro:#1}}
\newcommand{\propolab}[1]{\label{propo:#1}}
\newcommand{\lemlab}[1]{\label{lem:#1}}

\newcommand{\tablab}[1]{\label{tab:#1}}

\newcommand{\seclab}[1]{\label{sect:#1}}

\newcommand{\eval}[2][\right]{\relax \ifx#1\right\relax \left.\fi#2#1\rvert}
\renewcommand{\epsilon}{\varepsilon}
\newcommand{\epsb}{\bs{\epsilon}}

\newcommand{\yT}{\Tilde{\bs{y}}}
\newcommand{\nab}{\nabla_{\yb}}

\newcommand{\nDNN}{\texttt{nDNN}}
\newcommand{\TNet}{\texttt{TNet}}
\newcommand{\mcDNN}{\texttt{mcDNN}}

\newcommand{\myred}[1]{{\color[rgb]{0.65,0.0,0.0} #1}}

\begin{document}

\maketitle

\begin{abstract}
  Deep Learning (DL), in particular deep neural networks (DNN), by default is purely data-driven and in general does not require physics. This is the strength of DL but also one of its key limitations
when applied to science and engineering problems in which underlying
physical properties\textemdash such as stability, conservation, and
positivity\textemdash and desired accuracy need to be achieved. DL methods in their original forms are not capable of respecting
the underlying mathematical models or achieving desired accuracy even in big-data regimes. On the other hand, many data-driven science and engineering problems, such as inverse problems, typically have limited  experimental or
observational data, and DL would overfit the data in this case. Leveraging information encoded in the underlying mathematical models, we argue, not only compensates missing information in low data regimes but also provides opportunities to equip DL methods with the underlying physics, and hence promoting better generalization. This paper develops a model-constrained deep learning approach and its variant \TNet{}\textemdash a Tikhonov neural network\textemdash
that are capable of  learning not only information hidden in the training data but also in the underlying mathematical models to solve inverse problems governed by partial differential equations. We provide the constructions and some theoretical results for the proposed approaches. We show that data randomization can enhance not only the smoothness of the networks but also their generalizations.
Comprehensive numerical results 
not only confirm the theoretical findings but also show that with even as little as $20$ training data samples for
1D deconvolution, $50$ for inverse 2D heat conductivity problem, $100$ for inverse initial conditions for time-dependent 2D Burgers' equation, and $50$ for inverse initial conditions for 2D Navier-Stokes equations, \TNet{} solutions can be as accurate as Tikhonov solutions while being several orders of magnitude faster. This is possible owing to the model-constrained term, replications, and randomization. 


\end{abstract}



\section{Introduction}

Inverse problems  are
pervasive in scientific
discovery and decision-making for complex, natural, engineered, and
societal systems.
 They are perhaps the most popular mathematical
approaches for {\em enabling predictive scientific simulations} that
{\em integrate observational/experimental data, simulations and/or
models} \cite{OliverReynoldsLiu08,KaipioSomersalo05, Tarantola05}. Many engineering and science systems are governed by parametrized partial differential equations (PDE). Computational PDE-constrained inverse problems face not only the ill-posed nature\textemdash namely, non-existence, non-uniqueness, and instability of inverse solutions\textemdash but also the computational expense of solving the underlying PDEs. Computational inverse methods typically require the PDEs to be solved at many realizations of parameter and the cost is an (possibly exponentially) increasing function of the parameter dimension. The fast growth of this cost is typically associated with the curse of dimensionality.  Inverse problems for practical
  complex systems \cite{Alifanov94, OliverReynoldsLiu08,
    KomatitschRitsemaTromp02,
    Bui-ThanhBursteddeGhattasEtAl12_gbfinalist,
    LefebvreBozdaCalandraEtAl13} however possess this high dimensional parameter space challenge. 
Thus, mitigating the cost of repeatedly solving the underlying PDE has been of paramount importance in computational PDE-constrained inverse problems.

The field of Machine Learning (ML) typically refers
to computational and statistical methods for the automated detection of
meaningful patterns in data \cite{Bishop06, ShwartzEtAl14, MohriEtAl12}. While Deep Learning (DL) \cite{Goodfellow-et-al-2016}, a subset of machine learning, has
proved to be state-of-the-art methods in many fields of computer
sciences such as computer
vision, speech recognition, natural language processing, etc, and its presence in the
scientific computing community is, however, mostly limited to off-the-shelf applications of
deep learning. Unlike classical scientific
computational methods, such as finite element methods \cite{Ciarlet02,BrennerScott02,ErnGuermond04}, in which
solution accuracy and reliability are guaranteed under regularity
conditions, standard DL methods  are often far from providing reliable and accurate
predictions for science and engineering applications.  The reason is that though the approximation capability of deep learning,
 e.g. via Deep Neural Networks (DNN),
 is as good as classical methods in
 approximation theory \cite{Cybenko1989,hornik1989multilayer,Zhou17,johnson2018deep}, DL accuracy is hardly attainable in general due to
 limitation in training.
 It has been shown that the training problem is
 highly nonlinear and non-convex, and that the gradient of loss
 functions can explode or vanish \cite{Hochreiter01gradientflow}, thus possibly preventing any gradient-based
 optimization methods from reliably
 converging to a minimizer.
 Even when converged, the prediction of the
 (approximate) optimal deep learning model can be prone to over-fitting
 and can have poor
 generalization error.

Many data-driven inverse problems in science and engineering problems have limited  experimental or observational data, e.g. due to the cost of placing sensors (e.g. digging an oil well can cost million of dollars) or the difficulties of placing sensors in certain regions (e.g. deep ocean bottoms). DL, by design, does not require physics, but data. This is the strength of DL. It is also the key limitation to science and engineering problems in which underlying physics needs to be respected 
and higher accuracy is required. In this case, purely data-based DL approaches are prone to over-fitting and thus incapable of respecting the physics or providing the desired accuracy.
Similar to least squares finite element methods \cite{bochev2006least}, we can train a DNN solution constrained by the PDE residual as  a regularization \cite{raissi2017physics2, RaissiEtAl2019, 
  RaissiKarniadakis2018, RaissiEtAl2017, YangPerdikaris2019, TripathyBilionis2018, DeepXDE21,fPINNs19}). 
  Such an approach attempts to learn an approximate solution by making the $L^2$-norm of PDE residual  small. 
  While universal approximation results (see, e.g., \cite{Cybenko1989,hornik1989multilayer,Zhou17,johnson2018deep,UnifedUniversalBui21}) could ensure any desired accuracy with a sufficiently large number of neurons, practical network architectures are moderate in both depth and width, and hence the number of weights and biases. 
    Therefore, the accuracy of learning PDE solutions in function spaces can be limited.
    
    We are interested in {\em parametrized PDEs}\textemdash that are pervasive in design, control, optimization, inference, and uncertainty quantification. Attempts using pure data-driven deep learning to learn the parameter to observable map have been explored (see, e.g., \cite{Kojima17, WHITE20191118, Pestourie2020, Tahersima2019, Peurifoyeaar4206, Kojima17, DNNInverseNanoPhotonnics20, Jiang2020, singh2017machine}).
    Approaches using autoencoder spirit that train a forward network first and then an inverse network in tandem \cite{LiuEtAl18,Luo21} or both of them simultaneously \cite{goh2021solving} have also been proposed.
The work in     \cite{zhao2022learning}  proposes
to use a graph neural network to approximate forward solver and and fully connected neural network  to learn a regularization via the prior knowledge. Once trained, both networks are deployed in a Tikhonov-like regularization algorithm to obtain the inverse solution.
    While successes are reported, generalization capability, and hence success, could be limited to regimes seen in the training data as the governing equations\textemdash containing  most, if not all, information about the underlying physics\textemdash are not involved in the training. 
 
    In order to take into account  the underlying problem, a natural direction is to deploy deep learning methods as  surrogates for  expensive or difficult components in traditional methods. Such a 
    hybrid approach  can enjoy the benefits of both sides.
    For example, learning regularizers to penalize certain undesirable features has been proposed for both inverse  \cite{li2020nett,lunz2018adversarial} and imaging \cite{aggarwal2018modl} problems. Once trained, these regularizers can be used in any traditional inverse or imaging methods.
    The main disadvantage of these approaches is that they may still experience the same computational expense as traditional methods when the forward map is the most expensive part. Learning the forward map \cite{FouierOP,zhao2022learning, allen2022physical, pfaff2020learning} is thus desirable, though it may not be considered as a model-aware approach. 
   
    A logical alternative is thus to
    constrain the learning of inverse solution
    with the underlying governing equations and/or physics. The work in \cite{adler2017solving}  proposes to partially learn  the gradient of a Tikhonov functional and uses the learned gradient to perform a gradient-based optimization method for solving imaging problems. A natural extension of the physics-informed neural network framework \cite{Chen20, RAISSI2019686, DeepXDE21,lu2021physicsinformed} is to
     train two networks, one for  solutions and another for unknown parameters.  
    In an attempt to mimic the  traditional PDE-constrained approach, \cite{fan2020solving,BN17_FEniCS_plus_DNN}
     parametrize the unknown parameters using feed-forward neural networks whose weights/biases are then found by an optimization approach constrained by the Navier-Stokes equations and heat equations. 
     These methods, however, may not be efficient as new observational data (corresponding to new unknown parameters) requires retraining. It is also not clear how to extend them to statistical inverse problems. 
    
    Learning inverse maps constrained by the underlying governing equations has also been investigated. 
    The work in \cite{pakravan2021solving}, similar to \cite{fan2020solving,BN17_FEniCS_plus_DNN}, presents an autoencoder-like approach in which the encoder is the inverse map and the decoder is the numerical solutions of the underlying governing equations evaluated at observational points. The network weights/biases are found by minimizing the data misfit. Taking both the data misfit and the regularization into account as in the traditional Tikhonov inversion approach,
\cite{jin2020physics} solves 1D seismic inversion methods with promising results.
The beauty of this approach is that, once trained, the neural network can be deployed to approximately solve inverse problems in real-time. 

    The main contributions of this paper\textemdash a detailed extension of an approach set forth in a more general framework in \cite{NguyenBui_mcDNN21}\textemdash is as follows. Unlike similar and independent work in \cite{pakravan2021solving, fan2020solving,BN17_FEniCS_plus_DNN, jin2020physics}, our \mcDNN{} has a theoretical foundation, from which and numerical evidence, we infer that \mcDNN{} may not  be a good learning strategy for inverse problems as it could be biased by the training data, though it is interpretable compared to a purely data-driven counterpart. This motivates us to develop a new model-constrained deep learning approach, called \TNet{}, designed to learn the Tikhonov inverse solution, and indeed it recovers Tikhonov regularized solutions for linear inverse problems.
    We also propose to randomize the training data and rigorously justify randomization as an implicit regularization that could improve the generalization of the proposed deep-learning approaches. We provide comprehensive numerical results to support our developments for 1D deconvolution, inverse heat conductivity, and inverse initial conditions for both time-dependent 2D Burgers' and 2D Navier-Stokes equations.

  The paper is organized as follows. 
  In 
  \cref{sect:forwardInverse} we introduce nonlinear inverse problems, and a data-driven naive DNN (\nDNN) approach.
   The goal of \cref{sect:MCDL} is to present a model-constrained DNN (\mcDNN) approach designed to learn the inverse map while being constrained by the parameter-to-observable map of the underlying discretized PDE.  
   Though \mcDNN{} is interpretable, it could be biased toward training data. This leads us to develop \TNet{}\textemdash a Tikhonov neural network\textemdash in \cref{sect:TNET} that aims to learn the Tikhonov solver while removing unnecessary biases. We show that data randomization can make \TNet{} not only more robust but also generalize better: thanks to the model-constrained training. In \cref{sect:numerics} and the supplementary document, comprehensive numerical results supporting our developments are presented for 1D deconvolution, inverse heat conductivity, and inverse initial conditions for both time-dependent 2D Burgers' and 2D Navier-Stokes equations.
We conclude the paper with future research directions in \cref{sect:conclusions}. Proofs of the theoretical results, practical implementation aspects of our proposed approaches, and specifications of trainings are provided in the supplementary document.

\section{Introduction to forward and inverse problems}
\seclab{forwardInverse}
The following notations are used in the paper.    Boldface lowercases are reserved for (column) vectors, and uppercase letters are for matrices.
We denote by $\pb \in \R^\m$ the parameters  sought in the inversion or the parameter of interest (PoI), by $\wb \in \R^\s$ the  forward states, by  $\F: \R^\s \to \R^\n$ the forward map (computing some observable quantity of interest), and by $\yb \in \R^\n$ the observations given by
\begin{equation}\eqnlab{pointwiseObs}
\yb := \F\LRp{\wb\LRp{\ub}}  + \etab, 
\end{equation}
where $\etab$ is some additive observation noise.
The parameter-to-observable (PtO) map is the composition of the forward map $\F$ and the states, i.e., $\F\circ\wb$. However for simplicity of the exposition, we do not distinguish it from the forward map and thus we also write $\F: \R^\m \ni \ub \mapsto \F\LRp{\ub}:= \F\LRp{\wb\LRp{\ub}}  \in  \R^\n$.
 The forward
state is the solution of the forward equation
\begin{equation}
\eqnlab{forward}
\FW\LRp{\ub,\wb} = \fb.
\end{equation}

Assume
that \cref{eq:forward} is well-posed so that, for
a given set of parameters $\ub$, one can (numerically for example) solve for the corresponding
forward states $\wb = \wb\LRp{\ub} := \FW^{-1}\LRp{\fb}$. 
In the forward problem, we compute observational data $\yb$ via \cref{eq:pointwiseObs} given a set of parameter $\ub$. In the inverse problem, we seek to determine the unknown parameter $\ub$ given some observational data $\yb$, that is, we wish to construct the inverse of $\F$. Since $\m$ is typically (much) larger than $\n$ for many practical problems, the parameter-to-observable map $\F$ is not invertible even when $\F$ is linear. The inverse task is thus ill-posed and notoriously challenging as a solution for $\ub$ may not exist, even when it may, it is not unique nor stably depends on the data $\yb$. An approximate solution is typically sought via (either deterministic or statistical) regularization.

Given the popularity of emerging machine learning, in particular deep neural networks (DNN), methods, we may attempt to apply a naive pure data-driven DNN (\nDNN) to learn the (ill-posed) inverse of $\F$, e.g.,
\begin{equation}
\tag{\texttt{nDNN}}
	\eqnlab{optDNN}
	\min_{\bb,\W}  \mc{L}_{\nDNN}  = 
	\half\nor{\P - \DNN\LRp{\Y,\W,\bb}}^2 + \halfv{\alpha_1} \nor{\W}^2 + \halfv{\alpha_2} \nor{\bb}^2,
\end{equation}
where $\Psi$ is a DNN with weight matrix $\W$ and bias vector $\bb$ and the last two terms are regularizations for weights and biases with nonnegative regularization parameters $\alpha_1$ and $\alpha_2$.  Here, $\Y \in \R^{\n \times \nt}$ is the data matrix concatenating $\nt$ observational data $\yb^i$, $i =1,\hdots,\nt$, and $\P \in \R^{\m \times \nt}$ is the parameter matrix concatenating the corresponding parameter vectors $\pb^i$.
This approach completely disregards the underlying mathematical model \cref{eq:pointwiseObs}-\cref{eq:forward}.  Even for linear inverse problem\textemdash for example, $\F\LRp{\pb} = \G\pb$ and there is no error in computing the data so that $\Y = \G\P$\textemdash and linear DNN   such as 
$\DNN = \W\Y + \B$, where $\B := \bb\One^T$,
 and thus the optimal weight $\WO$ and bias $\bbO$ for \cref{eq:optDNN} are given as
\begin{align*}
    \WO &= \P\LRp{\I - \frac{1}{\nt + \alpha_2}\One\One^T}\Y^T \LRs{\Y\LRp{\I - \frac{1}{\nt + \alpha_2}\One\One^T}\Y^T + \alpha_1\I}^{\dagger}\\
    \bbO &= \frac{1}{1+\alpha_2/\nt}\LRp{\pbbar - \WO\ybbar}, \\
\end{align*}
where $\pbbar := \frac{1}{\nt}\P\One$ and $\dagger$ denotes the pseudo-inverse operation, 
 it is not clear if the \nDNN{} inverse solution
\[
    \ubNDL = \WO\ybobs + \bbO.
\]
provides an approximate solution to the original inverse problem 
\[
\min_{\ub}\nor{\ybobs - \G\ub}^2
\]
in an interpretable sense. This is a disadvantage of pure data-driven approaches.

The data-driven nature of DNN could be claimed as an advantage. However, DNN can be considered as an ``interpolation" method and thus can generalize well only for scenarios that have been seen in or are sufficiently close to the training data set $\LRc{\P,\Y}$. This implies a possible enormous amount of training data to learn the inverse of highly non-linear problems. 
In practical sciences and engineering problems, this extensive data regime is unfortunately rarely the case due to the high cost of placing sensors or the
difficulties in placing sensors in certain regions. {\em In order for a DNN to generalize well in insufficient/low data regimes, it should be equipped with information encoded in the forward model \cref{eq:pointwiseObs}-\cref{eq:forward} that is not covered in the data set. } 
{\em Such a physics encoding also supplies meaningful interpretations to DNN inverse solutions as we shall show.}
  The question is how to inform DNN about the underlying models?
In the following, we construct two {\em DNNs to learn the inverse of the PtO map $\F$ not only by information hidden in the training data but also by
satisfying the forward equations exactly 
at the training points}.

\section{Model-Constrained Deep Neural Network (\mcDNN{}) for learning the inverse map}
\seclab{MCDL}

\def\layersep{1.7cm}
\def\nodeinlayersep{1.0cm}
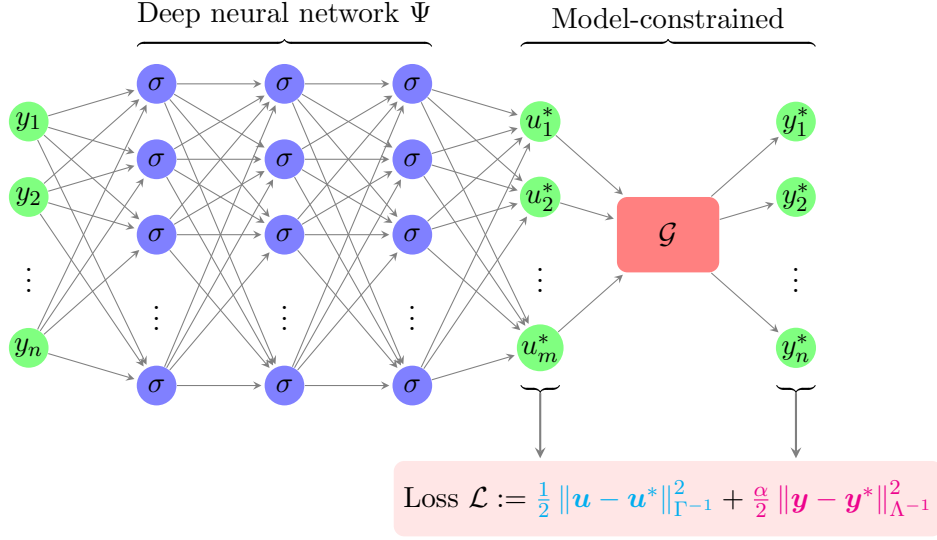
\begin{figure}[h!t!b!]
\centering
\begin{tikzpicture}[
    node distance=\layersep,
    edge/.style={-stealth,shorten >=1pt, draw=black!50, thin},
    neuron/.style={circle,fill=black!25,minimum size=15pt,inner sep=0pt},
    operator/.style={rectangle,fill=green!,minimum height= \nodeinlayersep, minimum width= 0.8 * \layersep, inner sep=0pt, rounded corners},
    input neuron/.style={neuron, fill=green!50,minimum size=15pt},
    output neuron/.style={neuron, fill=green!50,minimum size=15pt},
    hidden neuron/.style={neuron, fill=blue!50},
    Forward map/.style={operator, fill=red!50},
    annot/.style={text width=4em, text centered},
    every node/.style={scale=1.0},
    node1/.style={scale=2.0}
]
    \foreach \name / \y in {1,...,4}
        {\ifnum \y=3
            \node (I-\name) at (0,-\nodeinlayersep * \y - \nodeinlayersep * 0.5) {$\vdots$};
        \else
            \ifnum \y=4
                \node[input neuron] (I-\name) at (0,-\nodeinlayersep * \y - \nodeinlayersep * 0.5) {$y_{n}$};
            \else
                \node[input neuron] (I-\name) at (0,-\nodeinlayersep *\y - \nodeinlayersep * 0.5 ) {$y_{\y}$};
            \fi
        \fi}
        
    \foreach \name / \y in {1,...,4}
        {\ifnum \y=3
            \node (O-\name) at (4*\layersep,-\nodeinlayersep *\y - \nodeinlayersep * 0.5) {$\vdots$};
        \else
            \ifnum \y=4
                \node[input neuron] (O-\name) at (4*\layersep,-\nodeinlayersep *\y - \nodeinlayersep * 0.5) {$u_{m}^*$};
            \else
                \node[input neuron] (O-\name) at (4*\layersep,-\nodeinlayersep *\y - \nodeinlayersep * 0.5) {$u_{\y}^*$};
            \fi
        \fi}
        
    \node[Forward map] (FM) at (5*\layersep,-\nodeinlayersep * 3.) {$\mathcal{G}$};
    
    \foreach \name / \y in {1,...,4}
        {\ifnum \y=3
            \node (obs_p-\name) at (6*\layersep,-\nodeinlayersep *\y - \nodeinlayersep * 0.5) {$\vdots$};
        \else
            \ifnum \y=4
                \node[input neuron] (obs_p-\name) at (6*\layersep,-\nodeinlayersep *\y - \nodeinlayersep * 0.5) {$y_{n}^*$};
            \else
                \node[input neuron] (obs_p-\name) at (6*\layersep,-\nodeinlayersep *\y - \nodeinlayersep * 0.5) {$y_{\y}^*$};
            \fi
        \fi}

    \newcommand \Nhidden{3}
    \foreach \N in {1,...,\Nhidden} {
        \foreach \y in {1,...,5} { 
            \ifnum \y=4
                \node at (\N*\layersep,-\y*\nodeinlayersep) {$\vdots$};
            \else
                \node[hidden neuron] (H\N-\y) at (\N*\layersep,-\y*\nodeinlayersep ) {$\sigma$};
        \fi
      }
    }

    \foreach \source in {1,2,4}
        \foreach \dest in {1,...,3,5} 
            \draw[edge] (I-\source) -- (H1-\dest);
    
    \foreach [remember=\N as \lastN (initially 1)] \N in {2,...,\Nhidden}
      \foreach \source in {1,...,3,5} 
          \foreach \dest in {1,...,3,5} 
              \draw[edge] (H\lastN-\source) -- (H\N-\dest);
              
    \foreach \source in {1,...,3,5} 
        \foreach \dest in {1,2,4}
            \draw[edge] (H\Nhidden-\source) -- (O-\dest);
    
    \foreach \source in {1,2,4}
        \draw[edge] (O-\source) -- (FM);
        
    \foreach \source in {1,2,4}
        \draw[edge] (FM) -- (obs_p-\source);    
        
    \node[rectangle,fill=pink!40,minimum height= \nodeinlayersep, minimum width= 0.8 * \layersep, rounded corners] (L_misfit) at (5*\layersep, -6.5*\nodeinlayersep) { Loss $\mathcal{L} := \textcolor{cyan}{\frac{1}{2} \nor{\ub - \ub^*}_{\Ucovinv}^2} + \textcolor{magenta}{\frac{\alpha}{2} \nor{\yb - \yb^*}_{\Ycovinv}^2}$};
    
    \draw [decorate, decoration = {calligraphic brace,mirror}, thick] (3.85*\layersep,-5.0*\nodeinlayersep) -- (4.15*\layersep,-5.0*\nodeinlayersep);
    \draw[edge, thick] (4*\layersep,-5.1*\nodeinlayersep) -- (4*\layersep, -6.0*\nodeinlayersep);

    \draw [decorate, decoration = {calligraphic brace,mirror}, thick] (5.85*\layersep,-5.0*\nodeinlayersep) -- (6.15*\layersep,-5.0*\nodeinlayersep);
    \draw[edge, thick] (6*\layersep,-5.1*\nodeinlayersep) -- (6*\layersep, -6.0*\nodeinlayersep);
    
    \draw [decorate, decoration = {calligraphic brace}, thick] (.85*\layersep,-.5*\nodeinlayersep) -- (3.15*\layersep,-.5*\nodeinlayersep) node[pos=0.5,above=0.1cm,black]{ Deep neural network $\Psi$};
    
    \draw [decorate, decoration = {calligraphic brace}, thick] (3.85*\layersep,-.5*\nodeinlayersep) -- (6.15*\layersep,-.5*\nodeinlayersep) node[pos=0.5,above=0.1cm,black]{ Model-constrained};
    
\end{tikzpicture}
\caption{\normalsize Model-constrained neural network architecture \mcDNN. The observables $\yb$ is fed into the neural network $\Psi$. The parameter  $\ub^*$ predicted by the netork is pushed through the PtO map $\F$ to  generate the corresponding predicted observations $\yb^*$. Both predicted parameters and observations are compared with ground truth $\ub$ and $\yb$, respectively, to provide the mean-square error in the loss function $\mc{L}$.}
\figlab{mcDNN_architecture}
\end{figure}

We propose to learn the inverse map via DNN constrained by the forward map as 
\begin{equation}
\tag{\texttt{mcDNN}}
	\eqnlab{optNonlinear}
	\min_{\bb,\W} \mc{L}_{\mcDNN} :=\half\nor{\P - \DNN\LRp{\Y,\W,\bb}}_{\Ucovinv}^2 + \halfv{\alpha} \nor{\Y - \F\LRp{\DNN\LRp{\Y,\W,\bb}}}_{\Ycovinv}^2,
\end{equation}
where $\Psi$ is a DNN learning the map from observable data $\yb$ to parameter $\ub$ with weight matrix $\W$ and bias vector $\bb$.
We have introduced Frobenius norm weighted by $\Ucovinv$  in the first term as
\[
\nor{\P - \DNN\LRp{\Y,\W,\bb}}_{\Ucovinv}^2 := \nor{\Ucov^{-\half}\LRp{\P - \DNN\LRp{\Y,\W,\bb}}}^2, 
\]
and similarly for the second term weighted by $\Ycovinv$.
Unlike the naive purely data-driven DNN approach \cref{eq:optDNN}, the model-constrained \cref{eq:optNonlinear} makes the DNN $\Psi$ aware that the training data is generated by the forward map $\F$. This is done by requiring the output of the DNN\textemdash approximate unknown parameter $\ub$ for a given data $\yb$ as the input\textemdash when pushed through the forward model $\F$, reproduces the data $\yb$. The model-aware term $\halfv{\alpha} \nor{\Y - \F\LRp{\DNN\LRp{\Y,\W,\bb}}}_{\Ycovinv}^2$ can be considered as a physics-aware regularization approach for \mcDNN{} (compared to the non-physical regularizations in \cref{eq:optDNN}). The architecture of \mcDNN{} is presented in \cref{fig:mcDNN_architecture}.

In order to shed light on our \mcDNN{} approach let us choose a linear activation function such that the one-layer DNN model $\DNN\LRp{\Y,\W,\bb}$ for leaning the inverse map can be written as $\W\Y + \B$, where $\B := \bb\One^T$. We also assume that the forward map is linear. 
For linear inverse problem with linear DNN, the model-constrained training problem \cref{eq:optNonlinear} becomes
	\begin{equation}
	\eqnlab{optLinear}
	\min_{\bb,\W}\half\nor{\P - \LRp{\W\Y + \B}}_{\Ucovinv}^2 + \halfv{\alpha} \nor{\Y - \G\LRp{\W\Y + \B}}_{\Ycovinv}^2.
	\end{equation}
	\begin{lemma}
	\lemlab{TNET}
		The optimal solution $\WI$ and $\bbI$ of the DNN training problem \cref{eq:optLinear} satisfies
		\begin{align*}
		\bbI &= \LRp{\Ucovinv + \alpha \G^T \Ycovinv \G}^{-1}\LRs{\Ucovinv \ubbar + \alpha \G^T \Ycovinv \ybbar - \LRp{\Ucovinv \Pbar \, \Ybar^{\dagger} + \alpha\G^T \Ycovinv \Ybar \, \Ybar^{\dagger}} \ybbar}, \\
		\WI &= \LRp{\Ucovinv + \alpha \G^T \Ycovinv \G}^{-1}  \LRs{\Ucovinv \Pbar \, \Ybar^{\dagger} + \alpha\G^T \Ycovinv \Ybar \, \Ybar^{\dagger}},
		\end{align*}
		where $\pbbar := \frac{1}{\nt}\P\One$ and $\ybbar := \frac{1}{\nt}\Y\One$ are the column-average of the training parameters and data, $\Ybar := \Y - \ybbar\One^T$, and $\Pbar := \P - \pbbar\One^T$.
	\end{lemma}
	
\begin{corollary}[\mcDNN{} is a Tikhonov solver]
\corolab{optDNN}
For a given testing/observational data $\ybobs$, the \mcDNN{} inverse solution $\ubMCDL$ of \cref{eq:optLinear} is given by
\begin{multline*}
    \ubMCDL = \LRp{\Ucovinv + \alpha \G^T \Ycovinv \G}^{-1} \\  \LRs{\Ucovinv \ubbar + \alpha \G^T \Ycovinv \ybbar 
    + \LRp{\Ucovinv \Pbar \, \Ybar^{\dagger} + \alpha\G^T \Ycovinv \Ybar \, \Ybar^{\dagger}} \LRp{\ybobs - \ybbar}}
\end{multline*}
which is exactly the solution of the following Tikhonov regularized linear inverse problem
\[
\min_{\ub}\half\nor{\ybobs - \G\ub}_{\Ycovinv}^2 + \frac{1}{2\alpha}\nor{\ub - \ub_0^{\texttt{mcDNN}}}_{\Ucovinv}^2,
\]
where
\begin{equation}
\eqnlab{u0mcDNN}
\ub_0^{\texttt{mcDNN}} =  \pbbar +\Pbar \, \Ybar^{\dagger} \LRp{\ybobs - \ybbar} - \alpha \Ucov \G^T \Ycovinv \LRp{\I - \Ybar \, \Ybar^{\dagger}} \LRp{\ybobs - \ybbar}.
\end{equation}

\end{corollary}
The results of \cref{coro:optDNN} shows that the \mcDNN{} inverse solution $\ubMCDL$ is equivalent to a Tikhonov-regularized inverse solution with a data-informed reference parameter $\ub_0$ that depends on the training set $\LRc{\P,\Y}$ and the given observational data $\ybobs$. In other words, {\em the model-constrained deep learning \mcDNN{} approach is interpretable in the sense that it provides data-informed Tikhonov-regularized inverse solutions.}



\section{Tikhonov neural network (\TNet) for learning the inverse map}
\seclab{TNET}

We observe that the reference parameter $\ub_0^{\texttt{mcDNN}}$ in \cref{coro:optDNN} depends on the training data, and thus the model generalization depends on the amount of training data. 
In other words, \mcDNN{} solution $\ubMCDL$ could have a strong bias to the training data and may limit the generalization which is not desirable especially for scenarios that are not very close to the training ones.
On the other hand, in the classical Tikhonov regularization framework, the reference parameter is fixed and independent of the observable data. From a statistical point of view, the reference parameter is typically the mean of the prior distribution of the parameter of interest (PoI) $\ub$, which reflects the {\em a priori} belief on how the PoI should look like on average. Synergizing \mcDNN{} and Tikhonov regularization ideas, we propose a Tikhonov neural network \TNet\textemdash a semi-supervised  model-constrained learning approach\textemdash where, unlike \mcDNN{}, the unknown PoI predicted by the DNN $\Psi$ are forced to be close the PoI prior mean $\ub_0$ as 

\begin{equation}
\tag{\texttt{TNet}}
	\eqnlab{TNETNonlinear}
	\min_{\bb,\W} \mc{L}_{\TNet} :=\half\nor{\P_0 - \DNN\LRp{\Y,\W,\bb}}_{\Ucovinv}^2 + \halfv{\alpha} \nor{\Y - \F\LRp{\DNN\LRp{\Y,\W,\bb}}}_{\Ycovinv}^2,
\end{equation}
where $\P_0 = \ub_0 \One^T$ is the matrix whose columns are the mean prior of the PoI. 
Consequently, the architecture of \TNet{} is the same as \mcDNN{} in  \cref{fig:mcDNN_architecture} except with $\ub$ replaced by $\ub_0$. Applying \cref{coro:optDNN} to \TNet{} for linear inverse problem with linear DNN we have the following result.

	
\begin{corollary}[\TNet{} is a Tikhonov solver]
\corolab{TNETDNN}
For a given testing/observational data $\ybobs$, the \TNet{} inverse solution $\ubTNET$  is given by
\begin{equation}
\eqnlab{TNetsolution}
    \ubTNET = \LRp{\Ucovinv + \alpha \G^T \Ycovinv \G}^{-1} \\  \LRs{\Ucovinv \ub_0 + \alpha \G^T \Ycovinv \ybbar 
    + \alpha\G^T \Ycovinv \Ybar \, \Ybar^{\dagger} \LRp{\ybobs - \ybbar}}
\end{equation}
which is exactly the solution to the following Tikhonov regularized linear inverse problem
\[
\min_{\ub}\half\nor{\ybobs - \G\ub}_{\Ycovinv}^2 + \frac{1}{2\alpha}\nor{\ub - \ub_0^{\texttt{TNet}}}_{\Ucovinv}^2,
\]
where
\[
\ub_0^{\texttt{TNet}} =  \ub_0 - \alpha \Ucov \G^T \Ycovinv \LRp{\I - \Ybar \, \Ybar^{\dagger}} \LRp{\ybobs - \ybbar}.
\]
\end{corollary}

Two observations are in order. First, \cref{coro:TNETDNN} shows that the \TNet{} inverse solution $\ubTNET$ is exactly the Tikhonov-regularized inverse solution with the true prior mean $\ub_0$ as the reference parameter provided that the observation data $\Y$ is full row rank. This holds, for example,  when the number of independent data is at least the same as the number of observations. Even when this happens, \mcDNN{} solution in \cref{coro:optDNN} does not coincide with the Tikhonov solution as the first two terms on the right-hand side of \cref{eq:u0mcDNN} only reduce to $\ub_0$ in the limit of infinite training data (via the law of large numbers).
Second, training data for PoI $\ub$ is not needed (thanks to the semi-supervised learning nature of \TNet{}). This is particularly useful when we like to use actual observational data in training. 

The next result is a highlight of our method in that our model-constrained approaches satisfy the governing equation exactly at the training points. The proof and aspects on practical implementation are provided in \cref{sect:proofs} and \cref{sect:implementation}.
\begin{proposition}[Exactly satisfying the governing equations at training points]
Both \mcDNN{} and \TNet{} inverse solutions satisfy the governing equations exactly at all training points.
\propolab{HardC}
\end{proposition}

We now show that data randomization enhances not only the generalization of \TNet{} solution but also its robustness to observational noise. To begin, we randomize a generic data vector $\yb$, e.g. one column of $\Y$, as follows
\begin{equation}
    \yT = \yb + \epsb,
\end{equation}
where a Gaussian noise vector $\epsb \sim \mc{N}\LRp{0, \lambda^2 \I}$ with variances $\lambda^2$ is added to the data. We emphasize that the following  arguments also hold for any random noise vector with independent components, each of which is a random variable with zero mean and variances $\lambda^2$.  Let $\mathbb{E}\LRs{\cdot}$ denote the expectation with respect to $\epsb$. Following \cite{an1996effects}, for a generic loss function $\mc{L}\LRp{\yT}$, we perform the Taylor expansion around $\yb$ up to second order to obtain 
\begin{equation}
\eqnlab{Expect_form}
    \mathbb{E}\LRs{\mc{L}\LRp{\yT}} =  \mc{L}\LRp{\yb} + \mathbb{E}\LRs{\eval{\pp{\mc{L}}{\yb}}_{\yb} \epsb} +  \half \mathbb{E}\LRs{ \epsb^T \eval{\pp{^2 \mc{L}}{\yb^2}}_{\yb} \epsb} 
    + \mathbb{E}\LRs{o\LRp{\nor{\epsb}^2}} 
\end{equation}
where we have used sufficient small noise variance $\lambda^2$  so that the high-order term $o\LRp{\nor{\epsb}^2}$, using the standard ``small o" notation, is negligible. 

For training data set with $n_t$ samples, we randomize each sample as
$\yT^i = \yb^i + \epsb^i, i = 1, \hdots, n_t$, where $\epsb_i \sim \mc{N}\LRp{0, \lambda_i^2 \I}$. Note that we can use different noise levels for data randomization. In that case, the \cref{eq:TNETNonlinear} loss  becomes
\begin{equation}
    \eqnlab{TNET_data_aug_loss}
    \mc{L}_{\TNet}^{\text{rand}}  = \sum_{i=1}^{n_t} \underbrace{\half\nor{\ub_0 - \Psi(\yT^i)}_{\Ucovinv}^2 + \halfv{\alpha} \nor{\yT^i - \F\LRp{\Psi\LRp{\yT^i}}}_{\Ycovinv}^2}_{=: \mc{J}\LRp{\yT^i}}.
\end{equation}
Replacing $\mc{L}$ with $\mc{J}{\LRp{\yT_i}}$ in \cref{eq:Expect_form} yields
\begin{equation}
    \mathbb{E}\LRs{\mc{J}\LRp{\yT^i}} \approx  \mc{J}\LRp{\yb^i} + \lambda_i^2 \LRp{\mc{P}_1^i + \mc{P}_2^i + \mc{P}_3^i + \mc{P}_4^i},
\end{equation}
where 
\begin{equation}
    \eqnlab{no_noise_loss}
    \mc{J}\LRp{\yb^i} = \half\nor{\ub_0 - \Psi(\yb^i)}_{\Ucovinv}^2 + \halfv{\alpha} \nor{\yb^i - \F\LRp{\Psi\LRp{\yb^i}}}_{\Ycovinv}^2,
\end{equation}
and the {\em induced penalty} terms are given by
\begin{align*}
    \mc{P}_1^i &= \half \mb{Tr}\LRs{\LRp{\nab{\Psi\LRp{\yb^i}}}^T \Ucovinv \LRp{\nab{\Psi\LRp{\yb^i}}}}, \\
    \mc{P}_2^i &= \halfv{\alpha} \mb{Tr}\LRs{\LRp{\nab \LRs{\yb^i - \F \circ \Psi \LRp{\yb^i}}}^T {\Ycovinv} \LRp{\nab \LRs{\yb^i - \F \circ \Psi \LRp{\yb^i}}}},    \\
    \mc{P}_3^i &= \half \mb{Tr}\LRs{{{\nab^2{\Psi\LRp{\yb^i}}}} \odot {\Ucovinv\LRp{\Psi\LRp{\yb^i} - \ub_0}}}, \\
    \mc{P}_4^i &= \halfv{\alpha} \mb{Tr} \LRs{{\nab^2\LRs{\yb^i - \F \circ \Psi \LRp{\yb^i}}} \odot {{\Ycovinv} \LRp{\yb^i - \F \circ \Psi \LRp{\yb^i}}}},            
\end{align*}
in which $\mb{Tr}\LRp{\cdot}$ is the trace operator, and $\odot$ denotes the dot product of a third-order tensor and  a vector.
It  can be observed that the training loss with randomized data is the sum of the original loss plus four induced regularization terms.
 $\mc{P}_1^i$ is non-negative and  promotes the smoothness of the neural network.  The second term of \cref{eq:no_noise_loss} ensures that $\DNN$ is close to the right inverse of $\F$, and $\mc{P}_2^i$ strengthen this closeness by forcing the derivative of the $\F\circ \DNN$ to the identity. These two effects together behave like a Hermite interpolation in which not only the function values but also the derivatives are required to be matched closely at the training points.  The two terms in \cref{eq:no_noise_loss}
make  $\Psi\LRp{\yb^i} - \ub_0$ and 
$\yb^i - \F\circ \Psi\LRp{\yb^i}$
necessary small, and as a result, 
 $\mc{P}_3^i$ and $\mc{P}_4^i$ can be dominated by  $\mc{P}_1^i$ and $\mc{P}_2^i$, respectively. It is interesting to see that $\mc{P}_3^i$ and $\mc{P}_4^i$  can encourage the second derivatives (and hence extra smoothness) of the neural network $\DNN$ and $\I - \F\circ\DNN$ to be small. In other words, the beauty of  data randomization is that it can promote a $\mc{H}^2$-Sobolev-like Tikhonov regularization for the neural network $\DNN$ via \cref{eq:no_noise_loss}, $\mc{P}_1^i$, and $\mc{P}_3^i$. Moreover, it can further enforce $\DNN$ to be the same as the right inverse of the PtO map $\F$ up to second derivatives via  \cref{eq:no_noise_loss}, $\mc{P}_2^i$, and $\mc{P}_4^i$.
 


Accounting for the data randomization for all training data we can\textemdash after taking the expectation with the random noise $\epsb^i$, $i = 1,\hdots,n_t$\textemdash write \cref{eq:TNET_data_aug_loss} as
\begin{equation}
    \eqnlab{TNET_data_aug_loss_final}
    \begin{aligned}
        \mathbb{E}\LRs{\mc{L}_{\TNet}^{\text{rand}}}  \approx  \mc{L}_{\TNet} 
         + \half \sum_{i=1}^{n_t}\lambda_i^2 \LRp{\mc{P}_1^i + \mc{P}_2^i + \mc{P}_3^i  + \mc{P}_4^i}.
    \end{aligned}
\end{equation}
Thus, on average, the \TNet{} loss $\mc{L}_{\TNet}^{\text{rand}}$ with randomized data 
is approximately the sum of the original \TNet{} loss (without randomization) plus four regularization terms for each training data point. 
These induced regularization terms play a vital role in stimulating the robustness and accuracy of the neural network. Indeed, without data randomization, the \TNet{} loss \cref{eq:TNETNonlinear} simply requires the neural network outputs to be close to the parameter data via the data misfit (the first) term,
and the neural network, when pushed through the PtO map, resembles the observational data via the model-constrained (the second) term. 
Whereas, randomizing the data  enforces not only the smoothness of the neural network $\DNN$ up to second order derivative (through the first term) but also the agreement of the neural network and the right inverse of the PtO map $\F$ up to second order derivatives (through the model-constrained term). Let us summarize the above result in the following theorem.
\begin{theorem}
\theolab{randTNet}
Let $\yT^i = \yb^i + \epsb^i, i = 1, \hdots, n_t$, where $\epsb_i \sim \mc{N}\LRp{0, \lambda_i^2 \I}$. Then 
\begin{equation}
\eqnlab{randTNet}
  \mathbb{E}\LRs{\mc{L}_{\TNet}^{\text{rand}}}  =  \mc{L}_{\TNet} 
         + \half \sum_{i=1}^{n_t}\lambda_i^2 \LRp{\mc{P}_1^i + \mc{P}_2^i + \mc{P}_3^i  + \mc{P}_4^i} + 
\sum_{i=1}^{n_t}\mathbb{E}\LRs{o\LRp{\nor{\epsb^i}^2}}.
\end{equation}
\end{theorem}

\begin{remark}
Note that $\yb^i$ are not necessarily different from each other. However, the Hermite interpolation analogy tells us that we should have as many distinct baseline training points as possible for good generalization. It turns out that we just need a small number of distinct training points to have accurate results, as numerically shown in \cref{sect:numerics}.  
The above randomization approach also holds for \nDNN{} and \mcDNN{} approaches. Indeed, in the  the final expression \cref{eq:TNET_data_aug_loss_final}  we simply replace $\mc{L}_{\TNet}$ by $\mc{L}_{\mcDNN}$ (see \cref{eq:optNonlinear}) and $\ub_0$ by $\ub^i$ in $ \mc{P}_3^i$ for \mcDNN{}. Similarly for \nDNN{}, we  replace $\mc{L}_{\TNet}$ by  $\mc{L}_{\nDNN}$ (see \cref{eq:optDNN}) and remove $ \mc{P}_2^i$ and $ \mc{P}_4^i$.
\end{remark}

\section{Numerical results}
\seclab{numerics}

$ $
{\bf Noise realization.}
For all numerical results, we choose $\lambda =\delta \max \LRp{\yb}$ for all $\lambda_i$ in \cref{eq:randTNet}, where
$\delta$ denotes the relative noise level.

{\bf Data generation and training.}
For non-linear problems in  \cref{sect:Heat_problem}, \cref{sect:Burger} and \cref{sect:2D_NS}, we use a shallow neural network having one hidden layer with 5000  ReLU neurons.
We verified that a dense feed-forward neural network architecture with multiple layers could provide  comparable results but with large training data sets. In small data regimes, i.e. 100 samples,  deep networks perform poorly due to the vanishing gradient problem and/or the bias-variance trade-off problem. Moreover, training a deep learning network faces further challenges \cite{DEEPNET_difficulty_2010, DEEPNET_difficulty_2013} that are beyond the scope of this paper. We thus focus on neural networks with a single hidden layer and this is sufficient to demonstrate the proposed \TNet{} framework.
Regarding optimization algorithm, the default \texttt{ADAM} \cite{kingma2014adam} optimizer in \texttt{JAX}  \cite{jax2018github} is used.
In all numerical results, weights and biases of the neural network are initialized by standard Gaussian distribution and a zero vector, respectively, using the same random seed. Therefore, we begin the training process with the same network for all cases.

In order to be fair, within any comparison we use the same random seeds for noise. To ensure that more training data can offer more information,
the training data set is generated
in a nested manner,
e.g., $n_t = 50 \subset n_t = 100 \subset n_t = 200 \subset \hdots$, and so on, where { $n_t$ denotes the number of training samples}.
For any testing, except for the linear deconvolution problem in which we use $200$ testing samples,  a test data set of 500 samples is used to compare  approaches. 
The Tikhonov inverse solutions are obtained by the default BFGS algorithm \cite{nocedal1999numerical} in \texttt{Jax} \cite{jax2018github}.  
A summary of training parameters is presented in \cref{tab:Summary_training_params} in the supplementary.

{\bf Accuracy metric.}
To estimate the accuracy of each approach, we compute average relative errors from $M = \LRc{200,500}$ unseen random samples: the first based on pointwise values and the second on Euclidean norm of the inverse parameter vector as follows
\begin{equation}
\eqnlab{Errj}
\text{Err}_j = \frac{1}{M} \sum_{i=1}^{M} \frac{\LRp{\ub_{j}^{i,\text{pred}} - \ub_{j}^{\text{true}}}^2 }{\sfrac{\nor{\ub^{\text{true}}}^2}{m}} \times 100 \quad (\%),
\end{equation}
and 
\begin{equation}
\eqnlab{Err}
\text{Err} = \frac{1}{M} \sum_{i=1}^{M} \frac{\nor{\ub^{i,\text{pred}} - \ub^\text{true}}^2}{\nor{\ub^\text{true}}^2} = \frac{1}{m} \sum_{j=1}^{m} \text{Err}_j \quad (\%),
\end{equation}
where superscript $i$ denotes the $i$th sample, subscript $j$ denotes the $j$th component of the vector under consideration, and $m$ is the number of spatial grid points. Here, ``pred" stands for the solution predicted by the neural network, and ``true" for the synthetic ground truth  parameters.

\subsection{2D heat conductivity inverse problem}
\seclab{Heat_problem}

The heat equation we consider is the following
\begin{align*}
     -\nabla \cdot \LRp{e^u \nabla y} & = 20  \quad \text{in } \Omega = \LRp{0,1}^2\\
    y & = 0 \quad \text{ on } \Gamma^{\text{ext}} \\
    \textbf{n} \cdot \LRp{e^u \nabla y} & = 0 \quad \text{ on } \Gamma^{\text{root}},
\end{align*}
where the conductivity $u$ is the parameter of interest (PoI), $y$ is the temperature field, and $\textbf{n}$ is the unit outward normal vector on Neumann boundary part $\Gamma^{\text{root}}$ . \cref{fig:2D_Heat_Model} shows the domain (left subfigure) and a $16 \times 16$ mesh (right subfigure) together with the locations of 10 observational points of the state $y$. In this problem, we are interested in reconstructing the PoI field given a set of 10 pointwise observations.

{\bf Generating train and test data sets.}
We start with drawing the parameter conductivity samples via a truncated Karhunen-Lo\'eve expansion
\begin{equation*}
    u(x) = \sum_{i =1 }^n \sqrt{\lambda_i} \mb{\phi}_i(x) z_i, \quad x \in \LRs{0,1}^2,
\end{equation*}
where $\LRp{\lambda_i, \mb{\phi}_i}$ is the eigenpair of a 
two-point correlation function from \cite{constantine2016accelerating}, and $\textbf{z} = \LRp{z_i}_{i=1}^n \sim \mc{N}\LRp{0,\I}$ is a standard Gaussian random vector. 
We choose $n = 15$ eigenvectors corresponding to the first $15$ largest eigenvalues. For each sample, we discretize initial vorticity $u(x)$, denoted as $\ub$, and we solve the heat equation for the temperature $\yb$ by finite element method.
Observations are obtained by extracting values of the temperature field at $10$ observational points, which are then corrupted with additive Gaussian noise with a noise level of $\delta = 0.5\%$. A pair of conductivity field and its corresponding temperature distribution is depicted in the middle column of  \cref{fig:2D_Heat_20_database_predicted_test_samples}. We generate test pairs $\LRp{\ub, \yb}$ using the  same process.

\begin{figure}[h!t!b!]
     \centering
     \begin{subfigure}[b]{0.40\textwidth}
            \centering
            \includegraphics[width = 1.0\textwidth]{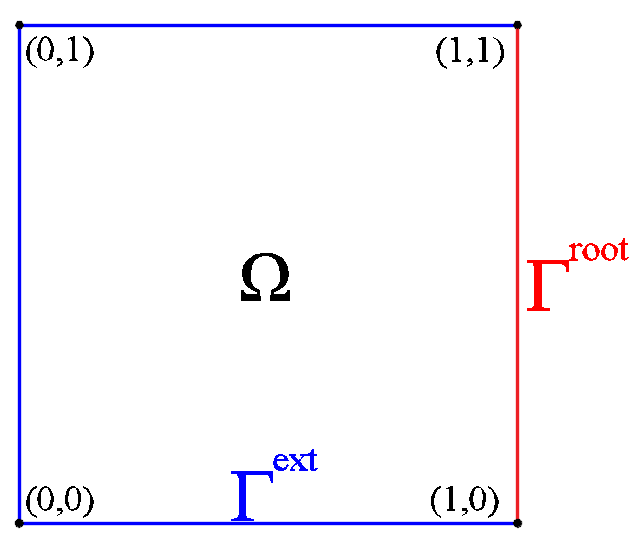}
    \end{subfigure}
    \begin{subfigure}[b]{0.43\textwidth}
            \centering
            \includegraphics[width =0.8\textwidth]{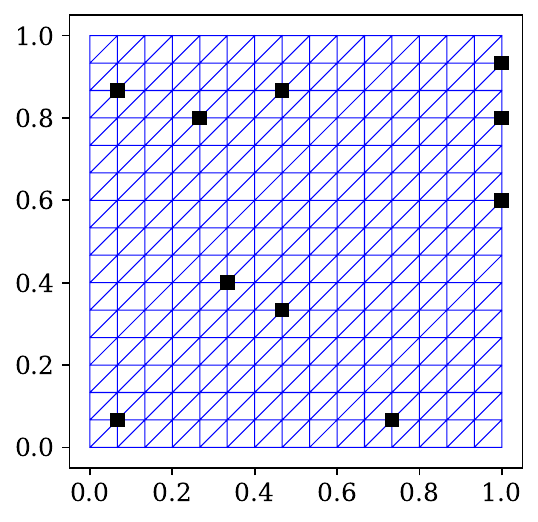}
     \end{subfigure}
    
    \caption{\textbf{2D heat conductivity inverse problem.} \textit{Left figure} the domain and the boundaries; \textit{Right figure} A $16 \times 16$ finite element mesh and $10$ observational locations.}
    \figlab{2D_Heat_Model} 
\end{figure}


Next, we consider two cases of train data for learning the inverse map from observations to conductivity. Case I: Full base, i.e., $n_t$ distinct training samples  are used; and Case II: we first pick a number of distinct baseline samples $n_b$ smaller than $n_t$, and then replicate and randomize them  to obtain $n_t$ samples for the train data set. For each case, the average relative error in \cref{eq:Err} is computed with 500 true test samples for  \nDNN{}, \mcDNN{} and \TNet{},  and is compared to the relative error of the Tikhonov regularization approach.

{\bf Case I: Training with full data sets $n_b = n_t \in \LRc{50, 100, 200}$.}
We train 
\nDNN{}, \mcDNN{} and \TNet{} networks
using three different full training data bases, $n_t = 50 \subset n_t = 100 \subset n_t = 200$ and present the smallest errors in \cref{tab:2D_Heat_full_data_base}. As can be seen, larger data sets provide more accurate inverse maps. In particular, the average smallest relative errors for \nDNN{} for these training sets are 60.41\%, 50.69\% and 49.07\%  which are higher than 57.27\%, 50.39\% and 48.40\%, respectively, for \mcDNN{}. With the smallest errors of 45.98\%, 45.35\% and, 44.98\%, correspondingly, \TNet{} outperforms \nDNN{} and \mcDNN{} by a significant margin, and is similar to Tikhonov (TIK) approach.
It is not surprising since \TNet{} approach is designed to learn Tikhonov method, as discussed in \cref{sect:TNET}. 
This is further confirmed by the fact that while regularization parameters for \mcDNN{}, \TNet{}, and Tikhonov approaches  are the same, namely $\alpha = 8000$, only \TNet{} and Tikhonov solutions agree well with each other for a wide range of regularization parameters, as shown in \cref{fig:2D_Heat_regularization}.
On the contrary, a data-driven approach such as \nDNN{} requires sufficient training data (more than 100 for this case as \cref{fig:2D_Heat_regularization} indicates) to provide a reasonable solution. We note that \mcDNN{} is not much more accurate than \nDNN{} for this example, perhaps due to strong bias from the data as suggested by \cref{coro:optDNN}.

The preceding discussion also alludes to an important point. In particular, identifying a good approximation of the optimal regularization parameter plays a vital part in \TNet{} performance. This  can be accomplished by finding a good regularization parameter for the Tikhonov approach and using it for \TNet{}. The subject of determining a suitable regularization parameter has been studied extensively in the literature using various approaches including the Morozov discrepancy principle, L-curve, and cross-validation \cite{mueller2012linear, tikhonov1995numerical, vogel2002computational}. The numerical results in \cref{fig:2D_Heat_regularization} show that  \TNet{} and \mcDNN{} results are robust in accuracy for a sufficiently large neighborhood around the optimal Tikhonov regularization parameter, and thus a reasonable regularization parameter is sufficient for \TNet{} and \mcDNN{} methods. Another important point that we show in the deconvolution \cref{sect:1D_Linear} is that the optimal regularization parameter for \TNet{} and \mcDNN{} are numerically independent of training data sets, while it varies drastically for \nDNN{} method. This implies \TNet{} and \mcDNN{} are more robust and reliable than \nDNN{}.

\begin{table}[htb!]
\centering
\caption{\textbf{2D heat conductivity inverse problem, Case I.} The average relative error \cref{eq:Err} over 500 test samples obtained by \nDNN{} (optimal $\alpha$ varies depending on the data set), \mcDNN{} ($\alpha = 8000$), \TNet{} ($\alpha = 8000$) with nested data sets $n_t = 50 \subset n_t = 100 \subset n_t = 200$, and Tikhonov (TIK) with $\alpha = 8000$.}
\tablab{2D_Heat_full_data_base}
\begin{tabular}{|c|c|c|c|c|}
\hline
     & \nDNN{}  & \mcDNN{} & \TNet{}  & TIK                    \\ \hline
$n_t =$ 50  & 60.41 & 57.27 & 45.98 & \multirow{3}{*}{44.99} \\ \cline{1-4}
$n_t =$ 100 & 50.69 & 50.39 & 45.35 &                        \\ \cline{1-4}
$n_t =$ 200 & 49.07 & 48.40 & 44.98 &                        \\ \hline
\end{tabular}
\end{table}

\begin{figure}[h!t!b!]
    \centering
    \includegraphics[width = 0.95\textwidth,clip]{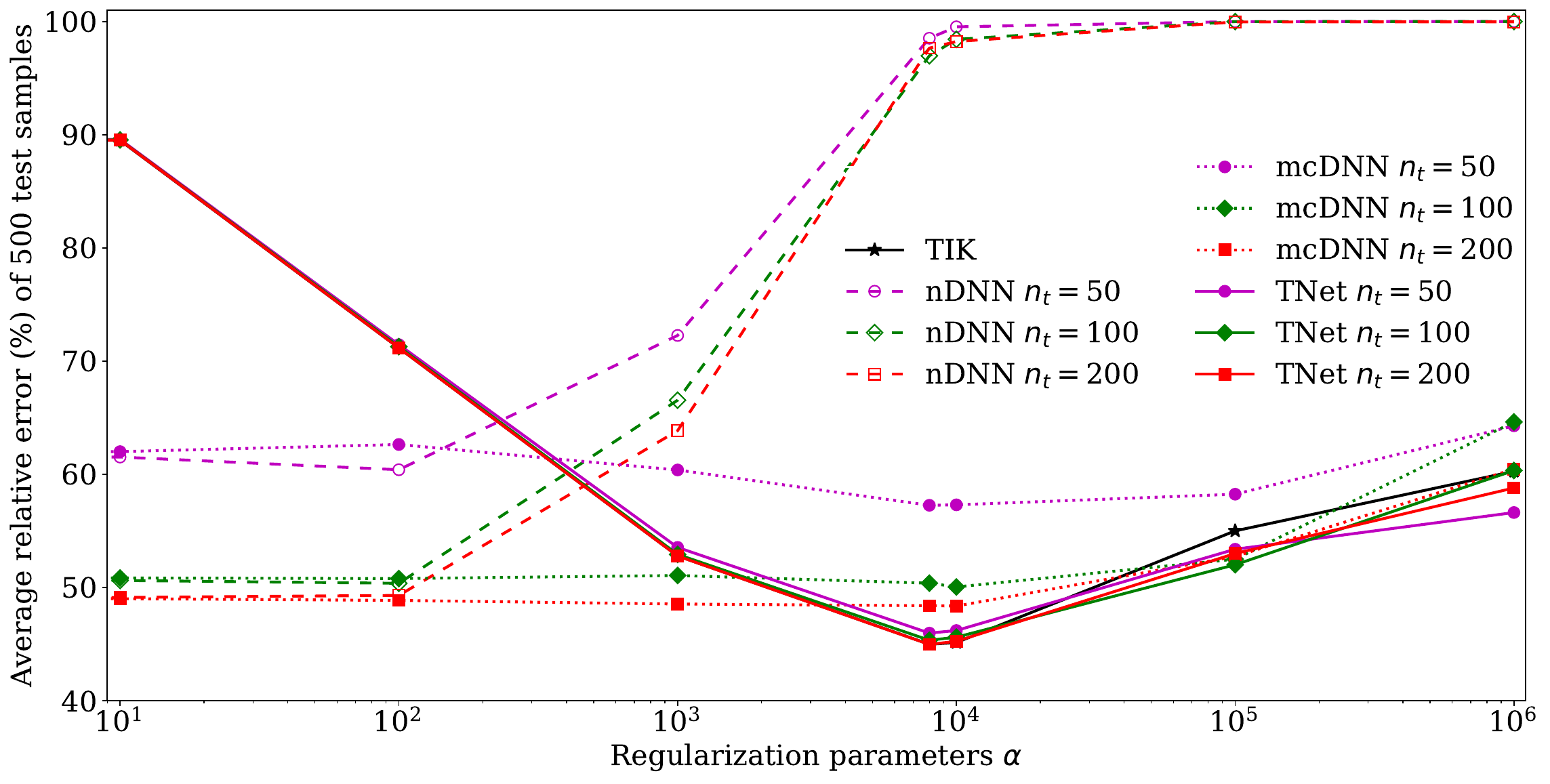}
    \caption{\textbf{2D heat conductivity inverse problem, Case I.} The average relative error \cref{eq:Err} over 500 test samples with nested data sets $n_t = 50 \subset n_t = 100 \subset n_t = 200$. The comparisons are done for  \nDNN{} (dashed curves), \mcDNN{} (dotted curves), \TNet{} (colored solids curves), and Tikhonov (TIK: black curve) over a wide range of regularization parameter values.}
    \figlab{2D_Heat_regularization}
\end{figure}

{\bf Case II: Training with $n_b = 20 < n_t \in \LRc{60, 100, 200, 1000, 2000, 5000}$.}
We now investigate how the data augmentation via randomization  performs with \nDNN{}, \mcDNN{} and \TNet{}. In particular, 20 noise-free baseline data pairs are replicated to create $N$ samples of training data sets ranging from $n_t = 20$ to $n_t = 5000$, which are then randomized with 2\% additive white noise. \cref{tab:2D_Heat_20_pair_data_base} shows the average relative error \cref{eq:Err} of the test data set obtained by \nDNN{}, \mcDNN{} and \TNet{}. 
In the first row  are the results for the baseline case with $n_t = n_b = 20$ and this is used as the reference for the other rows. 
It can be seen that data randomization and augmentation, though regularizes the smoothness of the network, negligibly improves the accuracy of \nDNN{}. 
Clearly,  \nDNN{} is not equipped with the forward map and completely depends on the limited information given in the baseline data. 
On the contrary, the accuracy for \mcDNN{} is improved by about $10\%$ for $n_t \ge 1000$. This is expected as  \cref{theo:randTNet} shows that randomization, via the model-constrained term, promotes the network solution to be the right inverse of the forward map up to second order.
However, \mcDNN{}'s accuracy level saturates with $n_t = 1000$ and  is still significantly higher than the Tikhonov approach (the last column). This is again due to data-dependent regularization nature (see \cref{coro:optDNN}), and hence biasing to the training data, of the \mcDNN{} approach despite of the effectiveness of model-constrained term. 
Unlike \nDNN{} and \mcDNN{} approaches, \TNet{} results are much more accurate regardless of any considered value of $n_t$. Furthermore, they seem to approach the Tikhonov accuracy as  $n_t$ increases from $20$ to $5000$.
 In particular, \TNet{} needs only about 100 samples replicated and randomized from  $n_b = 20$ distinct baseline samples to learn an inverse map as nearly accurate as the Tikhonov solution. 
 
 \cref{fig:2D_Heat_20_database_average_error} shows
 the pointwise average error over 500 test samples (see \cref{eq:Errj})
 for \nDNN{}, \mcDNN{}, \TNet{}, and Tikhonov (TIK) approaches for $n_b = 20$ and $n_t = 200$. While \nDNN{} and \mcDNN{} have a high level of error, \TNet{} has a similar  error as the TIK solver in both values and patterns. 
For all these methods,  we show in \cref{fig:2D_Heat_20_database_predicted_test_samples}  the
 reconstructed conductivities from a new unseen noisy data for $n_b = 20$ and $n_t = 200$. The synthetic ground truth conductivity and the corresponding temperature distribution are also presented for reference in the middle column. Again, the \TNet{} inverse solution is in good agreement with the Tikhonov one, and thus with the ground truth, while \nDNN{} and \mcDNN{} yield quite inaccurate reconstructions.

\begin{table}[htb!]
\centering
\caption{\textbf{2D heat conductivity inverse problem, Case II.} The average relative error \cref{eq:Err} for \nDNN{} (optimal $\alpha$ varies depending on the data set), \mcDNN{} ($\alpha = 8000$), \TNet{}($\alpha = 8000$), and Tikhonov (TIK) ($\alpha = 8000$)  over 500-sample test data set obtained by training with $n_b = 20$ baseline data pairs.} 
\tablab{2D_Heat_20_pair_data_base}
\begin{tabular}{|l|c|c|c|c|}
\hline
      & \nDNN{}  & \mcDNN{} & \TNet{}  & TIK                    \\ \hline
$n_t = $ 20   & 89.66 & 77.61 & 55.56 & \multirow{7}{*}{44.99} \\ \cline{1-4}
$n_t = $ 60   & 86.87 & 77.02 & 47.35 &                        \\ \cline{1-4}
$n_t = $ 100  & 87.19 & 72.58 & 46.27 &                        \\ \cline{1-4}
$n_t = $ 200  & 89.59 & 71.81 & 46.01 &                        \\ \cline{1-4}
$n_t = $ 1000 & 88.67 & 69.68 & 45.16 &                        \\ \cline{1-4}
$n_t = $ 2000 & 88.31 & 69.72 & 45.23 &                        \\ \cline{1-4}
$n_t = $ 5000 & 88.55 & 69.71 & 45.11 &                        \\ \hline
\end{tabular}
\end{table}

\begin{figure}[htb!]
    \centering
    \begin{tabular*}{\textwidth}{c c c c}
        \centering
        \raisebox{-0.5\height}{\nDNN{} (80.59)} &
        \raisebox{-0.5\height}{\mcDNN{} (71.81)} &
        \raisebox{-0.5\height}{\TNet{} (46.01)} &
        \raisebox{-0.5\height}{TIK (44.99)}
        \\
        \raisebox{-0.5\height}{\includegraphics[width=0.22\textwidth]{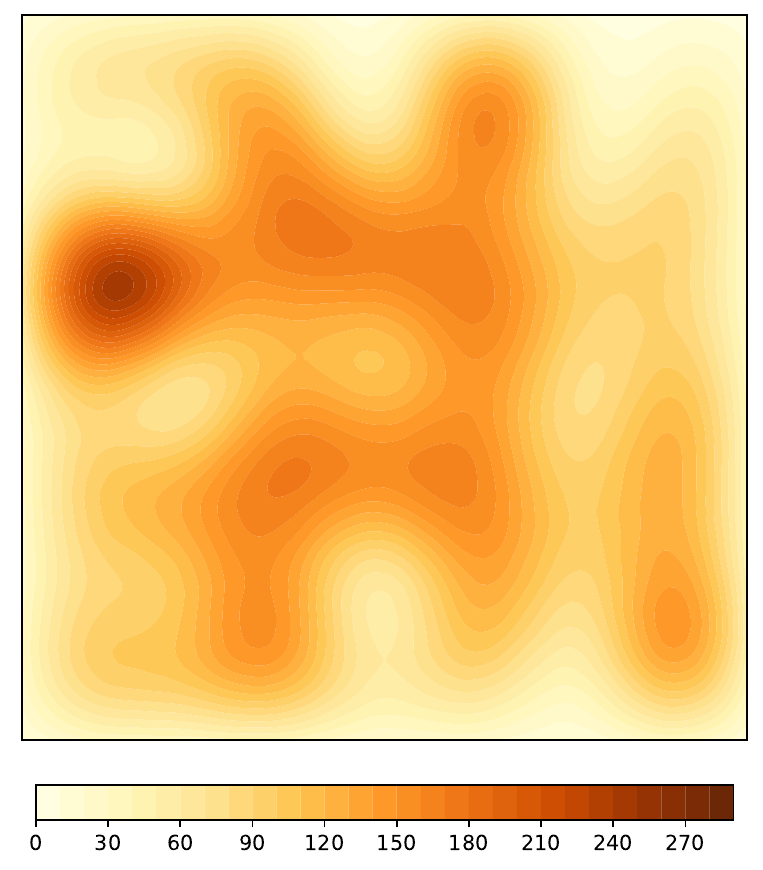}} &
        \raisebox{-0.5\height}{\includegraphics[width=0.22\textwidth]{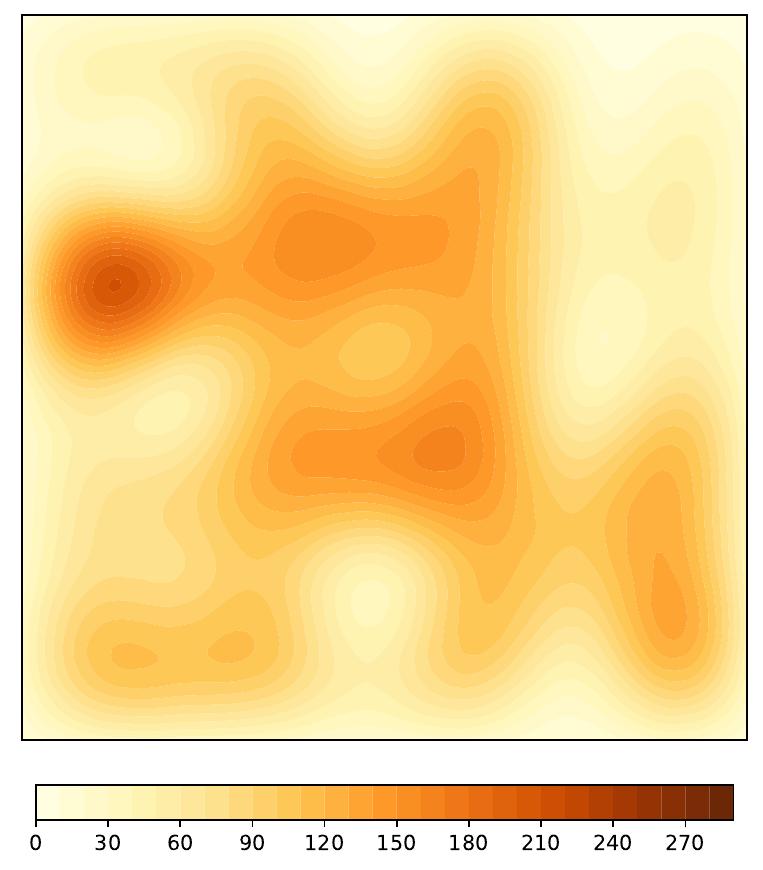}} &
        \raisebox{-0.5\height}{\includegraphics[width=0.22\textwidth]{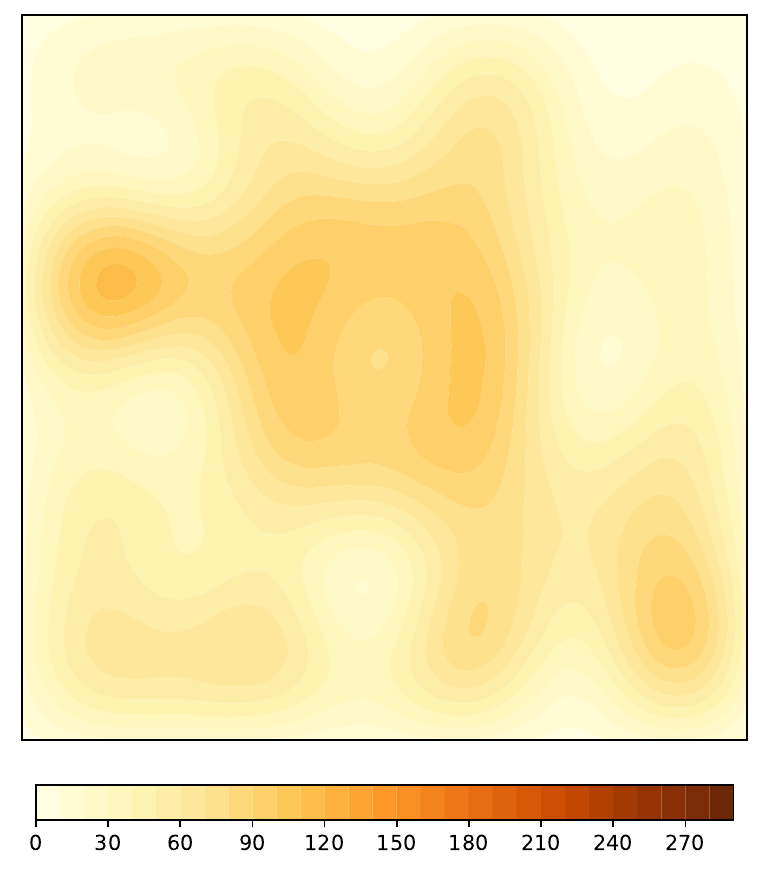}} &
        \raisebox{-0.5\height}{\includegraphics[width=0.22\textwidth]{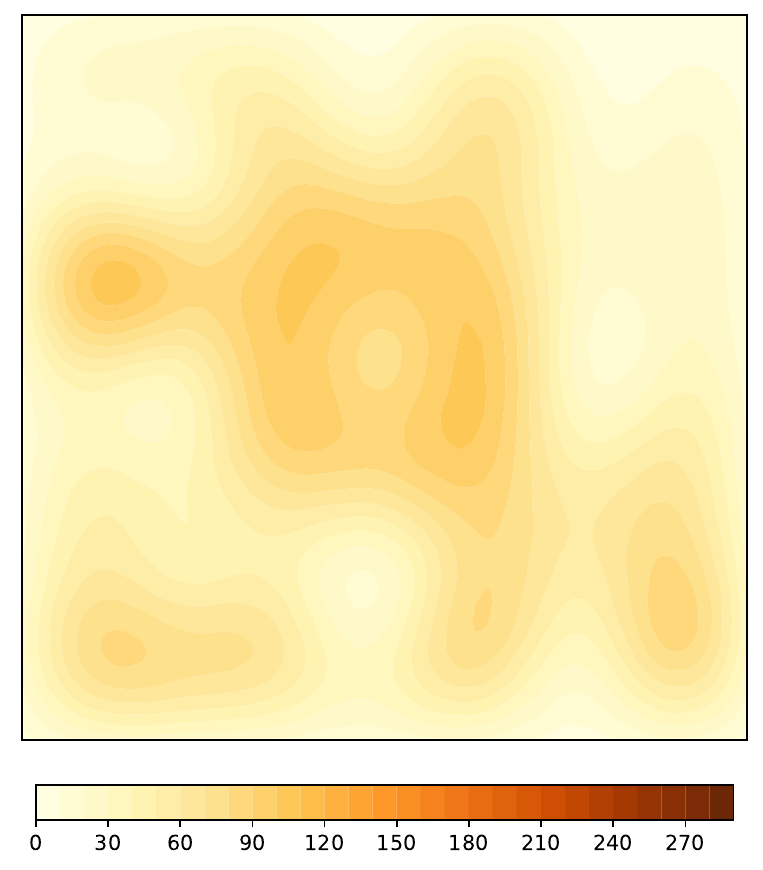}}

    \end{tabular*}
    \captionof{figure}{\textbf{2D heat conductivity inverse problem, Case II.} The distribution of average relative pointwise  error \cref{eq:Errj}  for \nDNN{}, \mcDNN{}, \TNet{}, and Tikhonov (TIK) over 500 test samples obtained with $n_b = 20$ and $n_t = 200$. The numbers in the parentheses are the average error \cref{eq:Err} incurred by these methods.}
    \figlab{2D_Heat_20_database_average_error}
\end{figure}

\begin{figure}[htb!]
    \centering
    \begin{tabular*}{\textwidth}{c c c}
        \centering
        \raisebox{-0.5\height}{\nDNN{}$\quad$} &
        \raisebox{-0.5\height}{Exact$\quad$} &
        \raisebox{-0.5\height}{\mcDNN{}$\quad$}
        \\
        \raisebox{-0.5\height}{\includegraphics[width=0.30\textwidth]{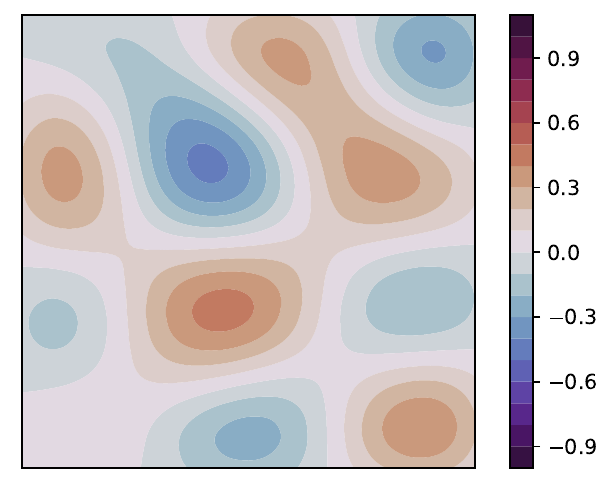}} &
        \raisebox{-0.5\height}{\includegraphics[width=0.30\textwidth]{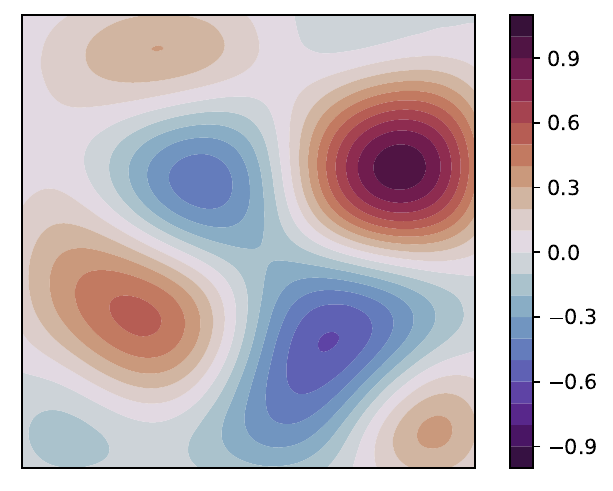}} &
        \raisebox{-0.5\height}{\includegraphics[width=0.30\textwidth]{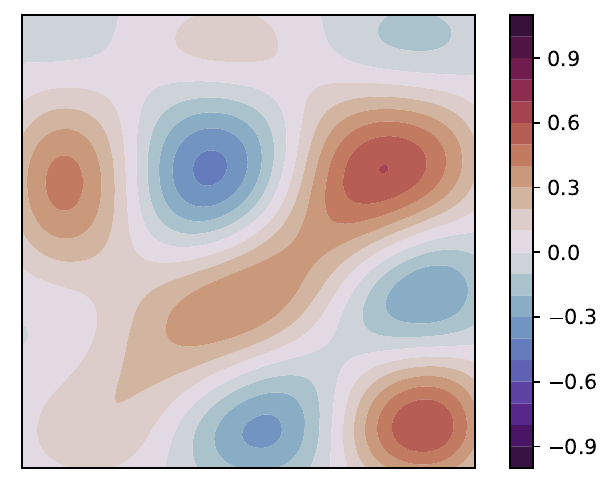}}
        \\ 
        \raisebox{-0.5\height}{\TNet{}$\quad$} &
        \raisebox{-0.5\height}{Temperature field \quad} &
        \raisebox{-0.5\height}{TIK$\quad$}
        \\
        \raisebox{-0.5\height}{\includegraphics[width=0.30\textwidth]{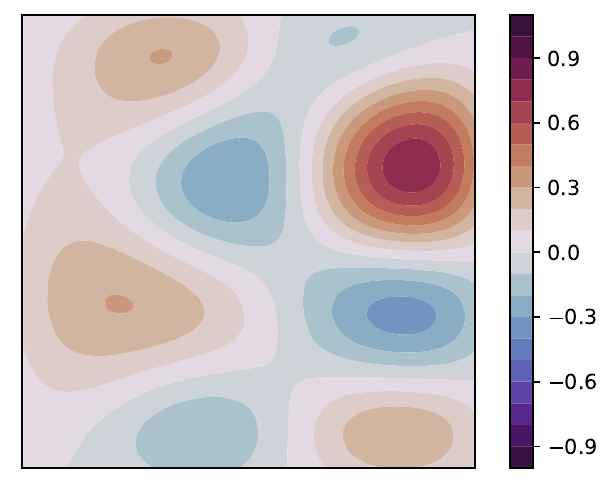}} &
        \raisebox{-0.5\height}{\includegraphics[width=0.3\textwidth]{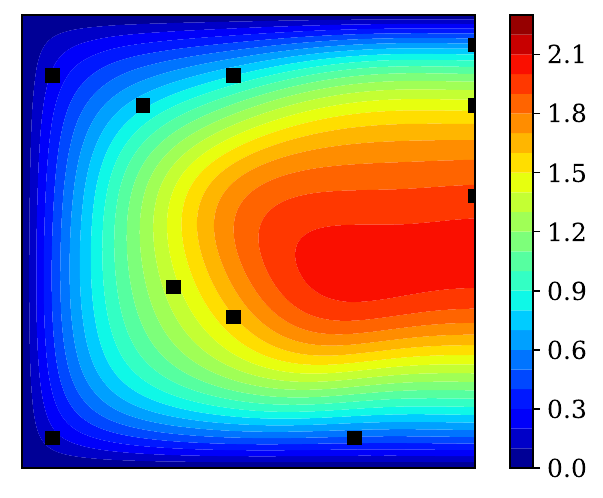}} &
        \raisebox{-0.5\height}{\includegraphics[width=0.30\textwidth]{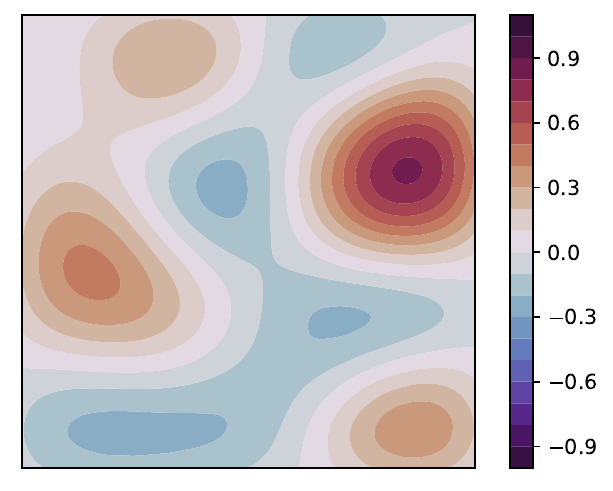}}
    \end{tabular*}
    \captionof{figure}{\textbf{2D heat conductivity inverse problem, Case II.} Predicted (reconstructed) heat conductivities for an unseen noisy test sample obtained by \nDNN{}, \mcDNN{}, and  \TNet{} neural networks along with the Tikhonov reconstruction for $n_b = 20$ and $n_t = 200$.  Shown in the middle column are the synthetic ground truth (Exact) conductivity and the corresponding temperature field for reference.}
    \figlab{2D_Heat_20_database_predicted_test_samples}
\end{figure}

{\bf How many  baseline pairs are sufficient for \TNet{}?} For this problem
we numerically study how many distinct baseline pairs are needed to achieve a reasonably accurate inverse solution  from the \TNet{} approach. \cref{fig:2D_Heat_data_base_convergence} shows that $n_b = 5$ baseline pairs are sufficient when $n_t \ge 1000$. For example, with $n_b = 5$ and $n_t = 1000$, \TNet{} achieves a relative error of approximately 46.7\% compared to 44.99 \% of the Tikhonov solution. We also observe that, given an inadequate number of distinct baseline pairs, i.e., one or two, it is challenging to learn a highly accurate inverse operator even with data augmentation, randomization, and large $n_t$ due to lacking information. This can be seen through the second term in \cref{eq:TNET_data_aug_loss_final}. Indeed, in this case, we have
\[
 \half \sum_{i=1}^{n_t}\lambda_i^2 \LRp{\mc{P}_1^i + \mc{P}_2^i + \mc{P}_3^i  + \mc{P}_4^i} = \frac{n_t}{2n_b} \sum_{i=1}^{n_b}\lambda_i^2 \LRp{\mc{P}_1^i + \mc{P}_2^i + \mc{P}_3^i  + \mc{P}_4^i},
\]
and thus the induced regularizations are active only at the distinct baseline samples. For small baseline samples, there is simply not enough information for \TNet{} to perform well. This again agrees with the Hermite interpolation analogy discussed in \cref{sect:TNET}.

\begin{figure}[h!t!b!]
    \centering
    \includegraphics[width = 0.95\textwidth,clip]{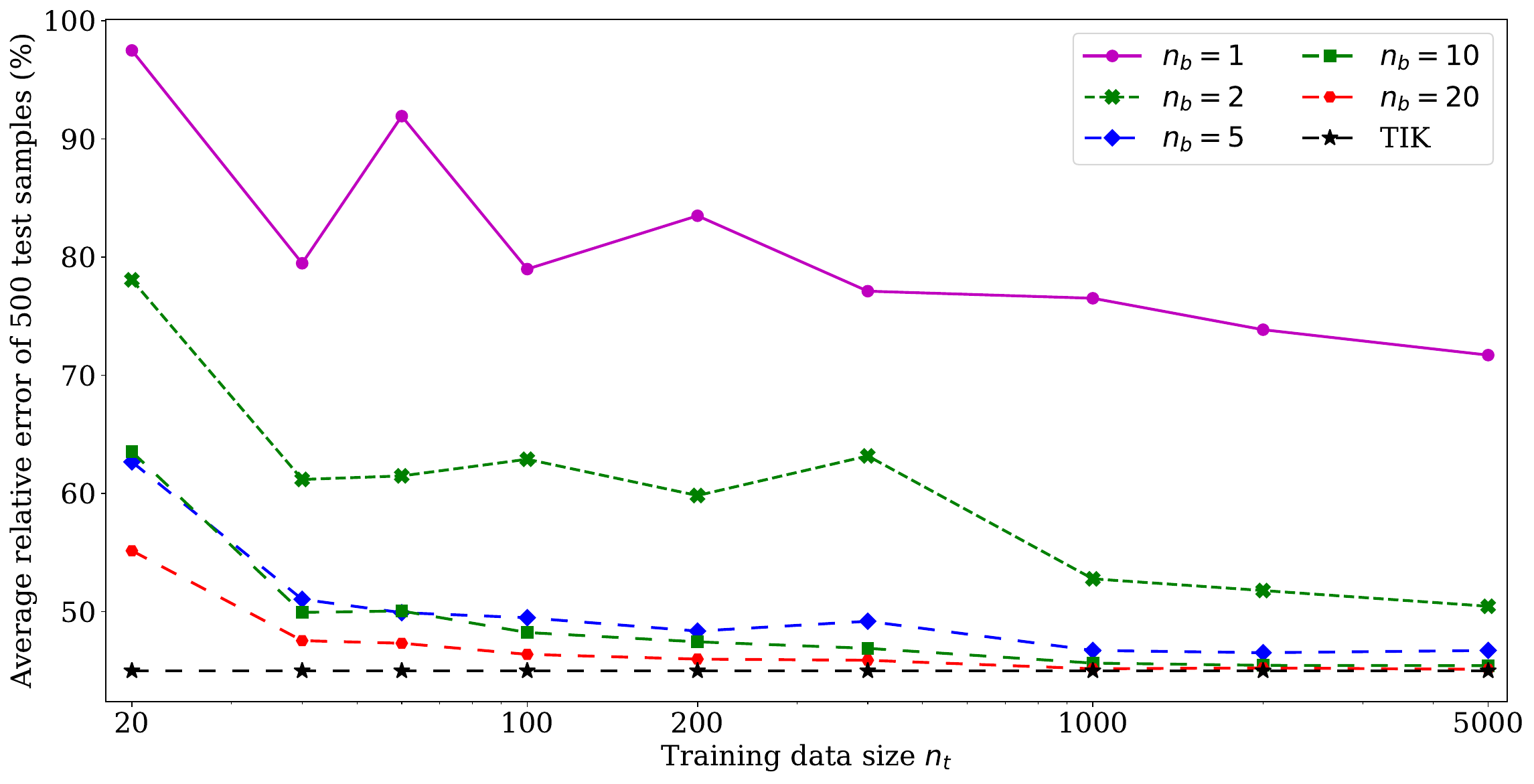}
    \caption{\textbf{2D heat conductivity inverse problem.} 
    The average relative error \cref{eq:Err} as a function of $n_b$ and $n_t$
for \TNet{} approach.}
    \figlab{2D_Heat_data_base_convergence}
\end{figure}

\subsection{2D Navier-Stokes equation}
\seclab{2D_NS}

The vorticity form of 2D Navier-Stokes equation for viscous and incompressible fluid \cite{FouierOP} is written as
\begin{equation*}
    \begin{aligned}
        \partial_t u(x,t) + v(x,t) \cdot \nabla u(x,t) &= \nu \Delta u(x,t) + f(x), & \quad x \in \LRp{0,1}^2, t \in (0, T] \\
        \nabla \cdot v(x,t) & = 0, & \quad x \in \LRp{0,1}^2, t \in (0, T] \\
        u(x,0) & = u_0(x),  & \quad x \in \LRp{0,1}^2 
    \end{aligned}
\end{equation*}
where $v \in \LRp{0,1}^2 \times (0, T]$ is the velocity field, $u = \nabla \times v$ is the vorticity, $u_0$
is the initial vorticity, $f(x) = 0.1 \LRp{\sin \LRp{2 \pi \LRp{x_1 + x_2}} + \cos\LRp{2 \pi \LRp{x_1 + x_2}}}$ is the forcing function, and $\nu = 10^{-3}$ is the viscosity coefficient. The spatial domain is discretized with $32 \times 32$ uniform mesh, while the time horizon $t \in \LRp{0, 10}$ is subdivided into 1000  time steps with $\Delta t = 10^{-2}$. We target to reconstruct the initial vorticity $u_0$ from the measurements of vorticity at $20$ observed points at the final time $T = 10$. 

{\bf Generating train and test data sets.}
A data pair of $\LRp{\ub, \yb}$ is generated by a similar procedure outlined for Burgers' equation  in Section \cref{sect:Burger}.
In particular, we draw samples of $u(x,0)$ using the truncated Karhunen-Loève expansion
\[
u(x,0) = \sum_{i=1}^{15} \sqrt{\lambda_i} \, {\bf \omega_i}(x) \, z_i,
\]
where $z_i \sim \mathcal{N} \LRp{0, 1}, i = 1, \hdots, 15$, and $\LRp{\lambda_i, {\bf \omega}_i}$ are eigenpairs obtained by the eigendecomposition of the covariance operator $ 7^{\frac{3}{2}} \LRp{-\Delta + 49 \textbf{I}}^{-2.5}$ with periodic boundary conditions. Next, we discretize an initial vorticity $u(x,0)$, denoted as $\ub_0$, and  we solve the Navier-Stokes equation by the stream-function formulation with a pseudospectral method \cite{FouierOP} to obtain a discrete representation $\ub_t$ of $u\LRp{x,t}$ at any time $t$. 
The observation operator is imposed on solution $\ub_{10}$ to form the synthetic observables $\yb$, then a realization of additive white noise 
with 
$\delta = 2\%$ is added to generate a noise-corrupted $\yb$ sample. A sample of $\LRp{\ub_0, \ub_{10}}$ pair together with the observation points is shown in the middle column of  \cref{fig:2D_NS_50_database_predicted_test_samples}.


Similar to the heat conductivity inverse problem in \cref{sect:Heat_problem} and Burgers' equations in \cref{sect:Burger}, we consider two cases of training data.
{\em Case I}: Full data with 
distinct training samples are used; and {\em Case II}: we first pick a number of distinct baseline samples $n_b$ smaller than $n_t$, and then replicate and randomize them  to obtain $n_t$ samples for the train data set. We shall compare and contrast results from \nDNN{}, \mcDNN{}, \TNet{}, and  Tikhonov solutions.

{\bf Case I:
Full distinct training samples $n_b = n_t =\LRc{50, 100, 200, 500}$.}
In \cref{fig:2D_NS_regularization} are 
the average relative error \cref{eq:Err} versus the regularization parameter $\alpha$ over 500 test samples with $\n_b = n_t = \LRc{50, 100, 200, 500}$. The results are shown for \nDNN{} (dashed curves), \mcDNN{} (dotted curves), \TNet{} (colored solids curves), and Tikhonov (TIK: black curve) solutions.
The general behavior of the error as a function of regularization parameter is similar to the results for Burgers and heat equations, and thus omitted.
 Here, {\em we focus on results at the ``best" regularization parameters for all methods.}
The optimal regularization parameters of \mcDNN{} and \TNet{} agree with that of Tikhonov methods, namely $\alpha = 2200$, due to the same reason as explained in the three other numerical problems. Whereas, \nDNN{} has small optimal regularization parameters. At the optimal regularization parameter, as summarized in \cref{tab:NS_full_daata}, with a given $n_t$, \TNet{} error is smaller than those of \nDNN{} and \mcDNN{} with twice amount of data. For example, \TNet{} solution with $n_t = 100$ incurs an error of $27.14\%$, smaller than $32.39\%$ and $28.54\%$ of \nDNN{}   and \mcDNN{} with $n_t = 200$. It is not surprising that \TNet{} solution tends to converge to Tikhonov (TIK) solution faster as it is designed to do so (see \cref{coro:TNETDNN}) while the others are not. 
Clearly, without being constrained to the forward map \nDNN{}  needs the largest amount of data to approximate the inverse map with the same level of accuracy.

\begin{figure}[h!t!b!]
    \centering
    \includegraphics[width = 0.95\textwidth,clip]{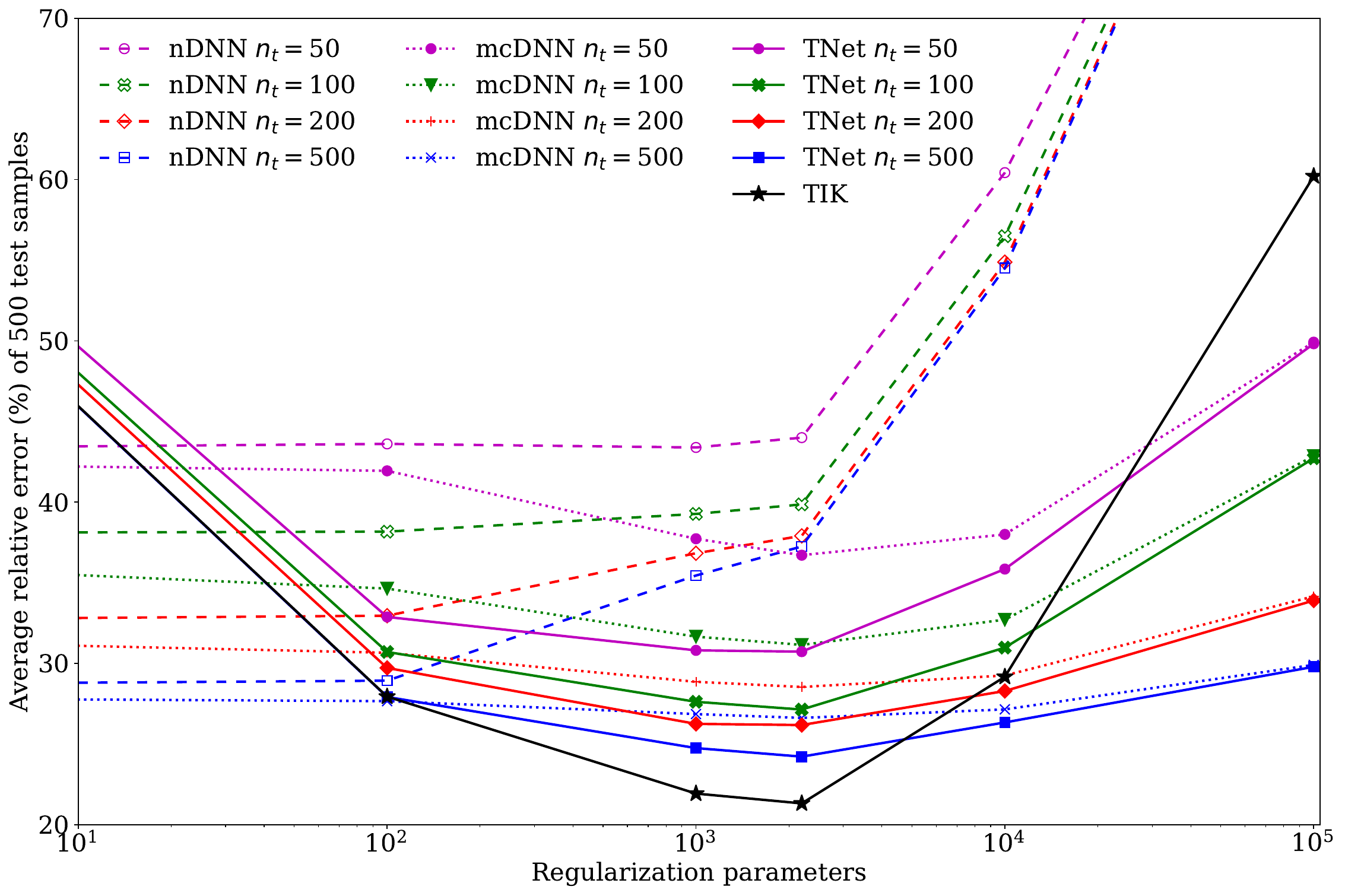}
    \caption{\textbf{2D Navier-Stokes equation, Case I.} The average relative error \cref{eq:Err} versus the regularization parameter $\alpha$ over 500 test samples with $\n_b = n_t = \LRc{50, 100, 200, 500}$. The results are shown for \nDNN{} (dashed curves), \mcDNN{} (dotted curves), \TNet{} (colored solids curves), and Tikhonov (TIK: black curve) solutions.}
    \figlab{2D_NS_regularization}
\end{figure}

\begin{table}[h!t!b!]
\caption{\textbf{2D Navier-Stokes equation, Case I.} The average relative error  \cref{eq:Err} over 500 test samples obtained by \nDNN{} ($\alpha \approx 10$), \mcDNN{} ($\alpha = 2200$), \TNet{} ($\alpha = 2200$) with nested data sets $n_t = 50 \subset n_t = 100 \subset n_t = 200 \subset n_t = 500$, and TIK ($\alpha = 2200$.)
}
\tablab{NS_full_daata}
\centering
\begin{tabular}{|c|c|c|c|c|}
\hline
      & \nDNN{}  & \mcDNN{} & \TNet{}  & TIK                    \\ \hline
$n_t =$ 50  & 43.00 & 37.86 & 30.71 & \multirow{4}{*}{21.93} \\ \cline{1-4}
$n_t =$ 100  & 37.98 & 31.15 & 27.14 &                        \\ \cline{1-4}
$n_t =$ 200 & 32.39 & 28.54 & 26.18 &                        \\ \cline{1-4}
$n_t =$ 500 & 28.08 & 26.62 & 24.22 &                        \\ \hline
\end{tabular}
\end{table}

{\bf Case II: Training with $n_b \in \LRc{10, 50} \le n_t \in \LRc{50, 100, 200, 500, 1000}$.}
\cref{tab:2D_NS_n_data_base} presents the relative error \cref{eq:Err}  of test data sets obtained by different approaches at the optimal regularization parameters. It can be seen that \nDNN{} results are improved as more distinct baseline data pairs are deployed in  training data sets. Nevertheless, for any  baseline case, replication and randomization to generate $n_t$ data samples, while being more computationally demanding, do not provide additional accuracy in \nDNN{} solutions. Again, this implies that the performance of \nDNN{} completely relies on the underlying information provided by distinct baseline data. The behavior of \mcDNN{} is, on the other hand, not predictable. In particular, 
 replication and randomization improves the results for $n_b = 10$ but not for $n_b = 50$: perhaps due to the data-driven nature, and hence bias, of $\ub_0^{\mcDNN{}}$ as discussed in \cref{sect:TNET} after \cref{coro:TNETDNN}.

On the contrary, \TNet{} results are consistently and significantly improved with replication and randomization for $n_b \in \LRc{10,50}$.  In particular, for a given $n_b$,  the larger $n_t$ is, the more accurate the \TNet{} solution. 
As further demonstrated in \cref{fig:2D_NS_50_database_average_error}, trained with $n_b = 50$ and $n_t = 1000$, the distribution of pointwise relative error for \TNet{} is closest to that of Tikhonov solution, and  is every where far lower than those of \mcDNN{} and \nDNN{}. 
Shown in \cref{fig:2D_NS_50_database_predicted_test_samples} are the predicted (reconstructed) initial vorticities for an unseen noisy test sample obtained by \nDNN{}, \mcDNN{}, and  \TNet{} neural networks for $n_b = 50$ and $n_t = 1000$ along with the Tikhonov reconstruction.  Shown in the middle column are the synthetic ground truth initial vorticity and the corresponding final vorticity for reference.
Again, \TNet{} reconstruction is closest to TIK inversion 
and \nDNN{} provides the most inaccurate solution.
 
\begin{table}[htb!]
\centering
\caption{\textbf{2D Navier-Stokes equation, Case II.} 
The average relative error \cref{eq:Err} for \nDNN{} ($\alpha\approx 10$), \mcDNN{} ($\alpha = 2200$), \TNet{}($\alpha = 2200$), and Tikhonov (TIK) ($\alpha = 2200$)  over 500-sample test data set obtained with $n_b = \LRc{10, 50}$ baseline data pairs.
}
\tablab{2D_NS_n_data_base}
\begin{tabular}{|c|ccc|ccc|c|}
\hline
\multirow{2}{*}{\begin{tabular}[c]{@{}c@{}}Training data\\ size ($n_t$)\end{tabular}} & \multicolumn{3}{c|}{$n_b = 10$}                               & \multicolumn{3}{c|}{$n_b = 50$}                               & \multirow{2}{*}{TIK}   \\ \cline{2-7}
& \multicolumn{1}{c|}{\nDNN{}}  & \multicolumn{1}{c|}{\mcDNN{}} & \TNet{}  & \multicolumn{1}{c|}{\nDNN{}}  & \multicolumn{1}{c|}{\mcDNN{}} & \TNet{}  &  \\ \hline
$n_t =$ 50 & \multicolumn{1}{c|}{85.09} & \multicolumn{1}{c|}{73.15} & 58.71 & \multicolumn{1}{c|}{43.00} & \multicolumn{1}{c|}{37.86} & 30.71 & \multirow{5}{*}{21.93} \\ \cline{1-7}
$n_t =$ 100 & \multicolumn{1}{c|}{85.37} & \multicolumn{1}{c|}{72.58} & 48.46 & \multicolumn{1}{c|}{42.03} & \multicolumn{1}{c|}{38.80} & 30.06 &                        \\ \cline{1-7}
$n_t =$ 200 & \multicolumn{1}{c|}{85.87} & \multicolumn{1}{c|}{64.04} & 39.75 & \multicolumn{1}{c|}{43.17} & \multicolumn{1}{c|}{37.74} & 29.06 &                        \\ \cline{1-7}
$n_t =$ 500    & \multicolumn{1}{c|}{85.85} & \multicolumn{1}{c|}{50.76} & 38.28 & \multicolumn{1}{c|}{43.29} & \multicolumn{1}{c|}{37.10} & 27.91 &                        \\ \cline{1-7}
$n_t =$ 1000 & \multicolumn{1}{c|}{85.52} & \multicolumn{1}{c|}{48.32} & 34.13 & \multicolumn{1}{c|}{42.85} & \multicolumn{1}{c|}{37.24} & 26.96 &                        \\ \hline
\end{tabular}
\end{table}

\begin{figure}[htb!]
    \centering
    \begin{tabular*}{\textwidth}{c c c c}
        \centering
        \raisebox{-0.5\height}{\nDNN{} (42.85)} &
        \raisebox{-0.5\height}{\mcDNN{} (37.24)} & 
        \raisebox{-0.5\height}{\TNet{} (26.96)} &
        \raisebox{-0.5\height}{TIK (21.93)}
        \\
        \raisebox{-0.5\height}{\includegraphics[width=0.22\textwidth]{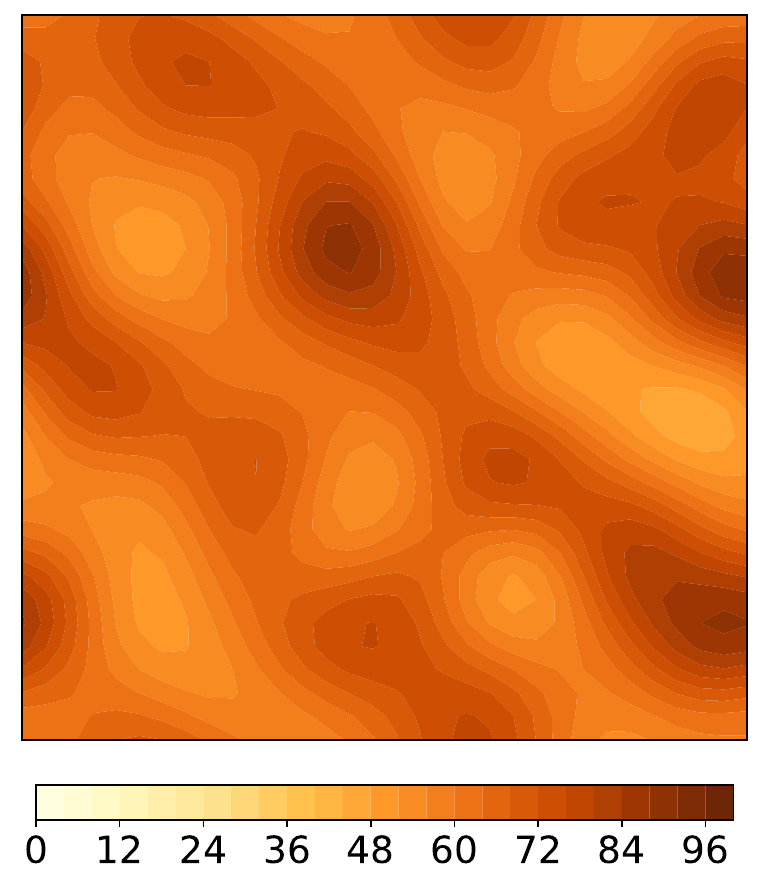}} &
        \raisebox{-0.5\height}{\includegraphics[width=0.22\textwidth]{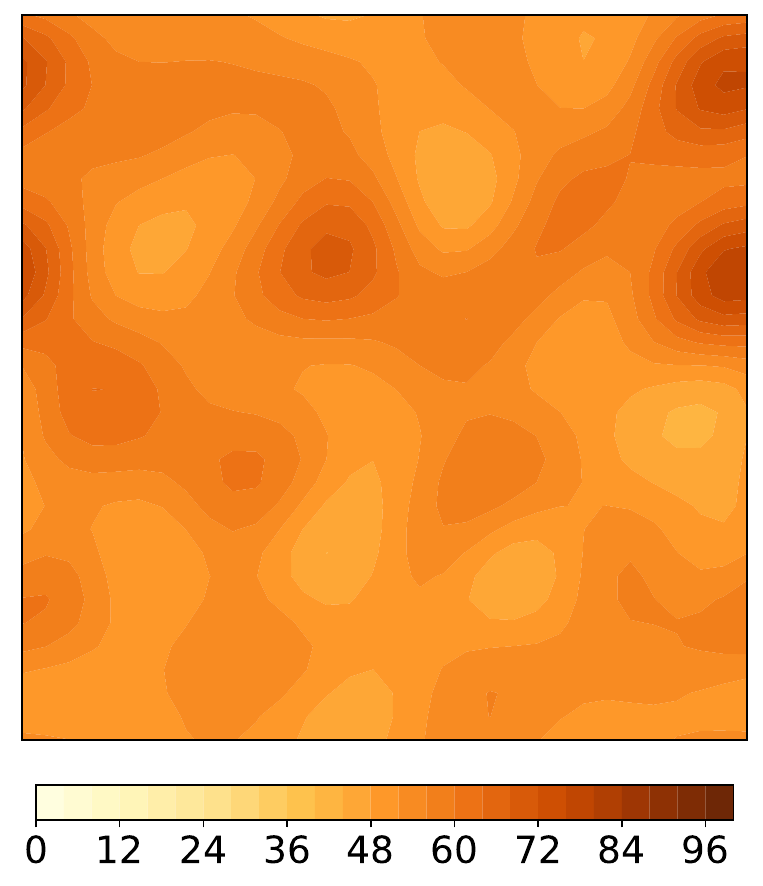}} & 
        \raisebox{-0.5\height}{\includegraphics[width=0.22\textwidth]{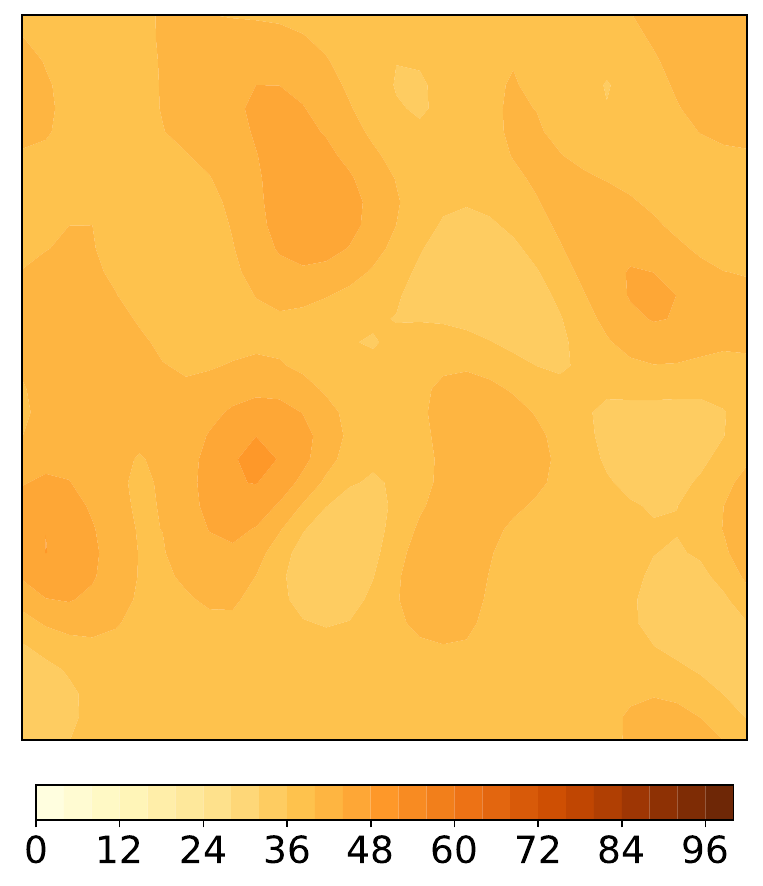}} &
        \raisebox{-0.5\height}{\includegraphics[width=0.22\textwidth]{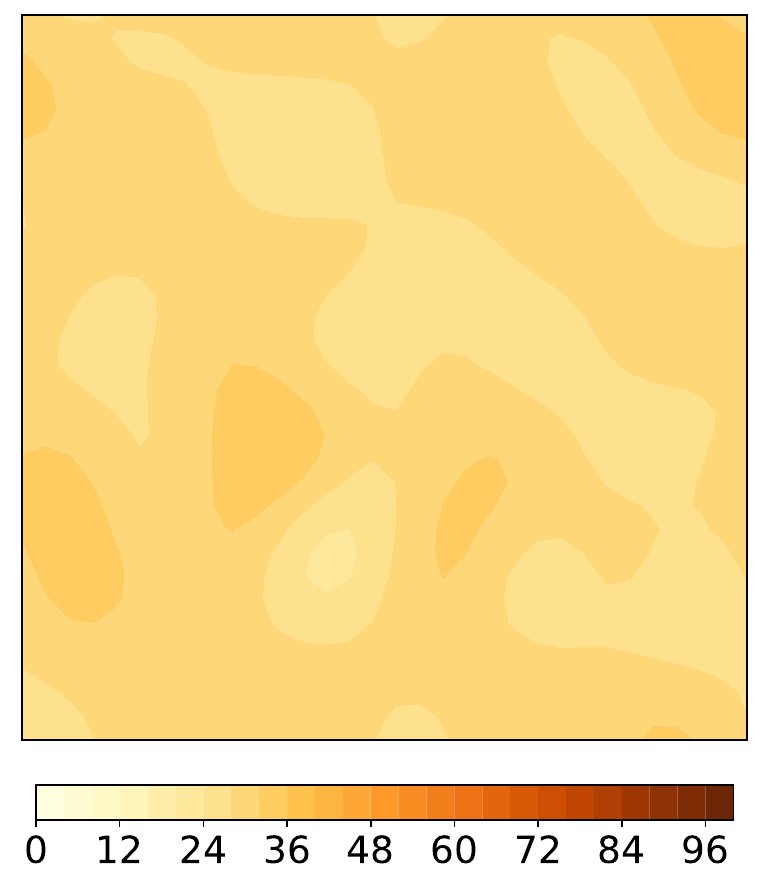}}
        
    \end{tabular*}
    \captionof{figure}{\textbf{2D Navier-Stokes equation, Case II.} 
    The distribution of average relative pointwise  error \cref{eq:Errj}  for \nDNN{}, \mcDNN{}, \TNet{}, and Tikhonov (TIK) over 500 test samples obtained with $n_b = 50$ and $n_t = 1000$. The numbers in the parentheses are the average error \cref{eq:Err} incurred by these methods.
    }
    \figlab{2D_NS_50_database_average_error}
\end{figure}

\begin{figure}[htb!]
    \centering
    \begin{tabular*}{\textwidth}{c c c}
        \centering
        \raisebox{-0.5\height}{\nDNN{}$\quad\quad$} &
        \raisebox{-0.5\height}{Exact$\quad\quad$} &
        \raisebox{-0.5\height}{\mcDNN{}$\quad\quad$}
        \\
        \raisebox{-0.5\height}{\includegraphics[width=0.30\textwidth]{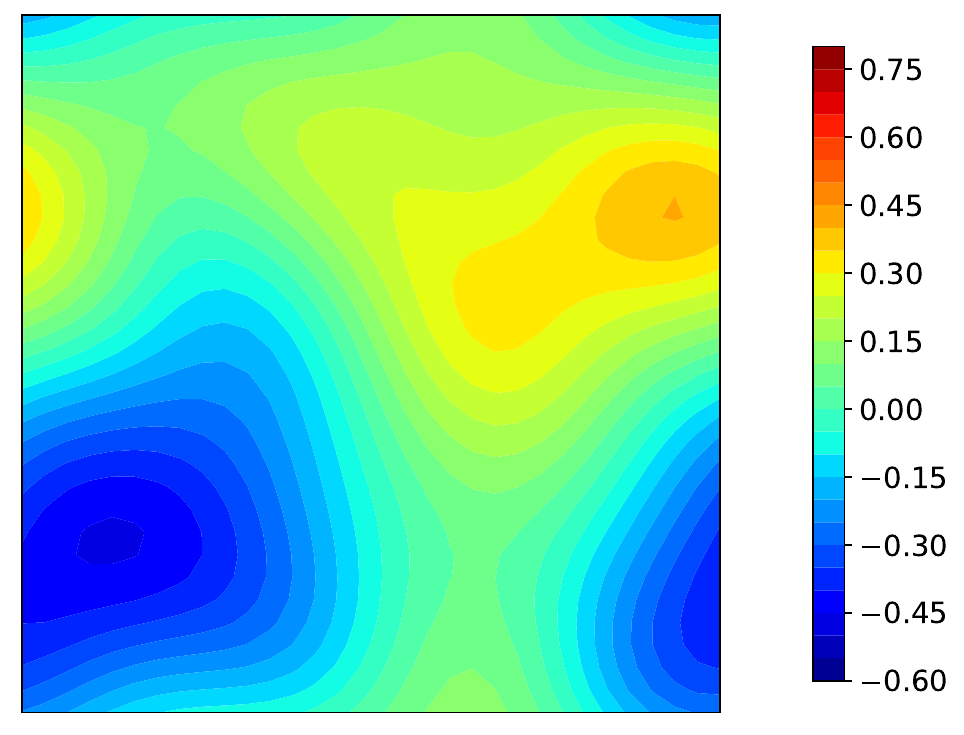}} &
        \raisebox{-0.5\height}{\includegraphics[width=0.30\textwidth]{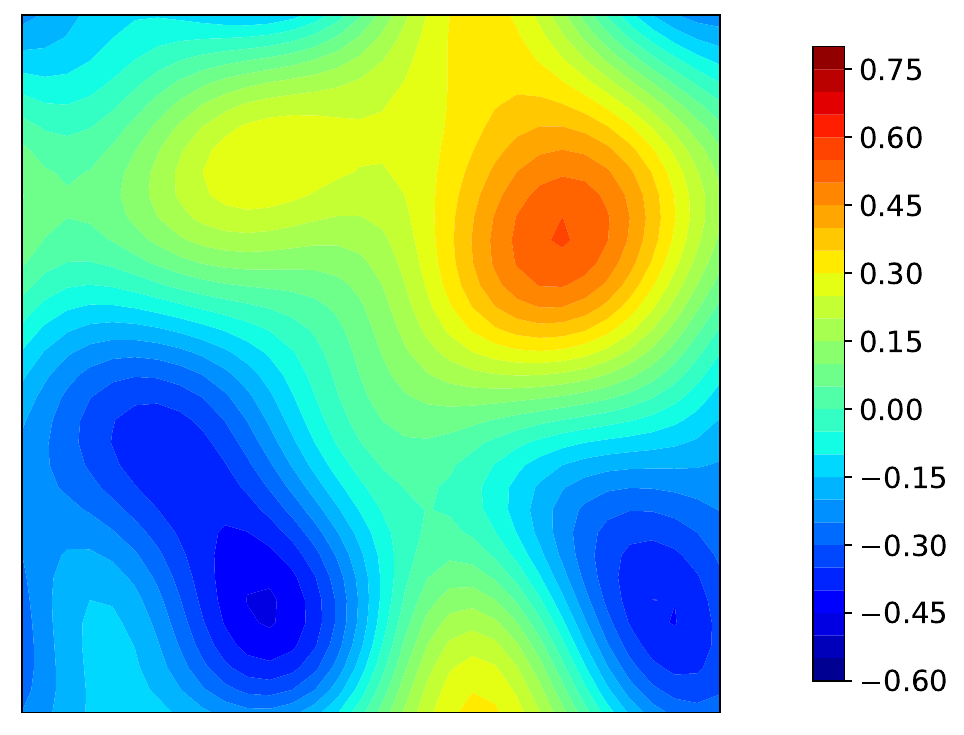}} &
        \raisebox{-0.5\height}{\includegraphics[width=0.30\textwidth]{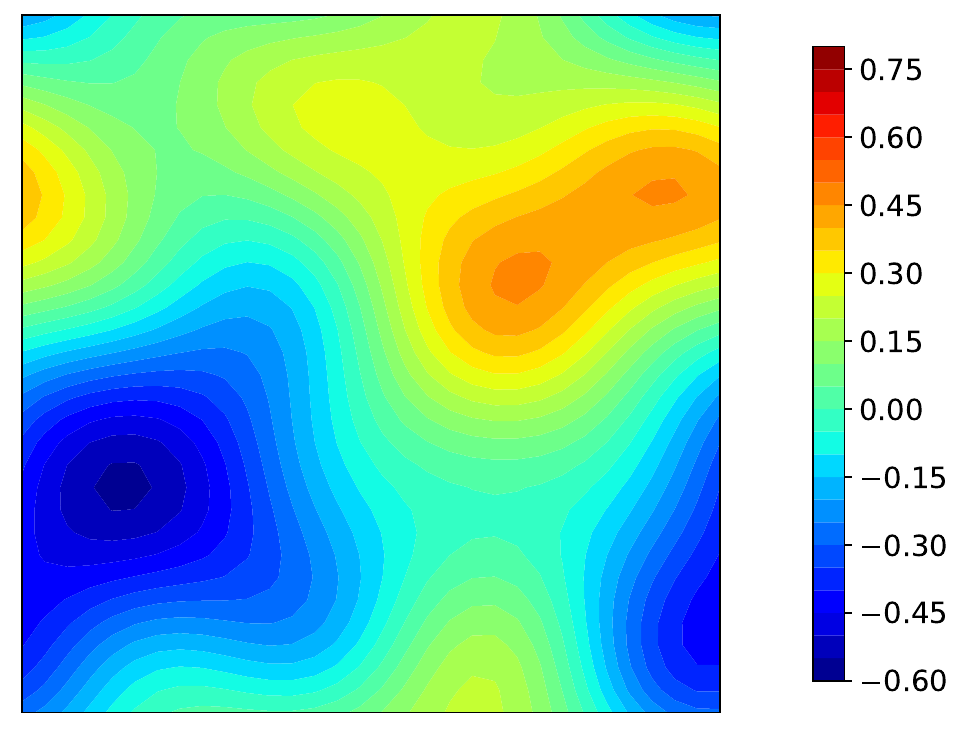}}
        \\ 
        \raisebox{-0.5\height}{\TNet{}$\quad\quad$} &
        \raisebox{-0.5\height}{Final vorticity $\quad\quad$} &
        \raisebox{-0.5\height}{TIK$\quad\quad$}
        \\
        \raisebox{-0.5\height}{\includegraphics[width=0.30\textwidth]{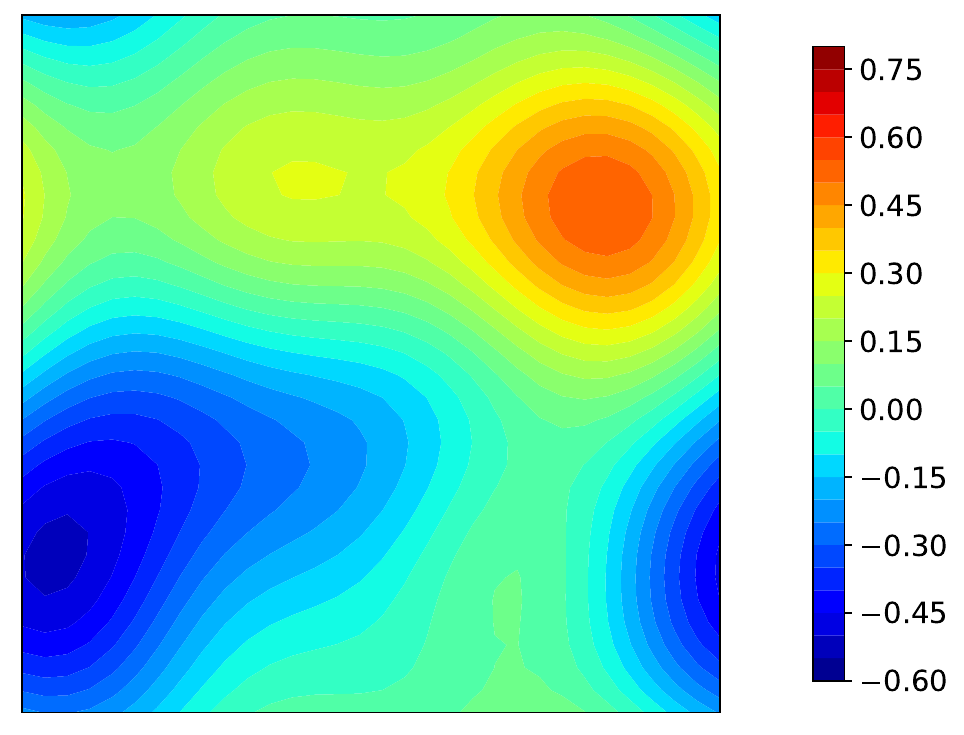}} &
        \raisebox{-0.5\height}{\includegraphics[width=0.30\textwidth]{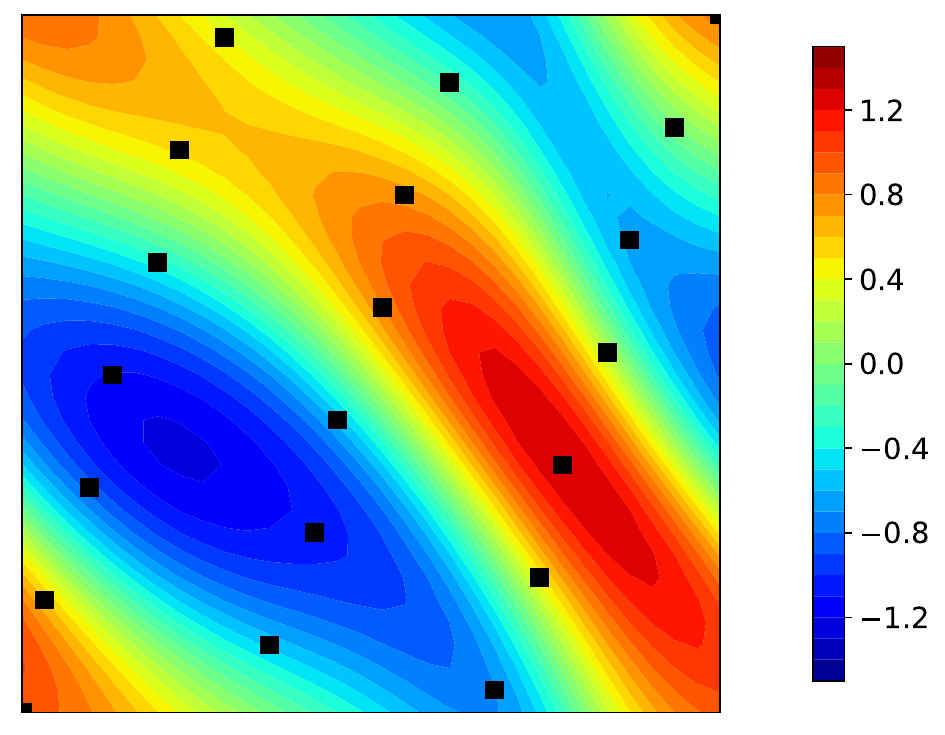}} &
        \raisebox{-0.5\height}{\includegraphics[width=0.30\textwidth]{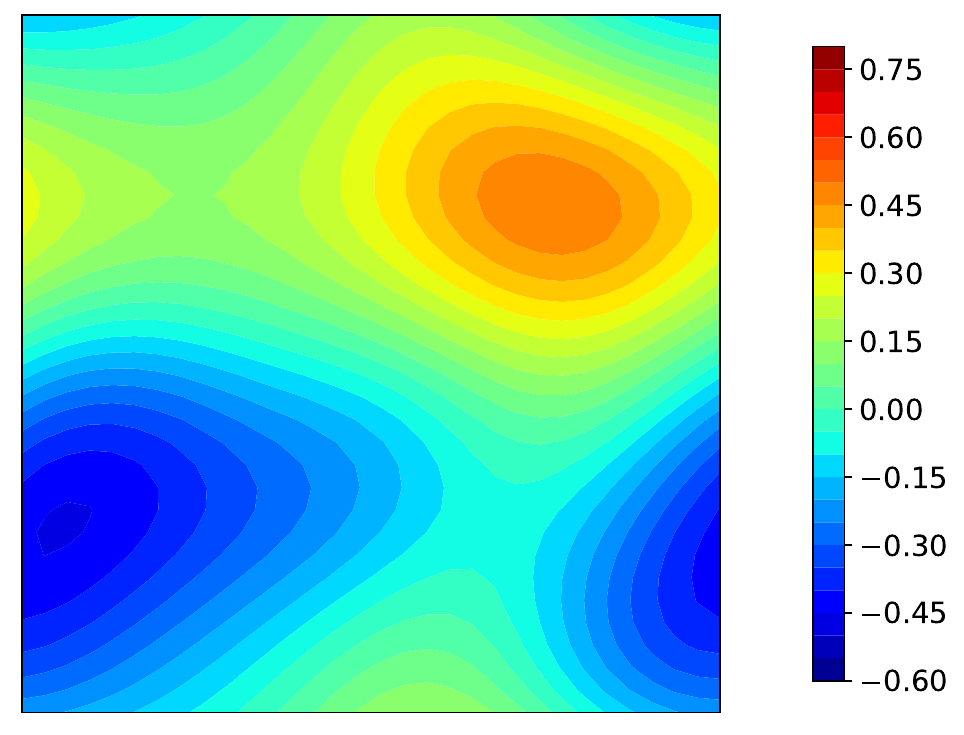}}
    \end{tabular*}
    \captionof{figure}{\textbf{2D Navier-Stokes equation, Case II.} 
    Predicted (reconstructed) initial vorticities for an unseen noisy test sample obtained by \nDNN{}, \mcDNN{}, and  \TNet{} neural networks along with the Tikhonov reconstruction for $n_b = 50$ and $n_t = 1000$.  Shown in the middle column are the synthetic ground truth (Exact) initial vorticity and the corresponding final vorticity for reference.
    }
    \figlab{2D_NS_50_database_predicted_test_samples}
\end{figure}

\subsection{Speedup with deep learning solutions}
We now compare the computational cost of reconstructing an unseen test sample by the classical Tikhonov (TIK) regularization technique and our proposed deep learning approach\footnote{Note that the cost for \nDNN{} and \mcDNN{} is the same as they have the same network architecture.} \TNet{} in \cref{tab:comp_cost}. We order the first column the increase in complexity from heat equation to Burger's equation to Navier-Stokes equation. Here, the complexity is estimated  based on the number of time steps, the operations carried out per time step, and the mesh size. As can be seen, the more complicated the problem is, the more time Tikhonov takes to obtain the solution.   Unlike the Tikhonov approach, regardless of the complexity of the problems, the learned \TNet{} inverse map using one hidden layer with 5000 neurons takes the same small amount of time: approximately $0.0003$ seconds. Note that the Tikhonov solver is implemented directly in \texttt{JAX} using the default BFGS algorithm with the gradient computed by the default Autograd functionality. Thus, the Tikhonov computation enjoys \texttt{JAX} optimized features including XLA (accelerated linear algebra), JIT (just-in-time compilation), and the nested primitive loop technique.
Even with such optimization, Tikhonov is still  orders of magnitude slower than \TNet{}.
In particular, for the Navier-Stokes equation, \TNet{} is $24,785$ times faster than Tikhonov.  We expect the computational gain is much more significant for larger-scale 3D time-dependent nonlinear forward problems. 
Clearly, once trained, obtaining \TNet{} solutions is simply a feed-forward neural network evaluation, which could be close to real-time or real-time depending on the depth and the width of the network.

\begin{table}[htb!]
\caption{The computational time (measured in seconds) for inverse solution governed by heat, Burgers, and Navier-Stokes equation using  \TNet{} (second column) and Tikhonov (third column) methods, and the speed-up (fourth column) of \TNet{} relative to Tikhonov using NVIDIA A100 GPUs on Lonestar6 at the Texas Advanced Computing Center (TACC).}
\tablab{comp_cost}
\centering
\begin{tabular}{|c|c|c|l|}
\hline
Inverse Problems               & \TNet{} (seconds)     & TIK (seconds)      & Speed-up \\ \hline
Heat equation          & $2.74 \times 10^{-4}$ & $4.36 \times 10^{-2}$ & \myred{159}          \\ \hline
Burger's equations        & $2.93  \times 10^{-4}$ & 1.08 & \myred{3,683}       \\ \hline
Navier-Stokes equation & $2.93  \times 10^{-4}$ & 7.26 & \myred{24,785}          \\ \hline
\end{tabular}
\end{table}

\section{Conclusions}
\seclab{conclusions}
We argue that in order for a DNN to generalize well in insufficient data regimes, it should be equipped with information encoded in the underlying mathematical model that is not or partially covered in the data set.  In other words, it is natural to require DNN to be aware of the underlying mathematical models (or discretizations) in order for it to be a {\em reliable and interpretable tool}  for sciences and engineering applications. Indeed, we have shown that the proposed model-constrained deep learning approaches are the same as Tikhonov regularization methods for linear inverse problems, while it is not clear if inverse solutions using purely data-driven DL methods are interpretable. We have shown that data randomization can further enhance the robustness and the generalization of our model-constrained deep neural networks. The numerical results 
not only confirm the theoretical findings but also show that  even with as little as $20$ training data samples for
1D deconvolution, $50$ for inverse heat conductivity, $100$ for inverse initial condition for time-dependent 2D Burgers' equation, and $50$ for inverse initial condition for 2D Navier-Stokes equations, \TNet{} solutions can be as accurate as Tikhonov solutions while being several orders of magnitude faster. This is not possible without the model-constrained term, replications, and randomization. Ongoing work is to understand under which conditions \TNet{} solution converges to Tikhonov solution for nonlinear inverse problems. Of interest is to estimate the minimum number of distinct baseline training samples to obtain a certain accuracy on average for unseen data. Extension to statistical inversions is also part of our future work.

\section*{Acknowledgments}
The authors would like to thank Krishnanunni Chandradath Girija, Jau-Uei Chen, Sheroze Sheriffdeen, Jonathan Wittmer, Hwan Goh, and Co Tran for fruitful discussions.
The authors also acknowledge the Texas Advanced Computing Center (TACC) at The University of Texas at Austin for providing {HPC, visualization, database, or grid} resources that have contributed to the research results reported within this paper. URL: http://www.tacc.utexas.edu

\bibliographystyle{plain}
\bibliography{references, referencesNew}

\begin{thebibliography}{10}

\bibitem{adler2017solving}
Jonas Adler and Ozan {\"O}ktem.
\newblock Solving ill-posed inverse problems using iterative deep neural
  networks.
\newblock {\em Inverse Problems}, 33(12):124007, 2017.

\bibitem{aggarwal2018modl}
Hemant~K Aggarwal, Merry~P Mani, and Mathews Jacob.
\newblock Modl: Model-based deep learning architecture for inverse problems.
\newblock {\em IEEE transactions on medical imaging}, 38(2):394--405, 2018.

\bibitem{Alifanov94}
O.~M. Alifanov.
\newblock {\em Inverse Heat Transfer Problems}.
\newblock Springer Verlag, Berlin, Heidelberg, New-York, 1994.

\bibitem{allen2022physical}
Kelsey~R Allen, Tatiana Lopez-Guevara, Kimberly Stachenfeld, Alvaro
  Sanchez-Gonzalez, Peter Battaglia, Jessica Hamrick, and Tobias Pfaff.
\newblock Physical design using differentiable learned simulators.
\newblock {\em arXiv preprint arXiv:2202.00728}, 2022.

\bibitem{an1996effects}
Guozhong An.
\newblock The effects of adding noise during backpropagation training on a
  generalization performance.
\newblock {\em Neural computation}, 8(3):643--674, 1996.

\bibitem{BN17_FEniCS_plus_DNN}
Jens Berg and Kaj Nyström.
\newblock Neural network augmented inverse problems for pdes, 2017.

\bibitem{Bishop06}
Christopher~M. Bishop.
\newblock {\em Pattern Recognition and Machine Learning (Information Science
  and Statistics)}.
\newblock Springer-Verlag, Berlin, Heidelberg, 2006.

\bibitem{bochev2006least}
Pavel~B Bochev and Max~D Gunzburger.
\newblock {\em Least-squares finite element methods}.
\newblock Springer, 2006.

\bibitem{jax2018github}
James Bradbury, Roy Frostig, Peter Hawkins, Matthew~James Johnson, Chris Leary,
  Dougal Maclaurin, George Necula, Adam Paszke, Jake Vander{P}las, Skye
  Wanderman-{M}ilne, and Qiao Zhang.
\newblock {JAX}: composable transformations of {P}ython+{N}um{P}y programs,
  2018.

\bibitem{BrennerScott02}
S.~C. Brenner and L.~R. Scott.
\newblock {\em The Mathematical Theory of Finite Element Methods}.
\newblock Springer Verlag, Berlin, Heidelberg, New York, second edition, 2002.

\bibitem{UnifedUniversalBui21}
Tan Bui-Thanh.
\newblock A unified and constructive framework for the universality of neural
  networks, 2021.

\bibitem{Bui-ThanhBursteddeGhattasEtAl12_gbfinalist}
Tan Bui-Thanh, Carsten Burstedde, Omar Ghattas, James Martin, Georg Stadler,
  and Lucas~C. Wilcox.
\newblock {Extreme-scale UQ for Bayesian inverse problems governed by PDEs}.
\newblock In {\em SC12: Proceedings of the International Conference for High
  Performance Computing, Networking, Storage and Analysis}, 2012.
\newblock {G}ordon {B}ell {P}rize finalist,
  http://users.ices.utexas.edu/\%7Etanbui/PublishedPapers/sc12.pdf.

\bibitem{CalvettiSomersalo07}
D.~Calvetti and E.~Somersalo.
\newblock {\em Introduction to Bayesian Scientific Computing: Ten Lectures on
  Subjective Computing}.
\newblock Springer, New York, 2007.

\bibitem{Chen20}
Yuyao Chen, Lu~Lu, George~Em Karniadakis, and Luca~Dal Negro.
\newblock Physics-informed neural networks for inverse problems in nano-optics
  and metamaterials.
\newblock {\em Opt. Express}, 28(8):11618--11633, Apr 2020.

\bibitem{Ciarlet02}
P.~G. Ciarlet.
\newblock {\em The finite element method for elliptic problems}, volume~40 of
  {\em Classics in Applied Mathematics}.
\newblock SIAM (SIAM), Philadelphia, PA, 2002.
\newblock Reprint of the 1978 original [North-Holland, Amsterdam; MR0520174 (58
  \#25001)].

\bibitem{constantine2016accelerating}
Paul~G Constantine, Carson Kent, and Tan Bui-Thanh.
\newblock Accelerating markov chain monte carlo with active subspaces.
\newblock {\em SIAM Journal on Scientific Computing}, 38(5):A2779--A2805, 2016.

\bibitem{Cybenko1989}
G.~Cybenko.
\newblock Approximation by superpositions of a sigmoidal function.
\newblock {\em Mathematics of Control, Signals and Systems}, 2(4):303--314, Dec
  1989.

\bibitem{ErnGuermond04}
Alexandre Ern and Jean-Luc Guermond.
\newblock {\em Theory and Practice of Finite Elements}, volume 159 of {\em
  Applied Mathematical Sciences}.
\newblock Spinger-{V}erlag, 2004.

\bibitem{fan2020solving}
Tiffany Fan, Kailai Xu, Jay Pathak, and Eric Darve.
\newblock Solving inverse problems in steady-state navier-stokes equations
  using deep neural networks.
\newblock {\em arXiv preprint arXiv:2008.13074}, 2020.

\bibitem{DEEPNET_difficulty_2010}
Xavier Glorot and Yoshua Bengio.
\newblock Understanding the difficulty of training deep feedforward neural
  networks.
\newblock In Yee~Whye Teh and Mike Titterington, editors, {\em Proceedings of
  the Thirteenth International Conference on Artificial Intelligence and
  Statistics}, volume~9 of {\em Proceedings of Machine Learning Research},
  pages 249--256, Chia Laguna Resort, Sardinia, Italy, 13--15 May 2010. PMLR.

\bibitem{goh2021solving}
Hwan Goh, Sheroze Sheriffdeen, and Tan Bui-Thanh.
\newblock Solving forward and inverse problems using autoencoders, 2019.

\bibitem{Goodfellow-et-al-2016}
Ian Goodfellow, Yoshua Bengio, and Aaron Courville.
\newblock {\em Deep Learning}.
\newblock MIT Press, 2016.
\newblock \url{http://www.deeplearningbook.org}.

\bibitem{Hochreiter01gradientflow}
Sepp Hochreiter, Yoshua Bengio, Paolo Frasconi, and Jürgen Schmidhuber.
\newblock Gradient flow in recurrent nets: the difficulty of learning long-term
  dependencies, 2001.

\bibitem{hornik1989multilayer}
Kurt Hornik, Maxwell Stinchcombe, and Halbert White.
\newblock Multilayer feedforward networks are universal approximators.
\newblock {\em Neural networks}, 2(5):359--366, 1989.

\bibitem{Jiang2020}
Jiaqi Jiang, Mingkun Chen, and Jonathan~A. Fan.
\newblock Deep neural networks for the evaluation and design of photonic
  devices.
\newblock {\em Nature Reviews Materials}, Dec 2020.

\bibitem{jin2020physics}
Yuchen Jin, Qiuyang Shen, Xuqing Wu, Jiefu Chen, and Yueqin Huang.
\newblock A physics-driven deep-learning network for solving nonlinear inverse
  problems.
\newblock {\em Petrophysics-The SPWLA Journal of Formation Evaluation and
  Reservoir Description}, 61(01):86--98, 2020.

\bibitem{johnson2018deep}
Jesse Johnson.
\newblock Deep, skinny neural networks are not universal approximators.
\newblock In {\em International Conference on Learning Representations}, 2019.

\bibitem{KaipioSomersalo05}
Jari Kaipio and Erkki Somersalo.
\newblock {\em Statistical and Computational Inverse Problems}, volume 160 of
  {\em Applied Mathematical Sciences}.
\newblock Springer-Verlag, New York, 2005.

\bibitem{kingma2014adam}
Diederik~P Kingma and Jimmy Ba.
\newblock Adam: A method for stochastic optimization.
\newblock {\em arXiv preprint arXiv:1412.6980}, 2014.

\bibitem{Kojima17}
Keisuke Kojima, Bingnan Wang, Ulugbek Kamilov, Toshiaki Koike-Akino, and Kieran
  Parsons.
\newblock Acceleration of fdtd-based inverse design using a neural network
  approach.
\newblock In {\em Advanced Photonics 2017 (IPR, NOMA, Sensors, Networks,
  SPPCom, PS)}, page ITu1A.4. Optical Society of America, 2017.

\bibitem{KomatitschRitsemaTromp02}
Dimitri Komatitsch, Jeroen Ritsema, and Jeroen Tromp.
\newblock The spectral-element method, {B}eowulf computing, and global
  seismology.
\newblock {\em Science}, 298:1737--1742, 2002.

\bibitem{LefebvreBozdaCalandraEtAl13}
Matthieu Lefebvre, Ebru Bozda, Henri Calandra, Judy Hill, Wenjie Lei, Daniel
  Peter, Norbert Podhorszki, David Pugmire, Herurisa Rusmanugroho, James Smith,
  and Jeroen Tromp.
\newblock A data centric view of large-scale seismic imaging workflows.
\newblock {\em Supercomputing (SC) 13}, 2013.
\newblock Invited paper.

\bibitem{li2020nett}
Housen Li, Johannes Schwab, Stephan Antholzer, and Markus Haltmeier.
\newblock Nett: Solving inverse problems with deep neural networks.
\newblock {\em Inverse Problems}, 36(6):065005, 2020.

\bibitem{FouierOP}
Zongyi Li, Nikola Kovachki, Kamyar Azizzadenesheli, Burigede Liu, Kaushik
  Bhattacharya, Andrew Stuart, and Anima Anandkumar.
\newblock Fourier neural operator for parametric partial differential
  equations, 2020.

\bibitem{LiuEtAl18}
Dianjing Liu, Yixuan Tan, Erfan Khoram, and Zongfu Yu.
\newblock Training deep neural networks for the inverse design of nanophotonic
  structures.
\newblock {\em ACS Photonics}, 5(4):1365--1369, 2018.

\bibitem{DeepXDE21}
Lu~Lu, Xuhui Meng, Zhiping Mao, and George~Em Karniadakis.
\newblock Deepxde: A deep learning library for solving differential equations.
\newblock {\em SIAM Review}, 63(1):208--228, 2021.

\bibitem{lu2021physicsinformed}
Lu~Lu, Raphael Pestourie, Wenjie Yao, Zhicheng Wang, Francesc Verdugo, and
  Steven~G. Johnson.
\newblock Physics-informed neural networks with hard constraints for inverse
  design, 2021.

\bibitem{Zhou17}
Zhou Lu, Hongming Pu, Feicheng Wang, Zhiqiang Hu, and Liwei Wang.
\newblock The expressive power of neural networks: A view from the width.
\newblock In I.~Guyon, U.~V. Luxburg, S.~Bengio, H.~Wallach, R.~Fergus,
  S.~Vishwanathan, and R.~Garnett, editors, {\em Advances in Neural Information
  Processing Systems}, volume~30. Curran Associates, Inc., 2017.

\bibitem{lunz2018adversarial}
Sebastian Lunz, Ozan {\"O}ktem, and Carola-Bibiane Sch{\"o}nlieb.
\newblock Adversarial regularizers in inverse problems.
\newblock {\em Advances in neural information processing systems}, 31, 2018.

\bibitem{Luo21}
Jie Luo, Xun Li, Xinyuan Zhang, Jiajie Guo, Wei Liu, Yun Lai, Yaohui Zhan, and
  Min Huang.
\newblock Deep-learning-enabled inverse engineering of multi-wavelength
  invisibility-to-superscattering switching with phase-change materials.
\newblock {\em Opt. Express}, 29(7):10527--10537, Mar 2021.

\bibitem{MohriEtAl12}
Mehryar Mohri, Afshin Rostamizadeh, and Ameet Talwalkar.
\newblock {\em Foundations of Machine Learning}.
\newblock The MIT Press, 2012.

\bibitem{mueller2012linear}
Jennifer~L Mueller and Samuli Siltanen.
\newblock {\em Linear and nonlinear inverse problems with practical
  applications}.
\newblock SIAM, 2012.

\bibitem{MuellerSiltanen12}
J.L. Mueller and S.~Siltanen.
\newblock {\em Linear and Nonlinear Inverse Problems with Practical
  Applications}.
\newblock SIAM, 2012.

\bibitem{NguyenBui_mcDNN21}
Hai~V. Nguyen and Tan Bui-Thanh.
\newblock Model-constrained deep learning approaches for inverse problems,
  2021.

\bibitem{nocedal1999numerical}
Jorge Nocedal and Stephen~J Wright.
\newblock {\em Numerical optimization}.
\newblock Springer, 1999.

\bibitem{OliverReynoldsLiu08}
Dean~S. Oliver, Albert~C. Reynolds, and Ning Liu.
\newblock {\em Inverse theory for petroleum reservoir characterization and
  history matching}.
\newblock Cambidge University Press, 2008.

\bibitem{pakravan2021solving}
Samira Pakravan, Pouria~A Mistani, Miguel~A Aragon-Calvo, and Frederic Gibou.
\newblock Solving inverse-pde problems with physics-aware neural networks.
\newblock {\em Journal of Computational Physics}, 440:110414, 2021.

\bibitem{fPINNs19}
Guofei Pang, Lu~Lu, and George~Em Karniadakis.
\newblock fpinns: Fractional physics-informed neural networks.
\newblock {\em SIAM Journal on Scientific Computing}, 41(4):A2603--A2626, 2019.

\bibitem{Pestourie2020}
Rapha{\"e}l Pestourie, Youssef Mroueh, Thanh~V. Nguyen, Payel Das, and
  Steven~G. Johnson.
\newblock Active learning of deep surrogates for pdes: application to
  metasurface design.
\newblock {\em npj Computational Materials}, 6(1):164, Oct 2020.

\bibitem{Peurifoyeaar4206}
John Peurifoy, Yichen Shen, Li~Jing, Yi~Yang, Fidel Cano-Renteria, Brendan~G.
  DeLacy, John~D. Joannopoulos, Max Tegmark, and Marin Solja{\v c}i{\'c}.
\newblock Nanophotonic particle simulation and inverse design using artificial
  neural networks.
\newblock {\em Science Advances}, 4(6), 2018.

\bibitem{pfaff2020learning}
Tobias Pfaff, Meire Fortunato, Alvaro Sanchez-Gonzalez, and Peter~W Battaglia.
\newblock Learning mesh-based simulation with graph networks.
\newblock {\em arXiv preprint arXiv:2010.03409}, 2020.

\bibitem{RaissiEtAl2019}
M.~Raissi, P.~Perdikaris, and G.E. Karniadakis.
\newblock Physics-informed neural networks: A deep learning framework for
  solving forward and inverse problems involving nonlinear partial differential
  equations.
\newblock {\em Journal of Computational Physics}, 378:686 -- 707, 2019.

\bibitem{RAISSI2019686}
M.~Raissi, P.~Perdikaris, and G.E. Karniadakis.
\newblock Physics-informed neural networks: A deep learning framework for
  solving forward and inverse problems involving nonlinear partial differential
  equations.
\newblock {\em Journal of Computational Physics}, 378:686--707, 2019.

\bibitem{RaissiKarniadakis2018}
Maziar Raissi and George~Em Karniadakis.
\newblock Hidden physics models: Machine learning of nonlinear partial
  differential equations.
\newblock {\em Journal of Computational Physics}, 357:125 -- 141, 2018.

\bibitem{RaissiEtAl2017}
Maziar Raissi, Paris Perdikaris, and George~Em Karniadakis.
\newblock Machine learning of linear differential equations using gaussian
  processes.
\newblock {\em Journal of Computational Physics}, 348:683 -- 693, 2017.

\bibitem{raissi2017physics2}
Maziar Raissi, Paris Perdikaris, and George~Em Karniadakis.
\newblock Physics informed deep learning (part ii): Data-driven discovery of
  nonlinear partial differential equations.
\newblock {\em arXiv preprint arXiv:1711.10566}, 2017.

\bibitem{ShwartzEtAl14}
Shai Shalev-Shwartz and Shai Ben-David.
\newblock {\em Understanding Machine Learning: From Theory to Algorithms}.
\newblock Cambridge University Press, USA, 2014.

\bibitem{singh2017machine}
Anand~Pratap Singh, Shivaji Medida, and Karthik Duraisamy.
\newblock Machine-learning-augmented predictive modeling of turbulent separated
  flows over airfoils.
\newblock {\em AIAA journal}, 55(7):2215--2227, 2017.

\bibitem{DNNInverseNanoPhotonnics20}
Sunae {So}, Trevon {Badloe}, Jaebum {Noh}, Jorge {Bravo-Abad}, and Junsuk
  {Rho}.
\newblock {Deep learning enabled inverse design in nanophotonics}.
\newblock {\em Nanophotonics}, 9(5):474, February 2020.

\bibitem{DEEPNET_difficulty_2013}
Ilya Sutskever, James Martens, George Dahl, and Geoffrey Hinton.
\newblock On the importance of initialization and momentum in deep learning.
\newblock In Sanjoy Dasgupta and David McAllester, editors, {\em Proceedings of
  the 30th International Conference on Machine Learning}, volume~28 of {\em
  Proceedings of Machine Learning Research}, pages 1139--1147, Atlanta,
  Georgia, USA, 17--19 Jun 2013. PMLR.

\bibitem{Tahersima2019}
Mohammad~H. Tahersima, Keisuke Kojima, Toshiaki Koike-Akino, Devesh Jha,
  Bingnan Wang, Chungwei Lin, and Kieran Parsons.
\newblock Deep neural network inverse design of integrated photonic power
  splitters.
\newblock {\em Scientific Reports}, 9(1):1368, Feb 2019.

\bibitem{Tarantola05}
Albert Tarantola.
\newblock {\em Inverse Problem Theory and Methods for Model Parameter
  Estimation}.
\newblock SIAM, Philadelphia, PA, 2005.

\bibitem{tikhonov1995numerical}
Andrei~Nikolaevich Tikhonov, AV~Goncharsky, VV~Stepanov, and Anatoly~G Yagola.
\newblock {\em Numerical methods for the solution of ill-posed problems},
  volume 328.
\newblock Springer Science \& Business Media, 1995.

\bibitem{TripathyBilionis2018}
Rohit~K. Tripathy and Ilias Bilionis.
\newblock Deep uq: Learning deep neural network surrogate models for high
  dimensional uncertainty quantification.
\newblock {\em Journal of Computational Physics}, 375:565 -- 588, 2018.

\bibitem{vogel2002computational}
Curtis~R Vogel.
\newblock {\em Computational methods for inverse problems}.
\newblock SIAM, 2002.

\bibitem{WHITE20191118}
Daniel~A. White, William~J. Arrighi, Jun Kudo, and Seth~E. Watts.
\newblock Multiscale topology optimization using neural network surrogate
  models.
\newblock {\em Computer Methods in Applied Mechanics and Engineering},
  346:1118--1135, 2019.

\bibitem{YangPerdikaris2019}
Yibo Yang and Paris Perdikaris.
\newblock Adversarial uncertainty quantification in physics-informed neural
  networks.
\newblock {\em Journal of Computational Physics}, 2019.

\bibitem{zhao2022learning}
Qingqing Zhao, David~B Lindell, and Gordon Wetzstein.
\newblock Learning to solve pde-constrained inverse problems with graph
  networks.
\newblock {\em arXiv preprint arXiv:2206.00711}, 2022.

\end{thebibliography}

\pagebreak
\begin{center}
\textbf{\large Supplemental Materials: \TNet: A Model-Constrained Tikhonov Network Approach for Inverse Problems}
\end{center}


\maketitle
\supplementary
\section{Proofs}
\seclab{proofs}

\begin{proof}[Proof of \cref{lem:TNET}]

Requiring the derivative of $ \mc{L}_{\mcDNN{}}$ in  \cref{eq:optLinear} with respect to $\bb$ to vanish yields
\begin{equation}
\eqnlab{bb_step1}
    \begin{aligned}
\LRp{\Ucovinv + \alpha \G^T \Ycovinv \G} \bb =     \LRs{\Ucovinv \LRp{\P - \W \Y} + \alpha \G^T \Ycovinv \LRp{ \Y - \G \W \Y }} \frac{\One}{n_t}.
    \end{aligned}
\end{equation}
Similarly, 
setting derivative of $ \mc{L}_{\mcDNN{}}$ with respect to $\W$ to zero gives
\begin{equation}
    \eqnlab{Wb_1}
    \begin{aligned}
     \Ucovinv \P \Y^T + \alpha \G^T \Ycovinv \Y \Y^T - \LRp{\Ucovinv + \alpha \G^T \Ycovinv \G} \LRp{\bb \One^T} \Y^T =  \LRp{\Ucovinv + \alpha \G^T \Ycovinv \G} \W \Y \Y^T.
    \end{aligned}
\end{equation}

Solving  \cref{eq:bb_step1} to \cref{eq:Wb_1} for $\bb$ and $\W$ we obtain
\begin{equation}
    \eqnlab{W1}
    \W = \LRp{\Ucovinv + \alpha \G^T \Ycovinv \G}^{-1} \LRp{\Ucovinv \Pbar \Ybar^\dagger + \alpha \G^T \Ycovinv \Ybar \Ybar^\dagger},
\end{equation}
and
\begin{equation}
    \eqnlab{b1}
    \bb = \LRp{\Ucovinv + \alpha \G^T \Ycovinv \G}^{-1}\LRs{\Ucovinv \ubbar + \alpha \G^T \Ycovinv \ybbar - \LRp{\Ucovinv \Pbar \, \Ybar^{\dagger} + \alpha\G^T \Ycovinv \Ybar \, \Ybar^{\dagger}} \ybbar},
\end{equation}
and this ends the proof.
\end{proof}

\begin{proof}[Proof of \cref{propo:HardC}]
Since \mcDNN{} and \TNet{} share the same model-constrained term, we only need to provide the proof for \TNet{}. Due to the assumption that \cref{eq:forward} is well-posed, we can write the \TNet{} training problem \cref{eq:TNETNonlinear} equivalently as
\begin{align}
	\eqnlab{TNetConstrained}
	\min_{\bb,\W} \mc{L}_{\TNet} &:=\half\nor{\P_0 - \DNN\LRp{\Y,\W,\bb}}_{\Ucovinv}^2 + \halfv{\alpha} \sum^{\nt}_{i=1}\nor{\yb^i - \F\LRp{\wb^i}}_{\Ycovinv}^2 \\
	\intertext{Subject to}
&\FW\LRp{\DNN\LRp{\yb^i,\W,\bb},\wb^i} = \fb, \quad i = 1,\hdots, \nt, \nonumber
\end{align}
which clearly shows that in fact \TNet{} formulation \cref{eq:TNETNonlinear} is a hard-constrained optimization problem that ensures the the forward equation \cref{eq:forward} to be satisfied exactly at all the training points during the training.
\end{proof}

\section{Practical implementation of \mcDNN{} and \TNet{}}
\seclab{implementation}
We discuss two approaches for training \mcDNN{} and \TNet{}. The first one is to implement the solution method of the forward equation \cref{eq:forward} inside a differentiable machine learning platform. This approach is suitable for small-to-medium size problems such as those in this paper. This is the approach that we take in this paper and \texttt{JAX} is the chosen platform. The advantage of this approach is that it exploits the automatic differentiation capability to straightforwardly apply any gradient-based learning algorithms. The main disadvantage is the implementation of a numerical solver in the machine learning platform under consideration. Clearly, this is not feasible for complex and/or large-scale codes.  

The second approach is to use the adjoint method to compute the gradient (and possibly Hessian-vector products) of $ \F$ with respect to $\DNN$ efficiently. 
This facilitates the reuse of existing adjoint-based large-scale inverse codes. The key hurdle to overcome is how to efficiently and scalably bridge  a machine learning platform (typically in \texttt{Python}) and an existing large-scale inversion framework (typically in \texttt{Fortran} or \texttt{C} or \texttt{C++}). Such a bridge is desirable as it enjoys the best from both sides. However, it is not a straightforward task as it requires interdisciplinary synergies among experts in numerical analysis, scientific computing, PDE-constrained optimization, and in computer sciences. Part of our ongoing work is to carry out this challenging task. 

\section{Training parameters}

\cref{tab:Summary_training_params} summarizes the specifications for neural network architectures, training settings, testing data sets, etc.
\begin{table}[H]
\centering
\caption{Summary of training parameters for \nDNN{}, \mcDNN{} and \TNet{} for nonlinear inverse  problems in \cref{sect:Heat_problem}, \cref{sect:Burger} and \cref{sect:2D_NS}.}
\begin{tabular}{|l l c |}
\hline
\multirow{5}{*}{Network}  & Architecture            & 1 layer with 5000 neurons                        \\ \cline{2-3} \\ [-10pt]
                          & Activation function     & ReLU                                             \\ \cline{2-3} \\ [-10pt]
                          & Weight initializer    & $\mc{N}\LRp{0, 0.02}$                            \\ \cline{2-3} \\ [-10pt]
                          & Bias initializer       & $\bs{0}$                                        \\ \cline{2-3} \\ [-10pt]
                          & Random seed             & 100                                              \\ \hline \\ [-10pt]
\multirow{3}{*}{Training} & Optimizer               & \texttt{ADAM}                                            \\ \cline{2-3} \\ [-10pt]
                          & Learning rate           & $10^{-3}$                                        \\ \cline{2-3} \\ [-10pt]
                          & Batch size              & 
                          $=\begin{cases}
                          n_t & \text{if }n_t \leq 500  \\
                          500 & \text{otherwise}
                          \end{cases}$
\\ [10pt] \hline \\ [-10pt]
                          
\multirow{4}{*}{Data}                      & Train data              & $n_t = p \subset n_t = q, \quad \text{if  } p < q \text{ and } p,q \in \mathbb{N}$ \\ \cline{2-3} \\ [-10pt]
                          & Test data               & 500 samples (drawn independently)                       \\ \cline{2-3} \\ [-10pt] 
                          & Train random seed & 18                                               \\ \cline{2-3} \\ [-10pt]
                          & Test random seed  & 28                                               \\ \hline  \\ [-10pt]
                          
\multirow{1}{*}{Precision}                      &               & Double precision  \\ \hline
\end{tabular}
\tablab{Summary_training_params}
\end{table}

\section{Additional numerical results}

\subsection{1D Linear deconvolution problem}
\seclab{1D_Linear}
In this problem, we compare the performance of \nDNN{}, \mcDNN{} and \TNet{} for linear inverse problem with linear neural network. In particular, we consider a one-dimensional deconvolution problem (see, e.g., \cite{CalvettiSomersalo07, MuellerSiltanen12} for the details) 
\[
y(s_j) = \int_0^1 \ \kappa(s_j, t) u(t) \, dt , \quad j = 1, \hdots,m,
\]
where $\kappa(s,t) = \frac{1}{\sqrt{2\pi \mu^2}} \exp \LRp{-\frac{1}{2\mu^2} \LRp{t-s}^2}$ is a Gaussian convolution kernel with $\mu = 0.05$, $s_j = \frac{j}{m} $ ($ 0 \leq j < m $), and $m =200$. A noise-free data pair $\LRp{\ub,\yb}$, with $\ub_j = u\LRp{s_j}$ and $\yb_j = y\LRp{s_j}$, is generated by two steps. Step 1: the parameter of interest $\ub$ is drawn from the prior distribution $\mc{N}\LRp{\bar{\ub}, \bs{\Gamma}}$, where
\[
\bar{\ub}_j = 10 \LRp{s_j-0.5} \exp \LRs{-50 \LRp{s_j-0.5}^2} -0.8 + 1.6 s_j, \quad j = 1, \hdots,m,
\]
and $\bs{\Gamma}$ is a  covariance matrix with Dirichlet boundary condition (again, see \cite{CalvettiSomersalo07}). Step 2: the observations are $20$ components randomly sampled from $m = 200$ components of $\yb$.

We consider three different cases of training data set: Case I: one pair of synthetic data $\LRp{\ub, \yb}$ is added with $n_t$ samples of noise vector $\epsb$ to produce a training data with $n_t$  samples;
Case II: ten pairs of synthetic data $\LRp{\ub, \yb}$, each of which is added with $n_t/10$ samples of noise vector $\epsb$ to produce a training data with $n_t$  samples;
Case III: $n_t$ pairs of synthetic data $\LRp{\ub, \yb}$ are randomized with different noise realizations. The relative noise level is $\delta = 5\%$ for all cases. The number of samples in the training data set, $n_t$, takes different values between 5 and  5000 (for Case II we take $n_t$ instead of $10$, if $n_t \le 10$), while a 200-pair noisy data set is generated for testing. In order to account for the random noise in the training data, we present the average behavior of the \TNet{} approach for each Case by first computing $100$ trained solutions, each with different noise samples, then averaging not only over the testing set but also over all the trained solutions. Since this is a linear inverse problem, we use a linear network, and thus \cref{coro:TNETDNN} applies. This means the ``trained" neural network solution is given explicitly by \cref{eq:TNetsolution} for each training data.

The comparison of relative errors of \nDNN{}, \mcDNN{}, \TNet{}, for Cases I, II, and III with $n_t \in  \LRs{5, 100}$, and the Tikhonov solution  is shown in \cref{fig:1D_all_data_base}. The top row shows the average relative error (see \cref{eq:Err}) versus the regularization parameter $\alpha$ and the training size $n_t$ for all Cases.  The pink, red, blue, and black curves are the minimum errors for different training sizes for \nDNN{}, \mcDNN{}, \TNet{}, and for Tikhonov. These curves are re-plotted  on the bottom row for clarity.
As can be seen, in all cases, \TNet{} solutions match the corresponding Tikhonov solutions when the number of training data size is larger than the number of observation points, regardless of which Case we consider. This numerically verifies the theoretical result in \cref{coro:TNETDNN}, as  $\Y$ has full row rank for these scenarios. This appealing property is not observed for \nDNN{} or \mcDNN{}. What we can see is that \nDNN{} and \mcDNN{} approaches provide more accurate reconstructions as we go from Case I to Case III since more distinct data are used to generate the training data set. In particular, \mcDNN{}'s errors are $10.42\%$, $10.39\%$, and $9.56\%$ for Cases I, II, and III, respectively. 
{\em While the optimal regularization parameter for both \TNet{} and \mcDNN{} are closed (in fact exactly for the former with $n_t \ge 20$) to the one for Tikhonov method, for \nDNN{} it erratically varies with the training size $n_t$}.
We also observe that, unlike \TNet{}, even though \mcDNN{} solutions are closed to\textemdash but not the same as\textemdash the Tikhonov ones. This is due to the data-driven nature of $\ub_0^{\mcDNN{}}$ as discussed in \cref{sect:TNET} after \cref{coro:TNETDNN}.
In summary, using Tikhonov solution as the reference, \TNet{} outperforms both \mcDNN{} and \nDNN{}. Solutions of  \mcDNN{} and \nDNN{} improve as the number of training data increases though the former is always more accurate: thanks to the model-constrained term (the second term in \cref{eq:optNonlinear}). 

\begin{figure}[htb!]
    \begin{center}
    \begin{tabular*}{\textwidth}{c c c}
        \centering
        \raisebox{-0.5\height}{\small Case I} &
        \raisebox{-0.5\height}{\small Case II} &
        \raisebox{-0.5\height}{\small Case III}
        \\
        \raisebox{-0.5\height}{\includegraphics[width=0.3\textwidth]{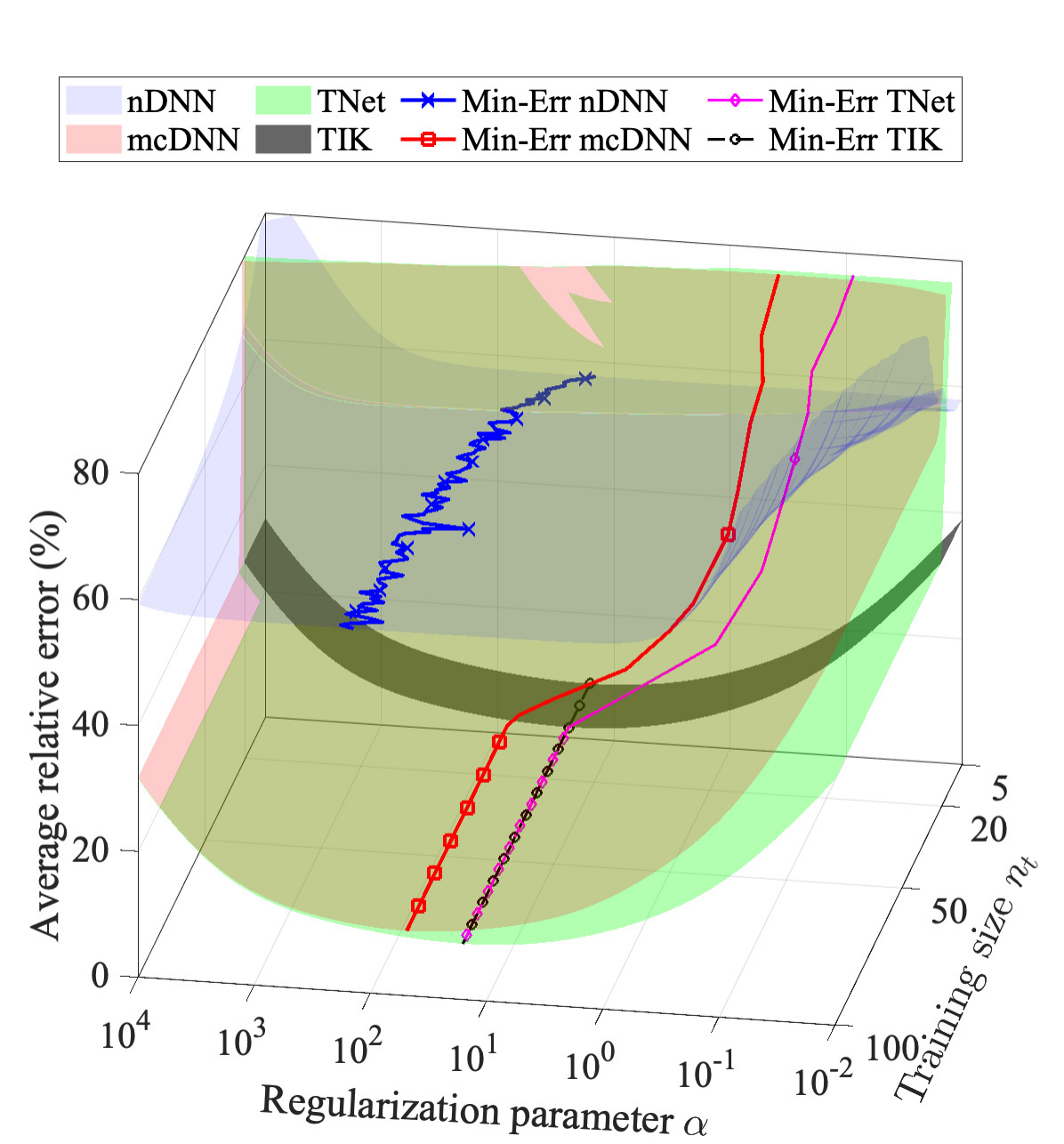}} &
        \raisebox{-0.5\height}{\includegraphics[width=0.3\textwidth]{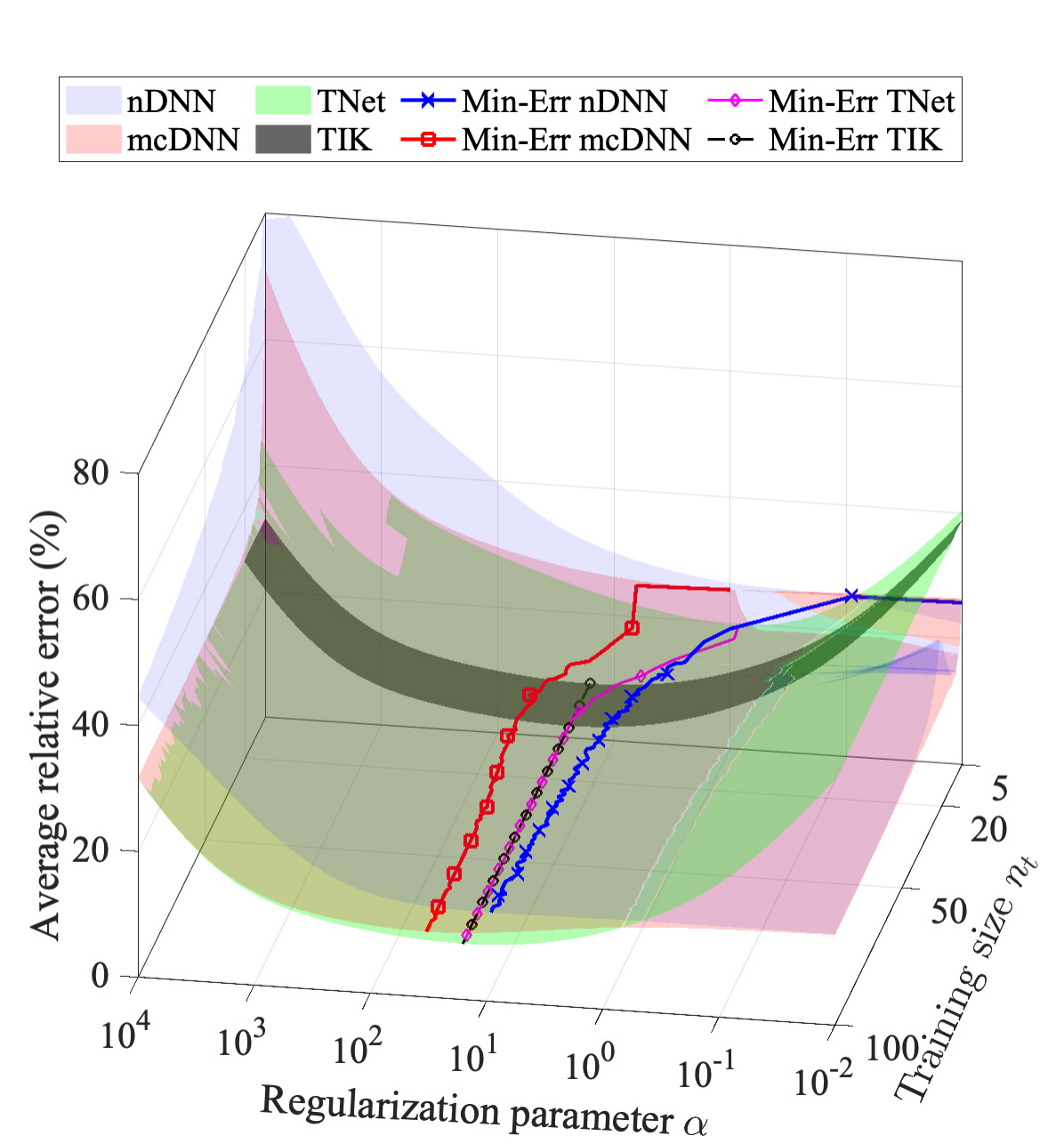}} & 
        \raisebox{-0.5\height}{\includegraphics[width=0.3\textwidth]{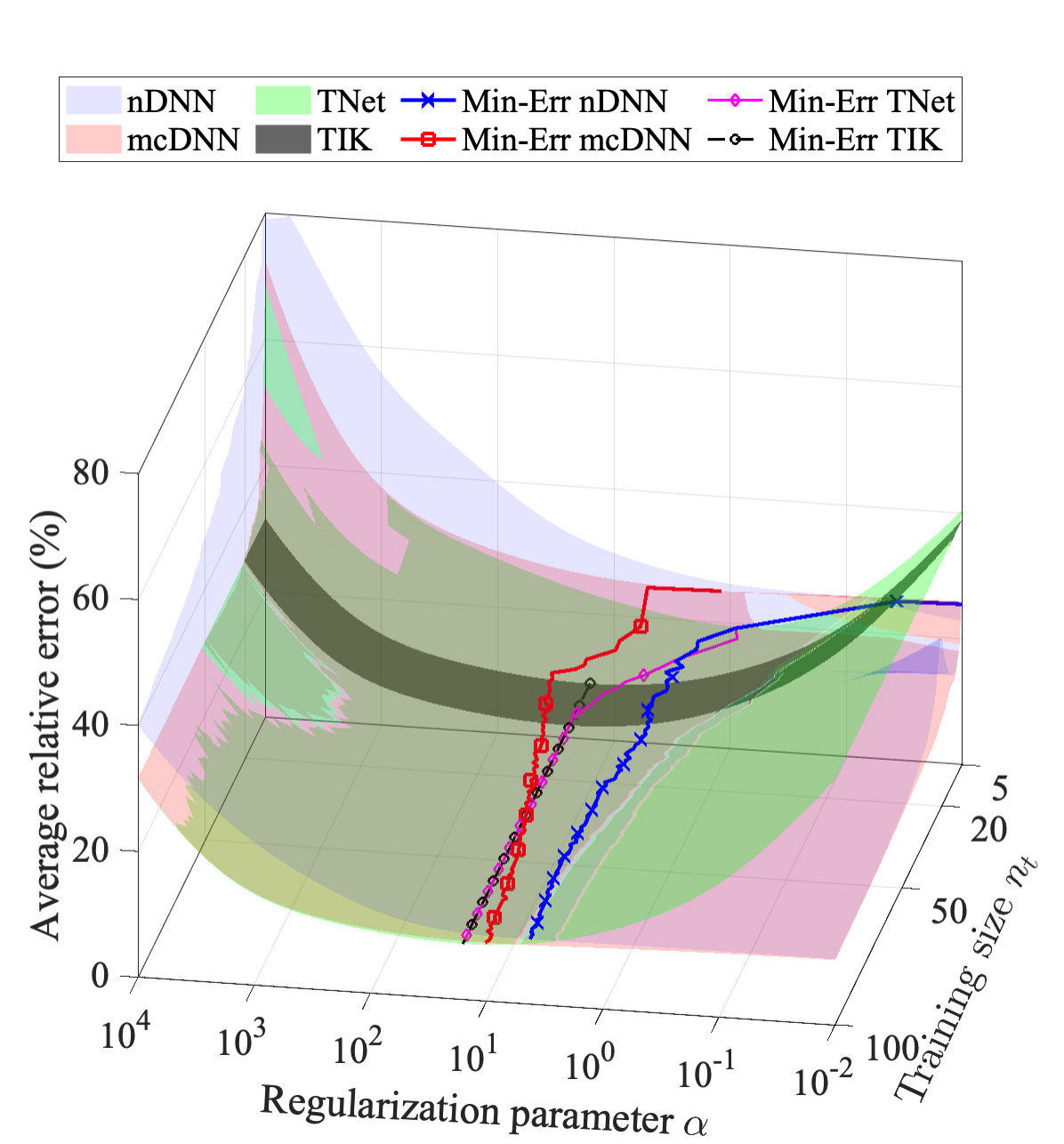}}
        \\
        \raisebox{-0.5\height}{\includegraphics[width=0.31\textwidth]{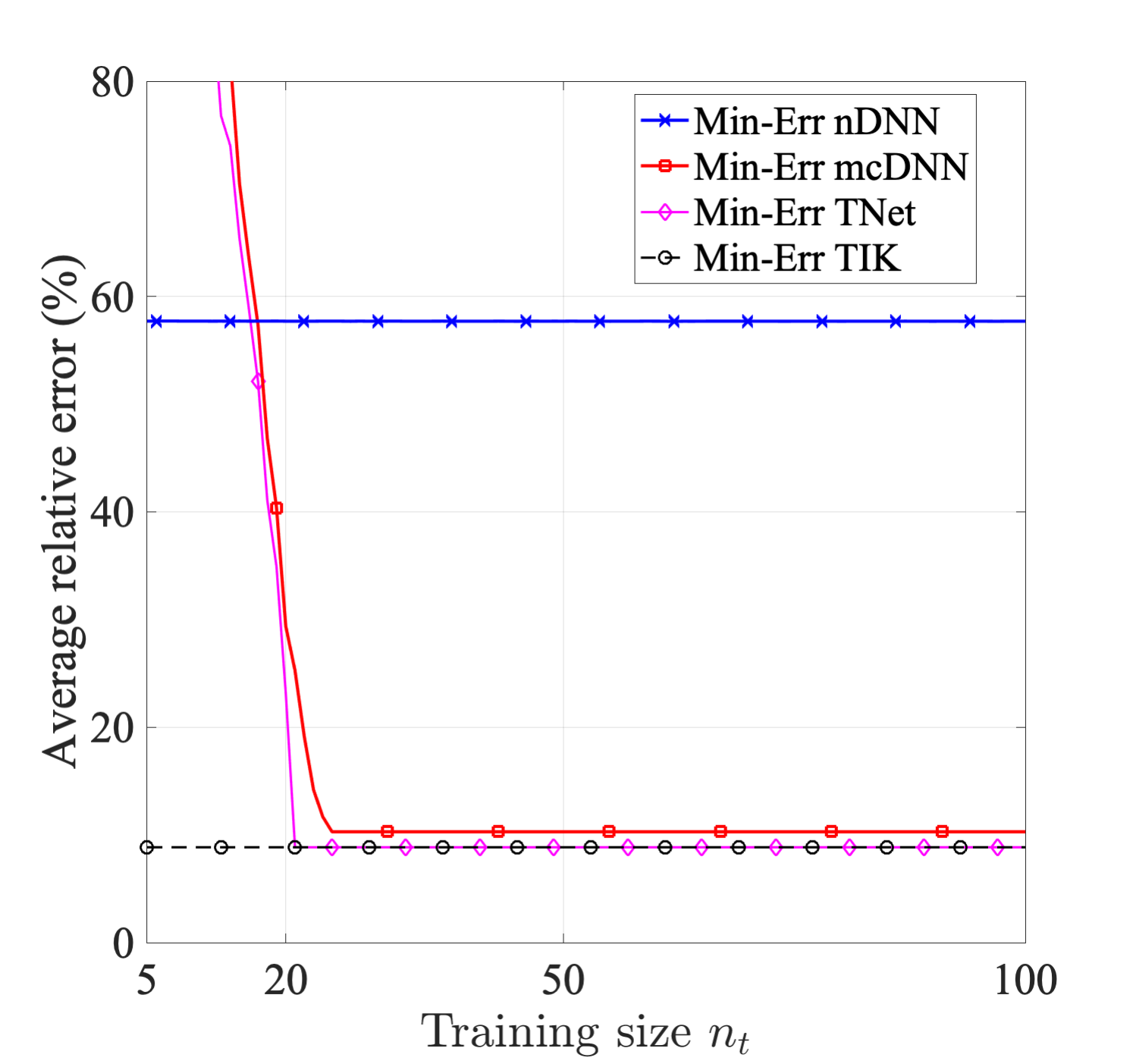}} & 
        \raisebox{-0.5\height}{\includegraphics[width=0.31\textwidth]{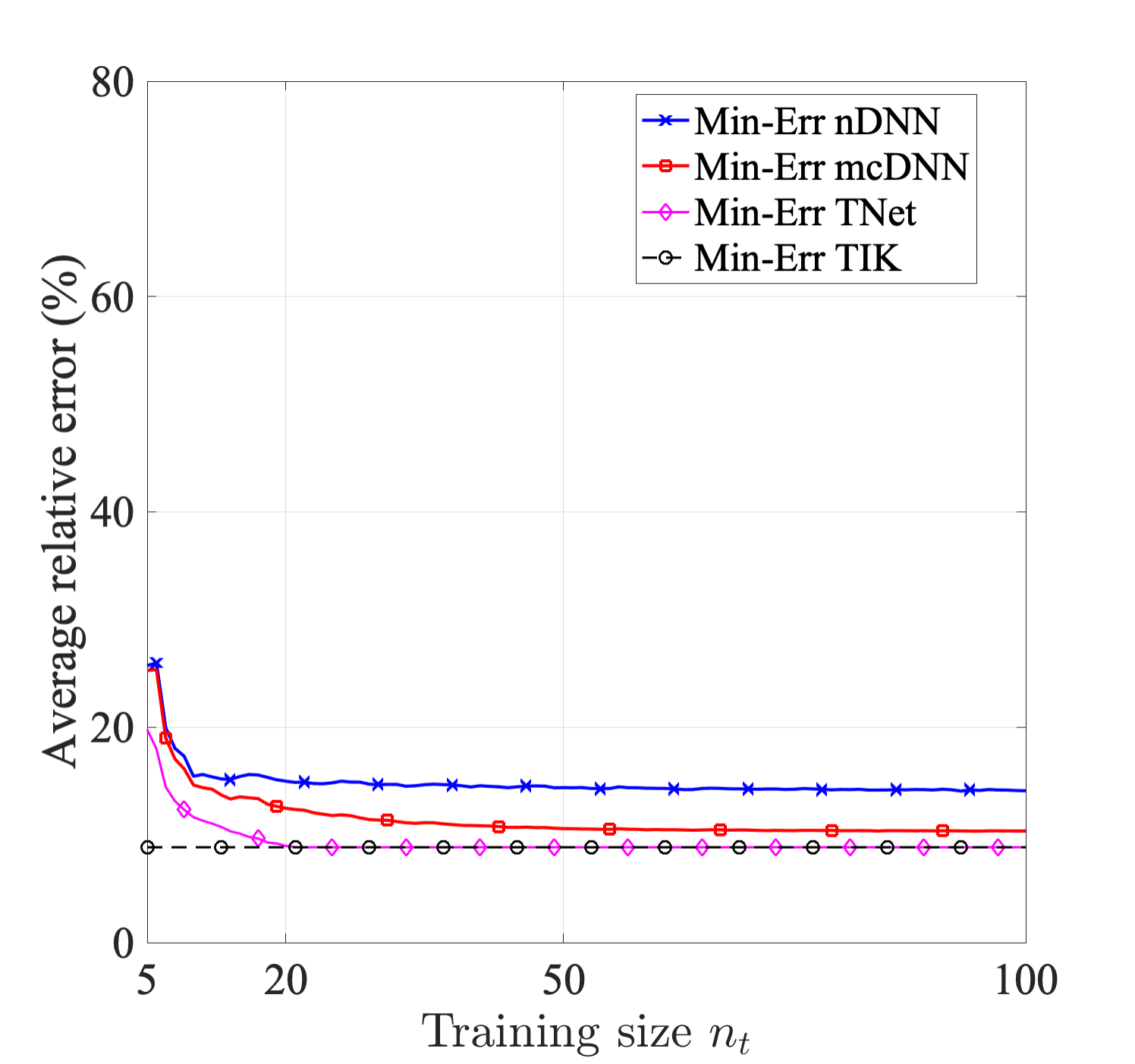}} & 
        \raisebox{-0.5\height}{\includegraphics[width=0.31\textwidth]{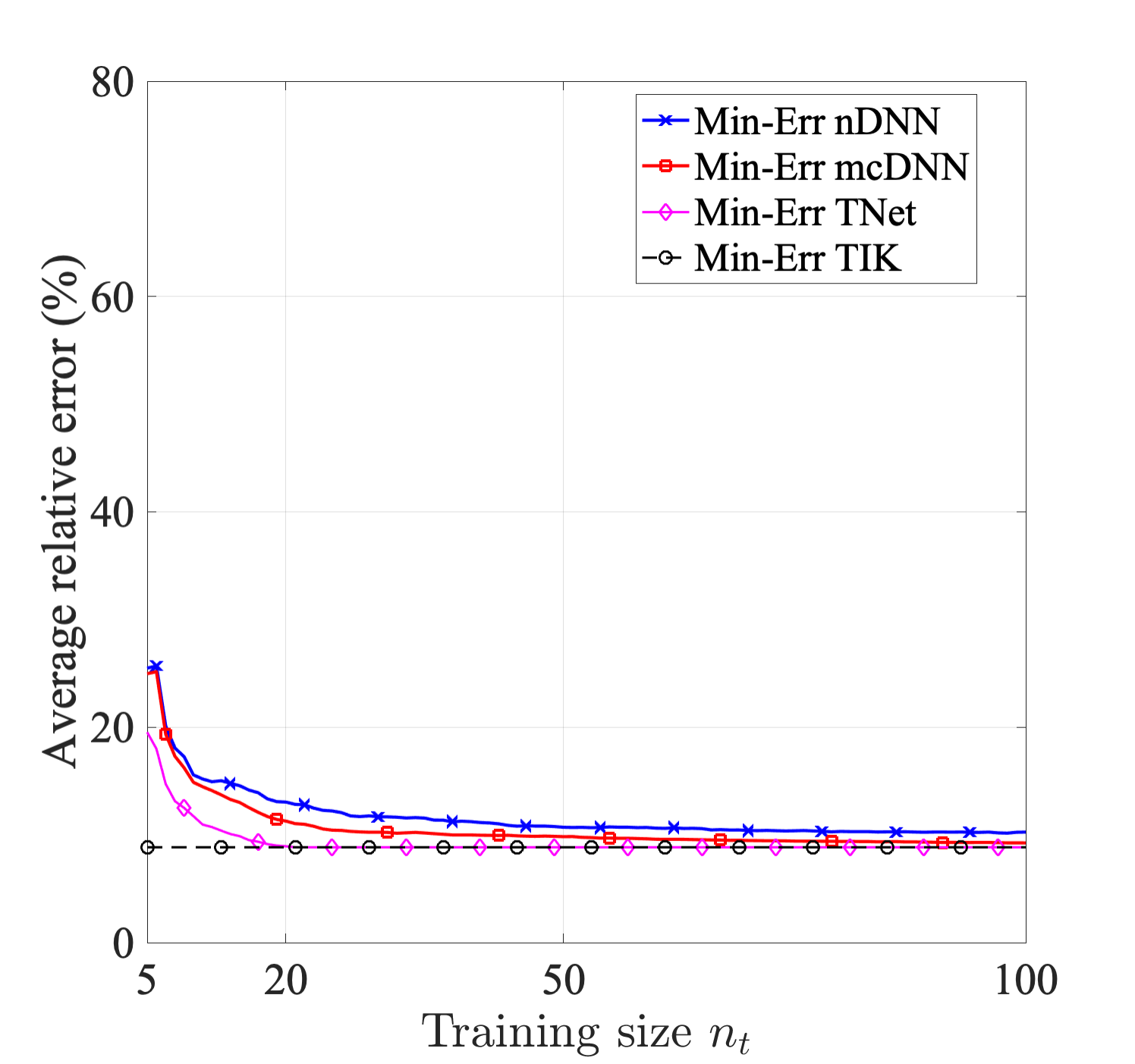}}
        
    \end{tabular*}
    \captionof{figure}{\textbf{1D Deconvolution.} The average relative error over $200$ test samples of \nDNN{}, \mcDNN{}, \TNet{} approaches
    for Cases I, II, and III, and the Tikhonov method.
\textit{Top row:} the relative error versus
the regularization parameter $\alpha$ and training size $n_t$. The pink, red, blue, and black curves are the minimum errors for different training sizes for \nDNN{}, \mcDNN{}, \TNet{}, and for Tikhonov. These curves are re-plotted  on the  \textit{Bottom row}  for clarity.
}
 \figlab{1D_all_data_base}
    \end{center}
\end{figure}

To understand how much data is needed for each method (except Tikhonov) until the minimum relative error starts saturating, we plot the average relative error at the optimal regularization parameter as a function of the training size for \nDNN{}, \mcDNN{}, \TNet{}, and  Tikhonov methods in  \cref{fig:1D_Relative_error}. Again, \TNet{} solutions reach the Tikhonov solutions when the number of training data samples is larger than the number of observation points, regardless of which Case we consider. 
However, \nDNN{} and \mcDNN{} approaches are still far away from Tikhonov solution even with $n_t = 5000$, though the latter is much more accurate. To be more specific, the relative error obtained by these methods for Case II (left subfigure) with training size between 1000 and 5000 samples have insignificant improvement. This is not surprising as the accuracy of \nDNN{} and \mcDNN{}  strongly depends on the information given in the training data set. The interesting point here is that data randomization, as predicted in \cref{sect:TNET}, improves the generalization (and thus accuracy) of \nDNN{} and \mcDNN{} approaches. Case III (the right subfigure) shows that more distinct training data provides  more accurate reconstructions for both \nDNN{} and \mcDNN. Furthermore, it is noticeable that \mcDNN{}, owing to the model-constrained term, converges to TIK at far fewer train data samples compared to \nDNN{}. In particular, around 500 training samples  \mcDNN{} quality is  almost the same as that of Tikhonov solution, while such a highly accurate result for \nDNN{} would need far beyond $5000$ training samples.

\begin{figure}[htb!]
    \centering
    \begin{tabular*}{\textwidth}{c c}
        \raisebox{-0.5\height}{Case II} &
        \raisebox{-0.5\height}{Case III}
        \\
        \raisebox{-0.5\height}{\includegraphics[width=0.47\textwidth]{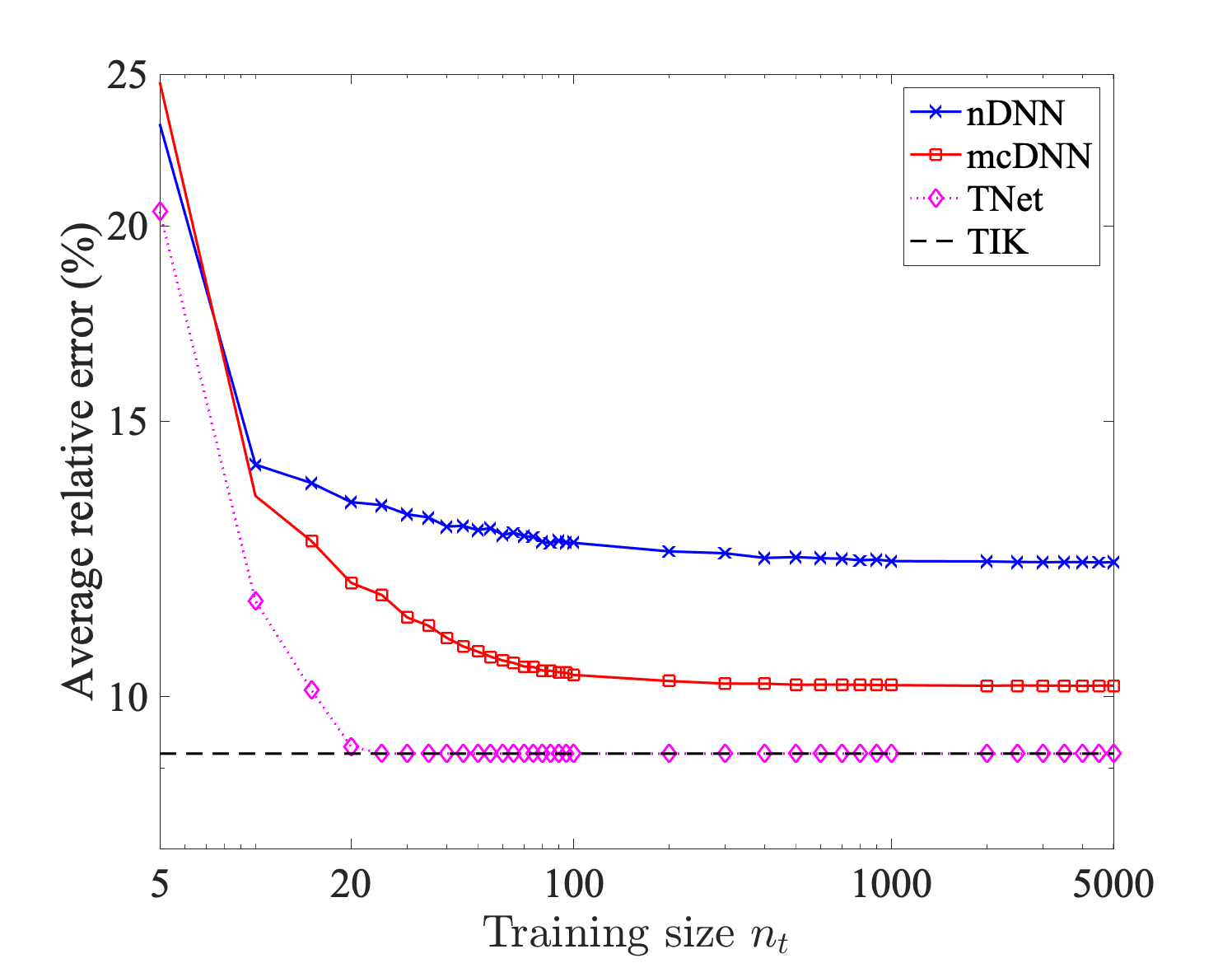}} &
        \raisebox{-0.5\height}{\includegraphics[width=0.47\textwidth]{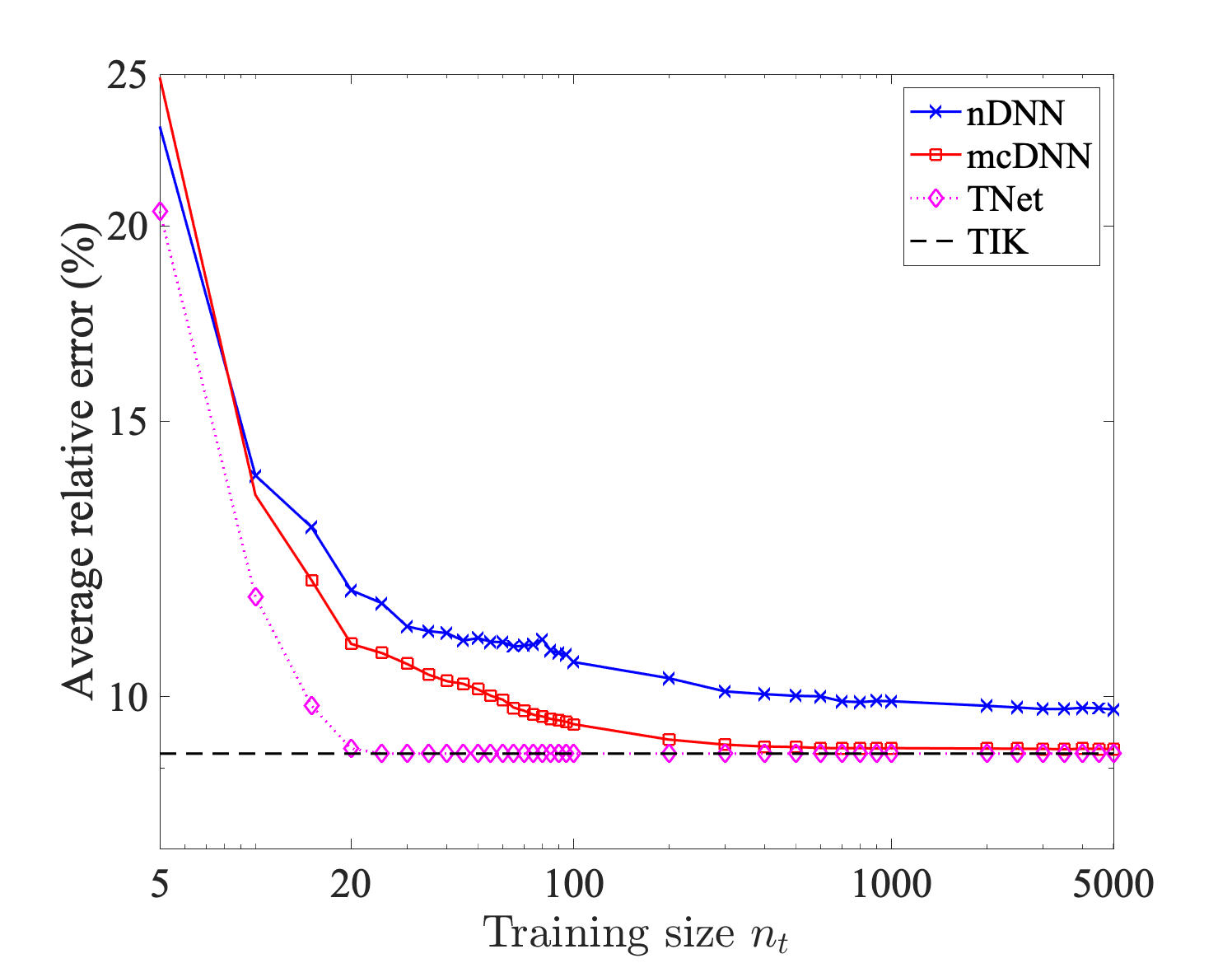}}
    \end{tabular*}
    \caption{\textbf{1D Deconvolution.} The average relative error at the optimal regularization parameter versus the training size for \nDNN{}, \mcDNN{}, \TNet{}, and  Tikhonov methods. The result for Case 
    II is on the left figure for training size up to $5000$, and a similar result is shown on the right figure for Case III.
    }
    \figlab{1D_Relative_error}
\end{figure}

\subsection{2D Burger's equations}
\seclab{Burger}

We consider the following viscous 2D Burger's equations
\begin{align*}
    & \pp{u}{t} + u \pp{u}{x} + u\pp{v}{y} = \nu \LRp{\pp{^2u}{x^2} + \pp{^2u}{y^2} } & x,y \in \LRp{0,1}, t \in (0,0.5], \\
    & \pp{v}{t} + v \pp{u}{x} + v\pp{v}{y} = \nu \LRp{\pp{^2v}{x^2} + \pp{^2v}{y^2} } & x,y \in \LRp{0,1}, t \in (0,0.5],
\end{align*}
subject to periodic boundary conditions, initial velocity components $v(x,y,0) = v_0(x,y) = 1$, $u(x,y,0) = u_0(x,y)$, and viscosity coefficient $\nu = 10^{-2}$. The spatial domain $\LRp{0,1}\times \LRp{0,1}$ is discretized with $n_x = 32$ and $n_y = 32$ mesh points in $x$ and $y$ directions, respectively, while the temporal domain $\LRp{0, 0.5}$ is subdivided into $n_t = 201$ time steps (including the initial time step $t = 0$). In this problem, the goal is to invert for the initial $x$-velocity $u_0$ from  20 pointwise values of  the vorticity $\ub_{0.5}$ (see the definition below) at the final time $T = 0.5$.

{\bf Construction of  train and test data sets.}
To generate training data for learning the inverse map, we draw periodic samples of $u(x,y,0)$ using a truncated Karhunen-Loève expansion
\[u(x,y,0) = \exp \LRp{\sum_{i=1}^{24} \sqrt{\lambda_i} \, \omega_i(x,y) \, z_i},\]
where $\textbf{z} = \LRc{z_i}_{i=1}^{24} \sim \mathcal{N} \LRp{0, \textbf{I}}$, and $\LRp{\lambda_i, \omega_i}$ are eigenpairs of the covariance $7^{\frac{3}{2}} \LRp{-\Delta + 49 \textbf{I}}^{-2.5}$ with periodic boundary conditions. 
Next, we discretize an initial vorticity $u(x,y,0)$, denoted as $\ub_0$, and
we solve the Burgers' equations via a finite difference method to compute the discrete x-velocity component at the final time: $\ub_{0.5}$. Finally, the values of $\ub_{0.5}$ at 20 locations 
are extracted and corrupted with an additive $2\%$ white noise to form  $\ybobs$. An example of a pair of $\LRp{\ub_0, \ub_{0.5}}$ is shown in
the middle column of \cref{fig:2D_Burger_50_database_predicted_test_samples}.


Similar to the heat conductivity inverse problem in \cref{sect:Heat_problem}, we consider two cases of train data.
Case I: Fully 
distinct training samples are used; and Case II: we first pick a number of distinct baseline samples $n_b$ smaller than $n_t$, and then replicate and randomize them  to obtain $n_t$ samples for the training data set. We shall compare and contrast results from \nDNN{}, \mcDNN{}, \TNet{}, and  Tikhonov solutions.

{\bf Case I: Training with fully distinct data sets $n_b = n_t \in \LRc{100, 200, 1000, 2000}$.}
\cref{fig:2D_Burger_regularization} presents the average relative error \cref{eq:Err} obtained by different approaches with different regularization values.
\begin{figure}[h!t!b!]
    \centering
    \includegraphics[width = 0.95\textwidth,clip]{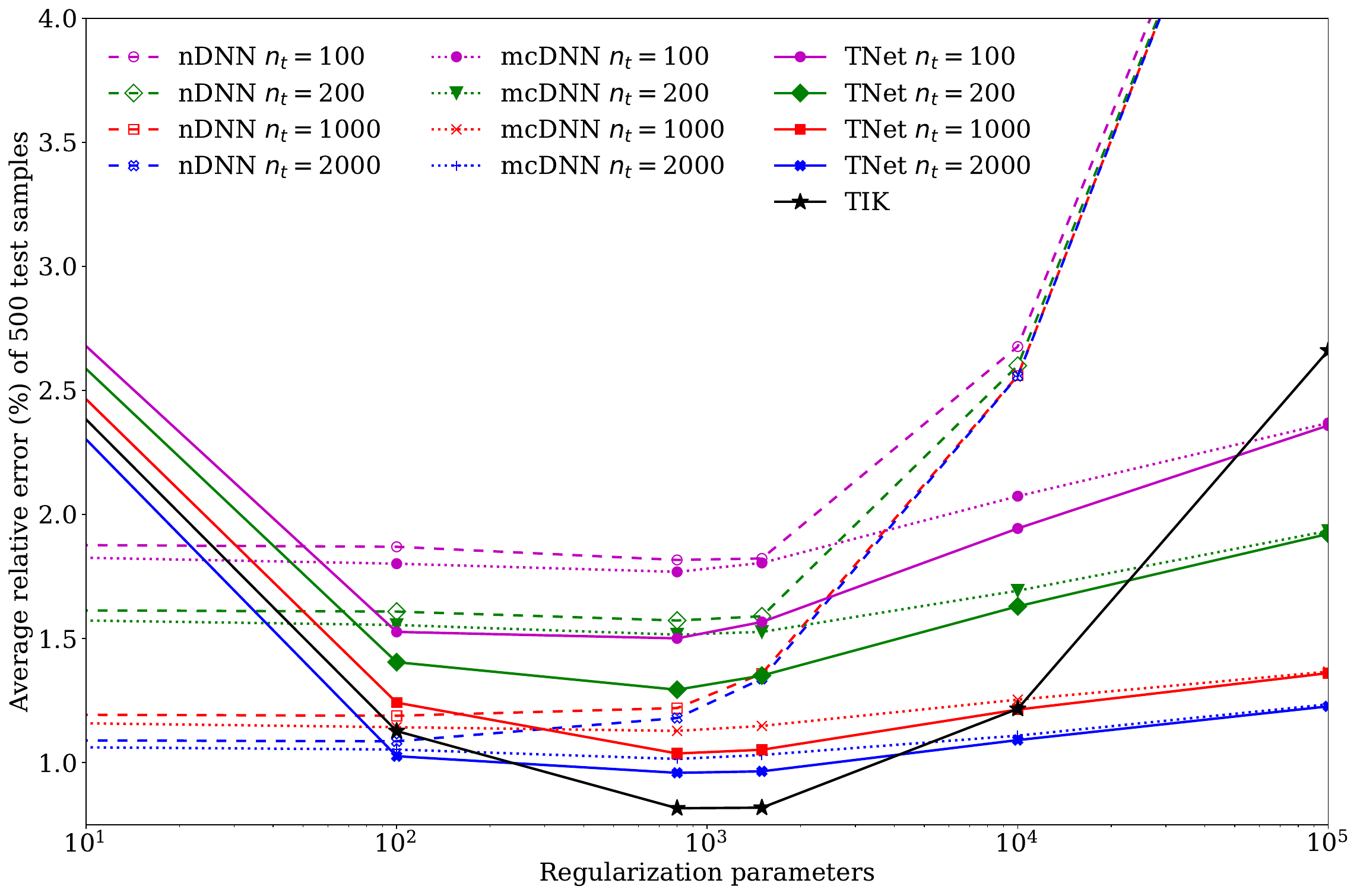}
    \caption{\textbf{2D Burger's equations, Case I.} The average relative error \cref{eq:Err} versus the regularization parameter $\alpha$ with 500 test samples with $\n_b = n_t = \LRc{100, 200, 1000, 2000}$. The results are shown for \nDNN{} (dashed curves), \mcDNN{} (dotted curves), \TNet{} (colored solids curves), and Tikhonov (TIK: black curve) solutions.}
    \figlab{2D_Burger_regularization}
\end{figure}
As can be observed, for regularization parameter range $\alpha \in \LRs{0.5\times 10^3, 10^5}$,  regardless of what $n_t$ is, \TNet{} has the highest accuracy, followed by \mcDNN{} and \nDNN{}, respectively. 
For $\alpha < 0.5\times 10^3$, \mcDNN{} is most accurate followed by \nDNN{} and \TNet{}. This is not surprising due to two reasons: i) moderately-to-significantly smaller regularization parameter relative to optimal Tikhonov regularization parameter (around $\alpha = 10^3$) is under-regularization and hence giving inaccurate results; and ii) \TNet{} aims to learn Tikhonov solutions. Consequently, the accuracy of \TNet{} is best for the regularization parameter inside a neighborhood of the optimal Tikhonov regularization parameter but degrades outside. A more detailed, but similar, discussion of the behavior of \TNet{} as a function of the regularization parameter can be found in \cref{sect:Heat_problem}. Here, {\em we focus on results at the ``best" regularization parameters for all methods.}

\begin{table}[h!t!b!]
\centering
\caption{\textbf{2D Burger's equations, Case I.} The average relative error \cref{eq:Err} over 500 test samples obtained by \nDNN{} (optimal $\alpha$ varies depending on the data set), \mcDNN{} ($\alpha = 800$), \TNet{} ($\alpha = 800$) with nested data sets $n_t = 100 \subset n_t = 200 \subset n_t = 1000 \subset n_t = 2000$, and Tikhonov (TIK) with $\alpha = 800$.
}
\tablab{Burger_full_data_base}
\begin{tabular}{|c|c|c|c|c|}
\hline
      & \nDNN{}  & \mcDNN{} & \TNet{}  & TIK                    \\ \hline
$n_t = $ 100  & 1.817 & 1.769 & 1.501 & \multirow{4}{*}{0.805} \\ \cline{1-4}
$n_t = $ 200  & 1.573 & 1.516 & 1.294 &                        \\ \cline{1-4}
$n_t = $ 1000 & 1.189 & 1.128 & 1.037 &                        \\ \cline{1-4}
$n_t = $ 2000 & 1.086 & 1.015 & 0.959 &                        \\ \hline
\end{tabular}
\end{table}

\begin{figure}[h!t!b!]
    \centering
    \includegraphics[width = 0.8\textwidth,clip]{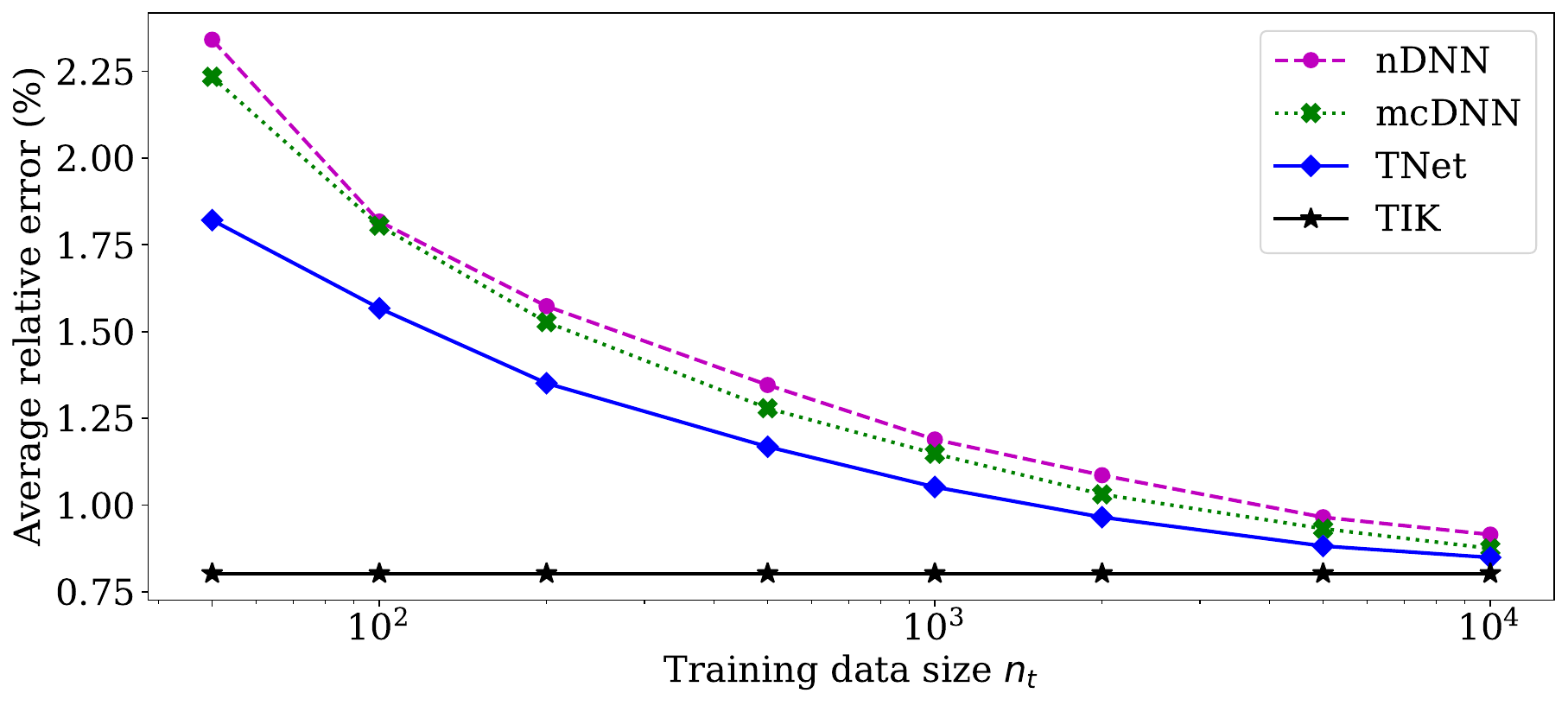}
    \caption{\textbf{2D Burger's equations, Case I.} Behavior of average error \cref{eq:Err} at the optimal regularization parameter for \TNet{}, \mcDNN{}, \nDNN{} as a function of training data size $n_t$.  Tikhonov (TIK) solution is used as the reference solution. }
    \figlab{2D_Burger_data_convergence}
\end{figure}

It is interesting to see that \TNet{} could get more accurate results even with smaller training data sets. For example, trained with $n_t = 100$ \TNet{} has  slightly better accuracy (1.501\%) compared to \mcDNN{} (1.516\%) and \nDNN{} (1.573\%) trained with $n_t = 200$. 
Unlike the deconvolution and heat conduction examples, \TNet{} behavior as a function of the regularization parameter does not follow Tikhonov closely, even with $n_t = 2000$. This is expected as Burger's equation is nonlinear and time-dependent, and we have only $20$ observations at the final time.

Nevertheless, \TNet{} is still significantly more accurate than \mcDNN{} and \nDNN{}.
 \cref{tab:Burger_full_data_base} provides the summary of relative error \cref{eq:Err} obtained by different approaches at the corresponding optimal regularization parameters. Clearly, \TNet{} is more accurate than \nDNN{} and \mcDNN{} approaches (see also \cref{fig:2D_Burger_data_convergence}). Similar to the other examples, unlike \nDNN{} for which the optimal regularization parameter depends on the training data set, \TNet{} and \mcDNN{} have the same optimal regularization parameter as Tikhonov approach regardless of training data. 
Shown in \cref{fig:2D_Burger_data_convergence} is the behavior of average error \cref{eq:Err} at the optimal regularization parameter as a function of training data size $n_t$. As can be seen,
though \mcDNN{}, \TNet{} and \nDNN{} seem to converge to the Tikhonov solution, \TNet{} has a faster convergence rate.

{\bf Case II: Training with $n_b \in \LRc{25, 100} \le n_t \in \LRc{100, 200, 500, 1000, 2000}$.}
Shown in \cref{tab:2D_Burger_n_data_base} are the average relative error \cref{eq:Err} for various training data sets. Recall that we start with $n_b \in  \LRc{25, 100} $ to generate training data sets with $n_t \in \LRc{100, 200, 500, 1000, 2000}$ by replication and randomization. As can be observed,
using train data sets with more distinct baseline pairs (i.e. with $n_b = 100$), and hence richer information, gives a more accurate inverse map for all approaches. 
Note that, within the same baseline case, \nDNN{} either does not improve or improves slightly when enriching the data with replication and randomization. 
In contrast, both \mcDNN{} and \TNet{} show better performance, and the latter is more accurate.

\begin{table}[htb!]
\caption{\textbf{2D Burger's equations, Case II.}
The average relative error \cref{eq:Err} for \nDNN{} (optimal $\alpha$ varies depending on the data set), \mcDNN{} ($\alpha = 800$), \TNet{}($\alpha = 800$), and Tikhonov (TIK) ($\alpha = 800$)  over 500-sample test data set with $n_b = \LRc{25, 100}$ baseline data pairs.
}
\tablab{2D_Burger_n_data_base}
\centering
\begin{tabular}{|c|ccc|ccc|c|}
\hline
\multirow{2}{*}{\begin{tabular}[c]{@{}c@{}}Training data\\ size ($n_t$)\end{tabular}} & \multicolumn{3}{c|}{$n_b = 25$}                          & \multicolumn{3}{c|}{$n_b = 100$} & \multirow{2}{*}{TIK}                        \\ \cline{2-7} 
 & \multicolumn{1}{c|}{\nDNN{}}  & \multicolumn{1}{c|}{\mcDNN{}} & \TNet{}  & \multicolumn{1}{c|}{\nDNN{}}  & \multicolumn{1}{c|}{\mcDNN{}} & \TNet{} & \\ \hline
$n_t = $ 100                                                                          & \multicolumn{1}{c|}{2.732} & \multicolumn{1}{c|}{2.596} & 2.077 & \multicolumn{1}{c|}{1.817} & \multicolumn{1}{c|}{1.805} & 1.566 & \multirow{5}{*}{0.805} \\ \cline{1-7}
$n_t = $ 200                                                                          & \multicolumn{1}{c|}{2.704} & \multicolumn{1}{c|}{2.260} & 1.871 & \multicolumn{1}{c|}{1.769} & \multicolumn{1}{c|}{1.650} & 1.439 & \\ \cline{1-7}
$n_t = $ 500                                                                          & \multicolumn{1}{c|}{2.725} & \multicolumn{1}{c|}{1.958} & 1.669 & \multicolumn{1}{c|}{1.764} & \multicolumn{1}{c|}{1.692} & 1.360 &  \\ \cline{1-7}
$n_t = $ 1000                                                                         & \multicolumn{1}{c|}{2.705} & \multicolumn{1}{c|}{1.789} & 1.576 & \multicolumn{1}{c|}{1.755} & \multicolumn{1}{c|}{1.515} & 1.276 &  \\ \cline{1-7}
$n_t = $ 2000                                                                         & \multicolumn{1}{c|}{2.707} & \multicolumn{1}{c|}{1.715} & 1.489 & \multicolumn{1}{c|}{1.761} & \multicolumn{1}{c|}{1.373} & 1.150 & \\ \hline
\end{tabular}
\end{table}

The distribution of average relative pointwise  errors \cref{eq:Errj}  for \nDNN{}, \mcDNN{}, \TNet{}, and Tikhonov (TIK) over 500 test samples obtained with $n_b = 100$ and $n_t = 2000$ are shown in \cref{fig:2D_Burger_100_database_average_error}.
 The numbers in the parentheses in the titles of subfigures are the average error \cref{eq:Err} incurred by each method.
 It can be seen that the pointwise error in the \TNet{} reconstruction is closest to that of Tikhonov solution, and  \nDNN{} has the largest error.
We would like to emphasize that the average error metrics \cref{eq:Errj} and \cref{eq:Err} (or similar) should be used to assess the performance of a neural network prediction as it reflects the overall average behavior of a neural network solution. The reason is that a network prediction can be less accurate for one case but more for another. In practice, after training and testing, we need to predict the inverse solution for a particular data. To see the solution of each method for a particular test sample, we pick a sample from 500 test samples and plot in
\cref{fig:2D_Burger_50_database_predicted_test_samples} the corresponding reconstructions from \nDNN{}, \mcDNN{}, \TNet{} and Tikhonov (TIK) methods. Clearly, overall \TNet{} result is in good agreement with TIK, and thus the ground truth, while 
\mcDNN{} and \nDNN{} results are less accurate. Again, this conclusion does not always hold as there are test samples on which \TNet{} could less accurate than either \nDNN{} or \mcDNN{}.

\begin{figure}[h!t!b!]
    \centering
    \begin{tabular*}{\textwidth}{c c c c}
        \centering
        \raisebox{-0.5\height}{\nDNN{} (1.761)} &
        \raisebox{-0.5\height}{\mcDNN{} (1.373)} &
        \raisebox{-0.5\height}{\TNet{} (1.150)} &
        \raisebox{-0.5\height}{TIK (0.805)}
        \\
        \raisebox{-0.5\height}{\includegraphics[width=0.22\textwidth]{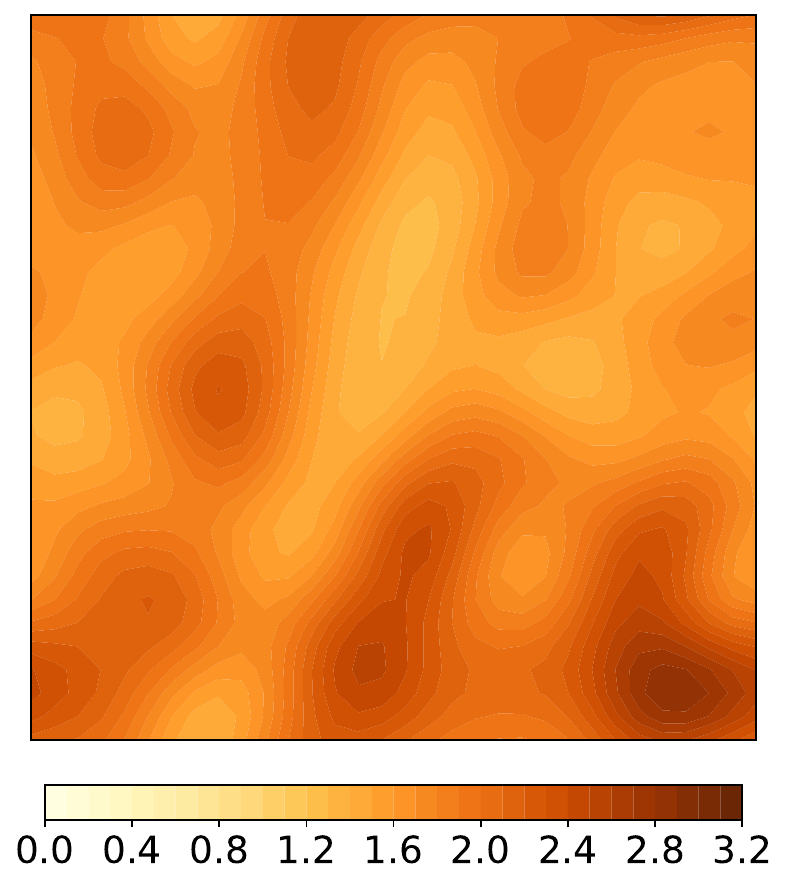}} &
        \raisebox{-0.5\height}{\includegraphics[width=0.22\textwidth]{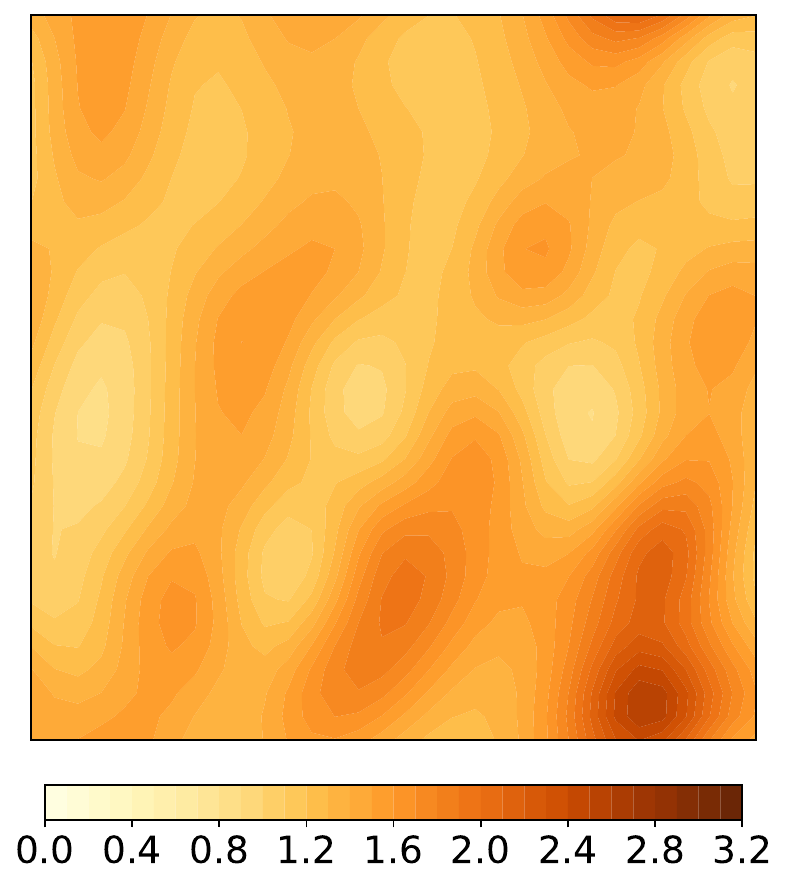}} &
        \raisebox{-0.5\height}{\includegraphics[width=0.22\textwidth]{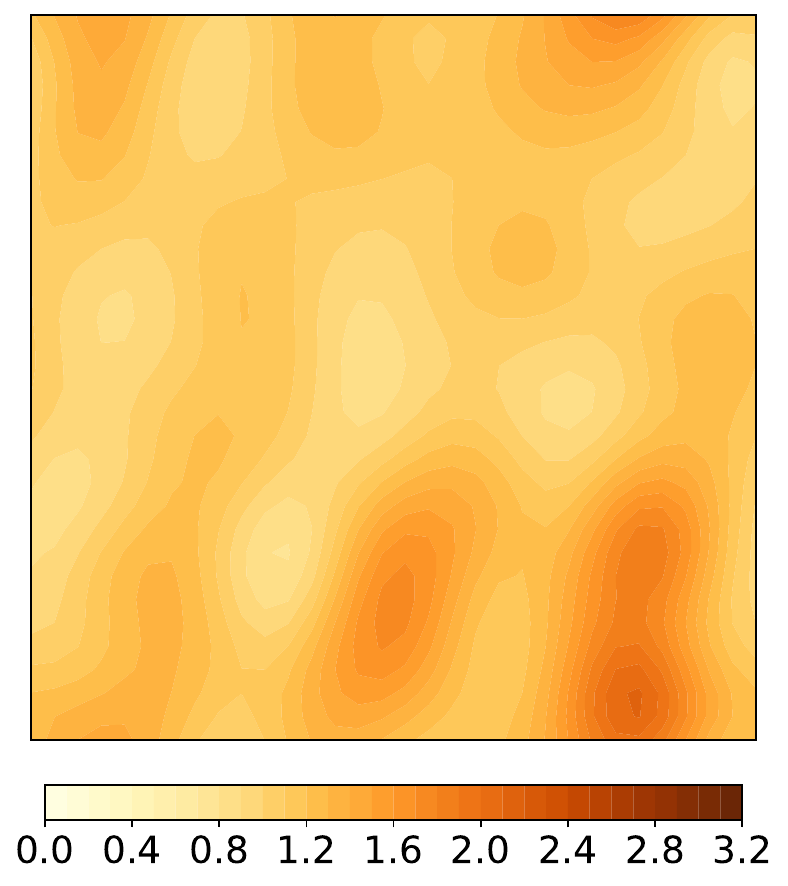}} &
        \raisebox{-0.5\height}{\includegraphics[width=0.22\textwidth]{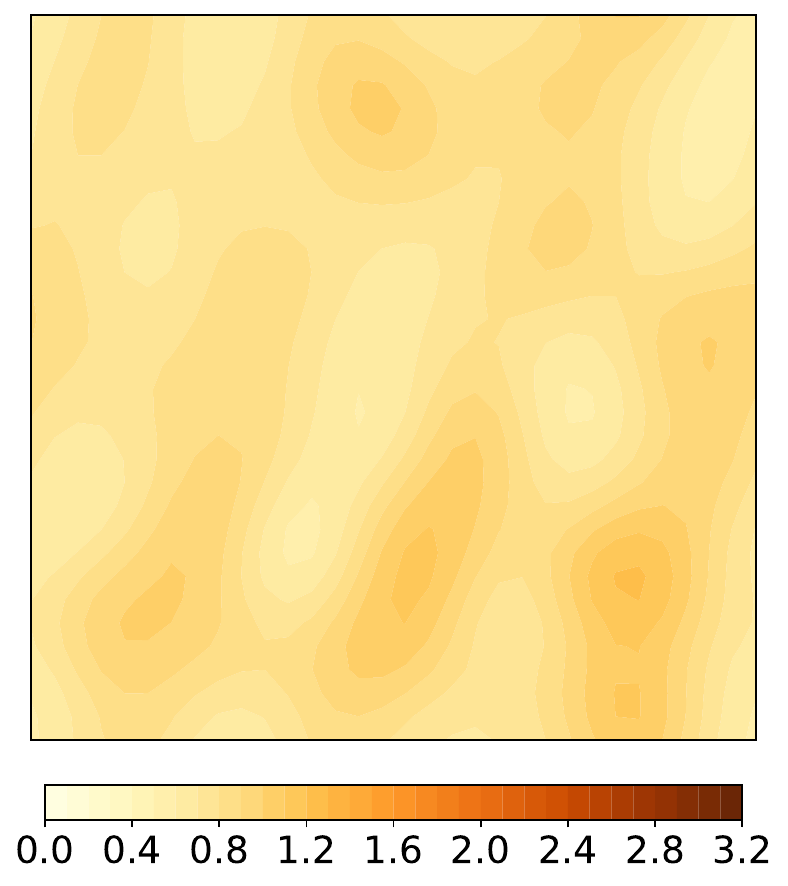}}
    \end{tabular*}
    \captionof{figure}{\textbf{2D Burger's equations, Case II.} 
    The distribution of average relative pointwise  error \cref{eq:Errj}  for \nDNN{}, \mcDNN{}, \TNet{}, and Tikhonov (TIK) over 500 test samples obtained with $n_b = 100$ and $n_t = 2000$. The numbers in the parentheses are the average error \cref{eq:Err} incurred by these methods.
    }
    \figlab{2D_Burger_100_database_average_error}
\end{figure}

\begin{figure}[h!t!b!]
    \centering
    \begin{tabular*}{\textwidth}{c c c}
        \centering
        \raisebox{-0.5\height}{\nDNN{}$\quad\quad$} &
        \raisebox{-0.5\height}{Exact$\quad\quad$} &
        \raisebox{-0.5\height}{\mcDNN{}$\quad\quad$}
        \\
        \raisebox{-0.5\height}{\includegraphics[width=0.30\textwidth]{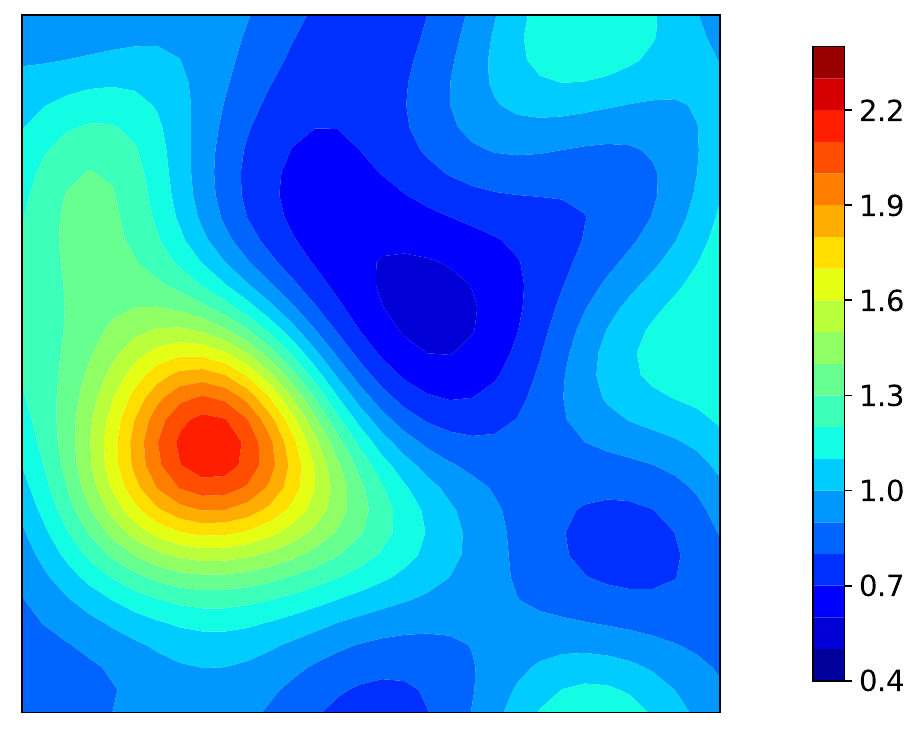}} &
        \raisebox{-0.5\height}{\includegraphics[width=0.30\textwidth]{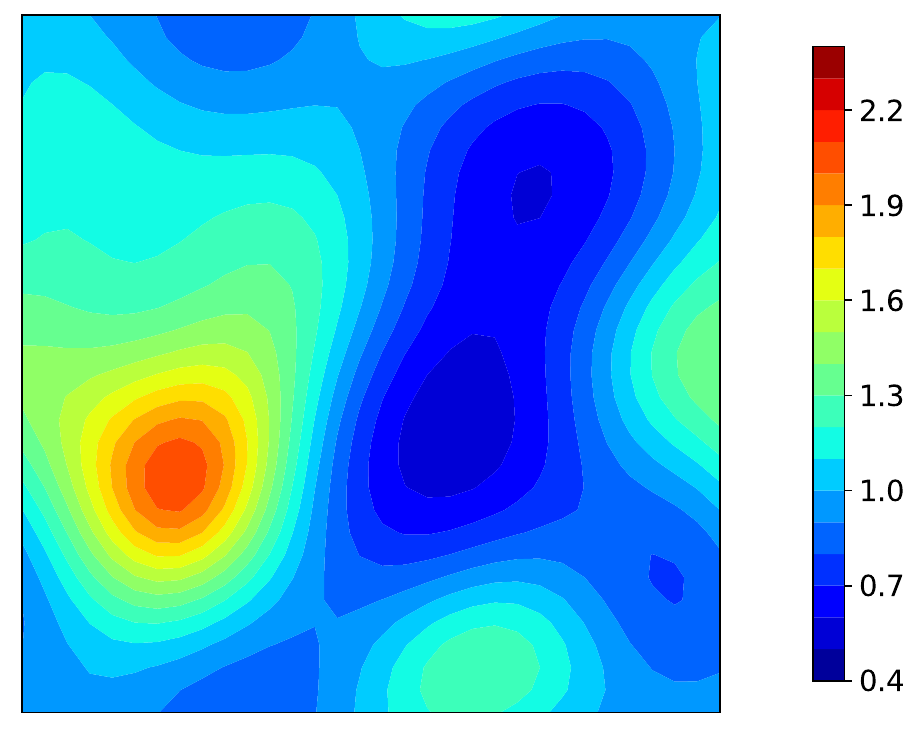}} &
        \raisebox{-0.5\height}{\includegraphics[width=0.30\textwidth]{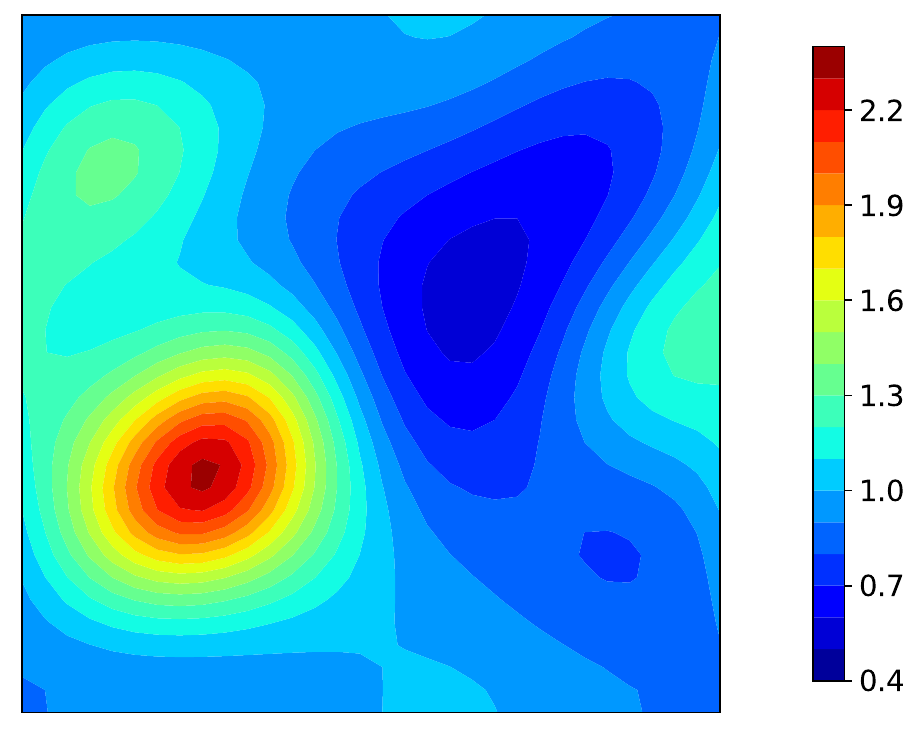}}
        \\ 
        \raisebox{-0.5\height}{\TNet{}$\quad\quad$} &
        \raisebox{-0.5\height}{Final $x-$velocity $\quad\quad$} &
        \raisebox{-0.5\height}{TIK$\quad\quad$}
        \\
        \raisebox{-0.5\height}{\includegraphics[width=0.30\textwidth]{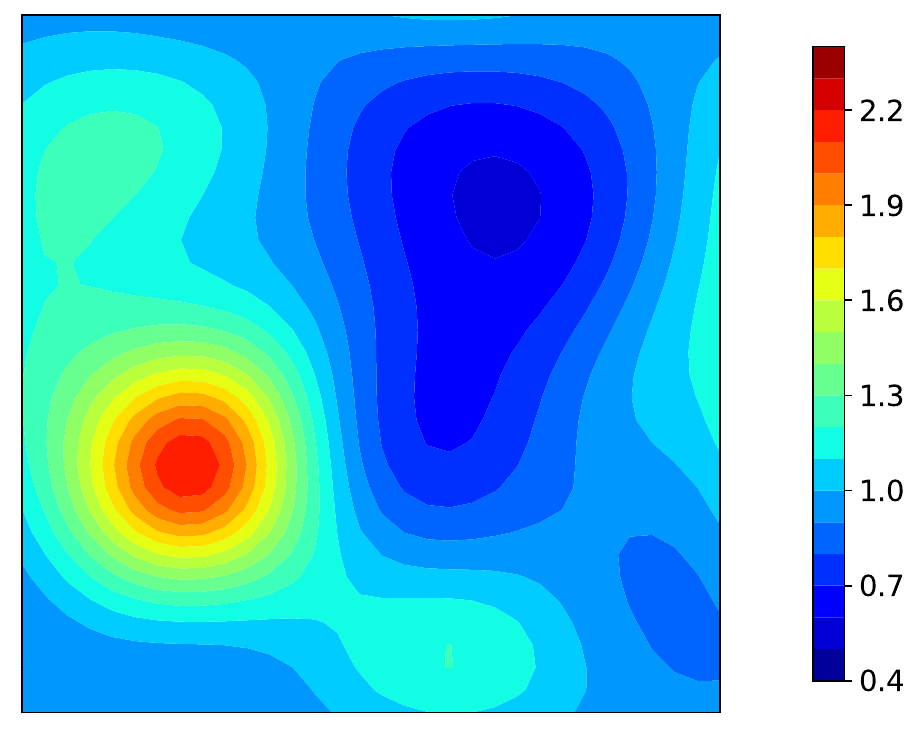}} &
        \raisebox{-0.5\height}{\includegraphics[width=0.3\textwidth]{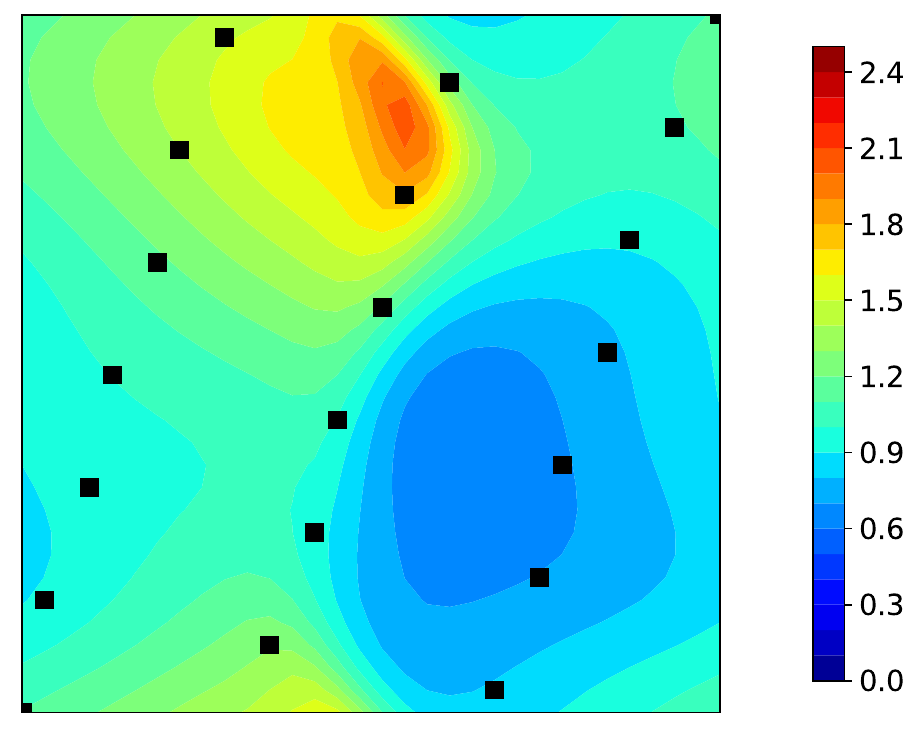}} &
        \raisebox{-0.5\height}{\includegraphics[width=0.30\textwidth]{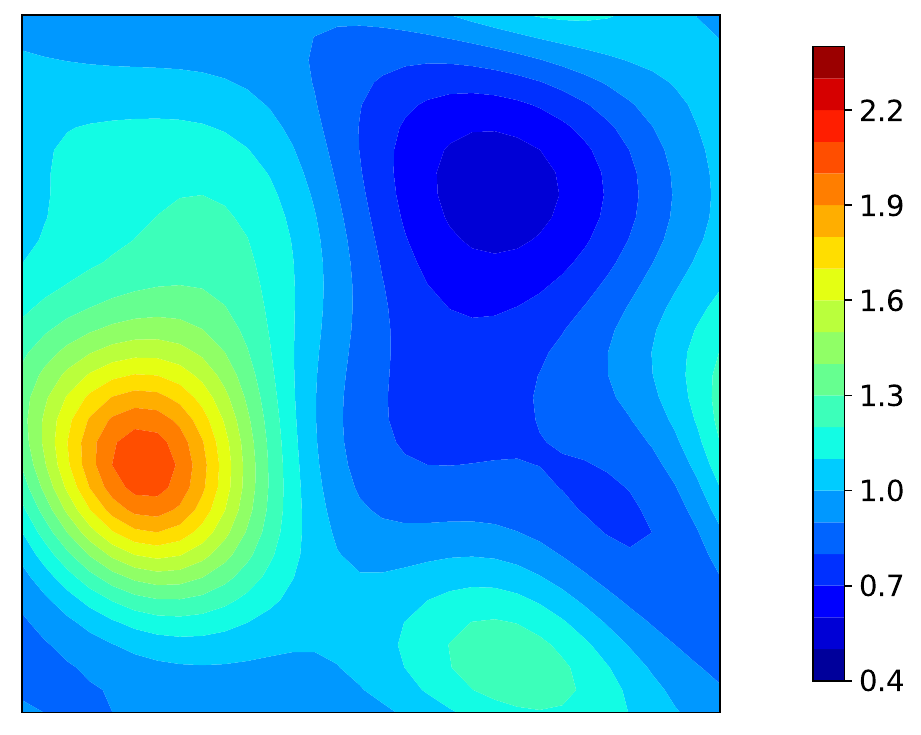}}
    \end{tabular*}
    \captionof{figure}{\textbf{2D Burger's equations, Case II.} 
    Predicted (reconstructed) initial $x-$velocity component for an unseen noisy test sample obtained by \nDNN{}, \mcDNN{}, and  \TNet{} neural networks along with the Tikhonov reconstruction for $n_b = 100$ and $n_t = 2000$.  Shown in the middle column are the synthetic ground truth (Exact) $x-$velocity and the corresponding $x-$velocity at the final time for reference.}
    \figlab{2D_Heat_20_database_predicted_test_samples
    }
    \figlab{2D_Burger_50_database_predicted_test_samples}
\end{figure}


\end{document}